\documentclass[10pt,twocolumn,letterpaper]{article}

\usepackage{cvpr}              
\usepackage{todonotes}

\usepackage{enumitem}

\usepackage{placeins}

\definecolor{cvprblue}{rgb}{0.21,0.49,0.74}
\usepackage[pagebackref,breaklinks,colorlinks,allcolors=cvprblue]{hyperref}

\def\paperID{14} 
\def\confName{MORSE}
\def\confYear{2026}

\title{DINO Soars: DINOv3 for Open-Vocabulary Semantic Segmentation of Remote Sensing Imagery}

\author{Ryan Faulkenberry\\
University of Houston\\
{\tt\small rfaulken@cougarnet.uh.edu}
\and
Saurabh Prasad\\
University of Houston\\
{\tt\small sprasad2@central.uh.edu}}

\usepackage[T1]{fontenc}
\begin{document}
\twocolumn[{%
\renewcommand\twocolumn[1][]{#1}%
\maketitle

\hspace{0.05\linewidth}
\includegraphics[width=.27\linewidth]{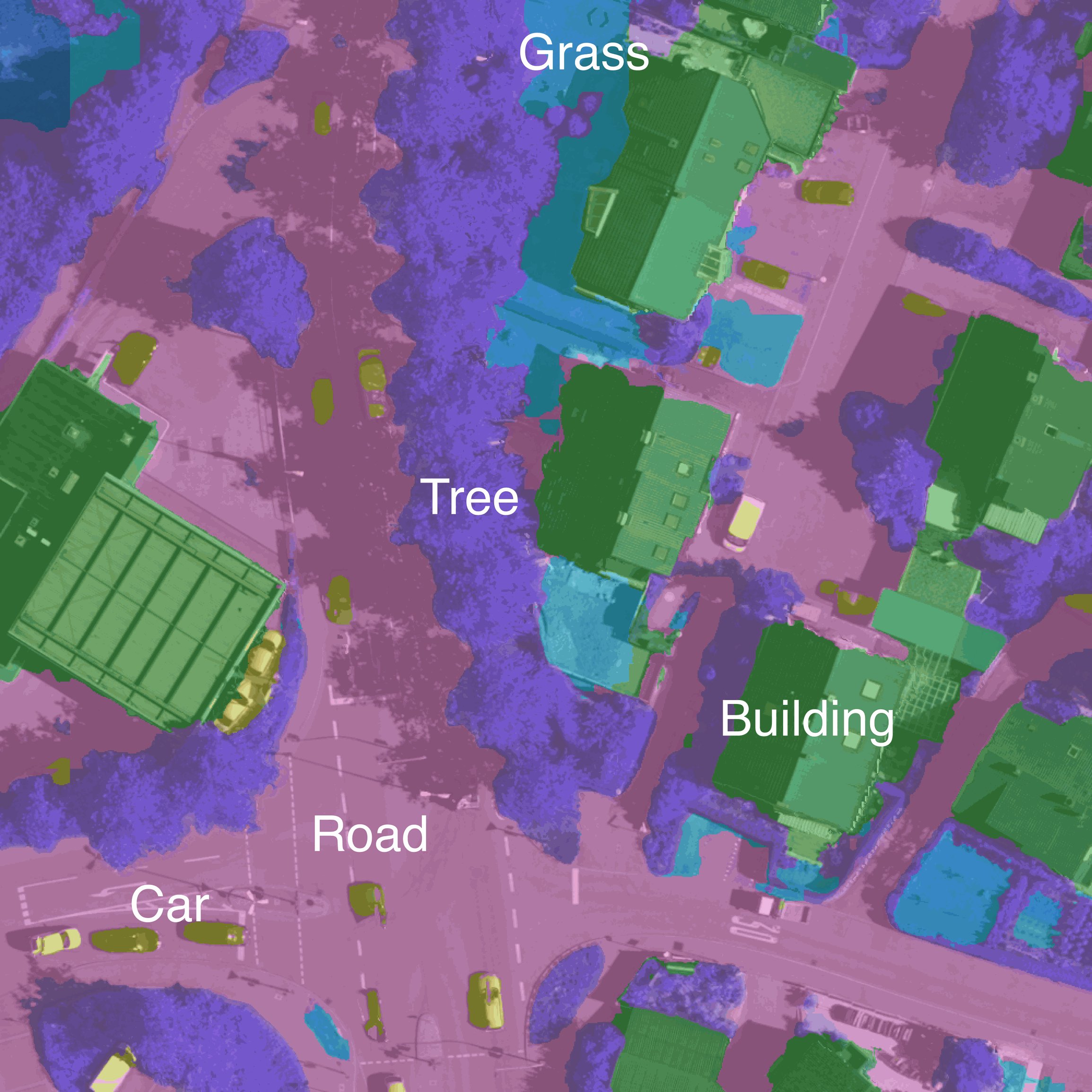}
\includegraphics[width=.27\linewidth]{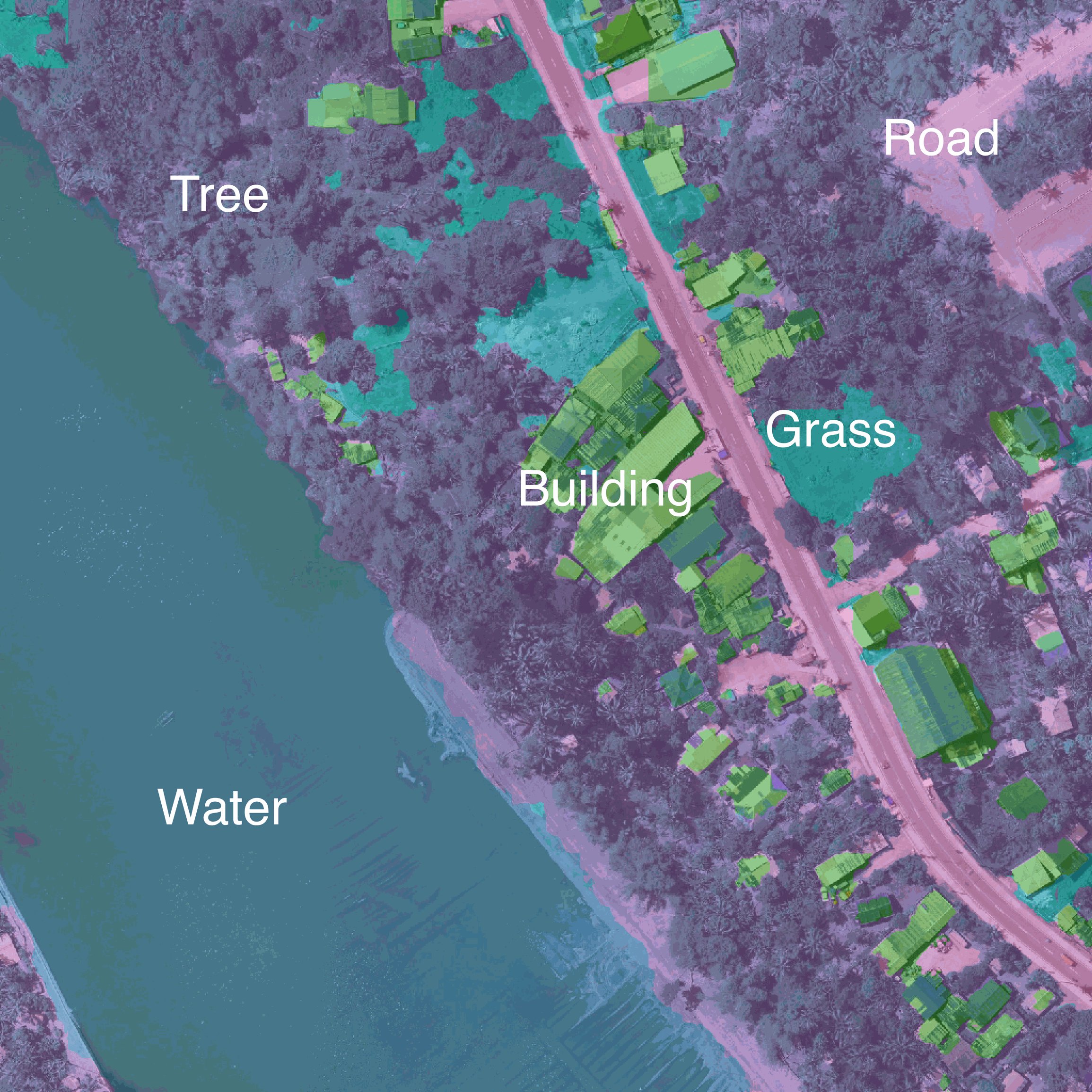}
\includegraphics[width=.27\linewidth]{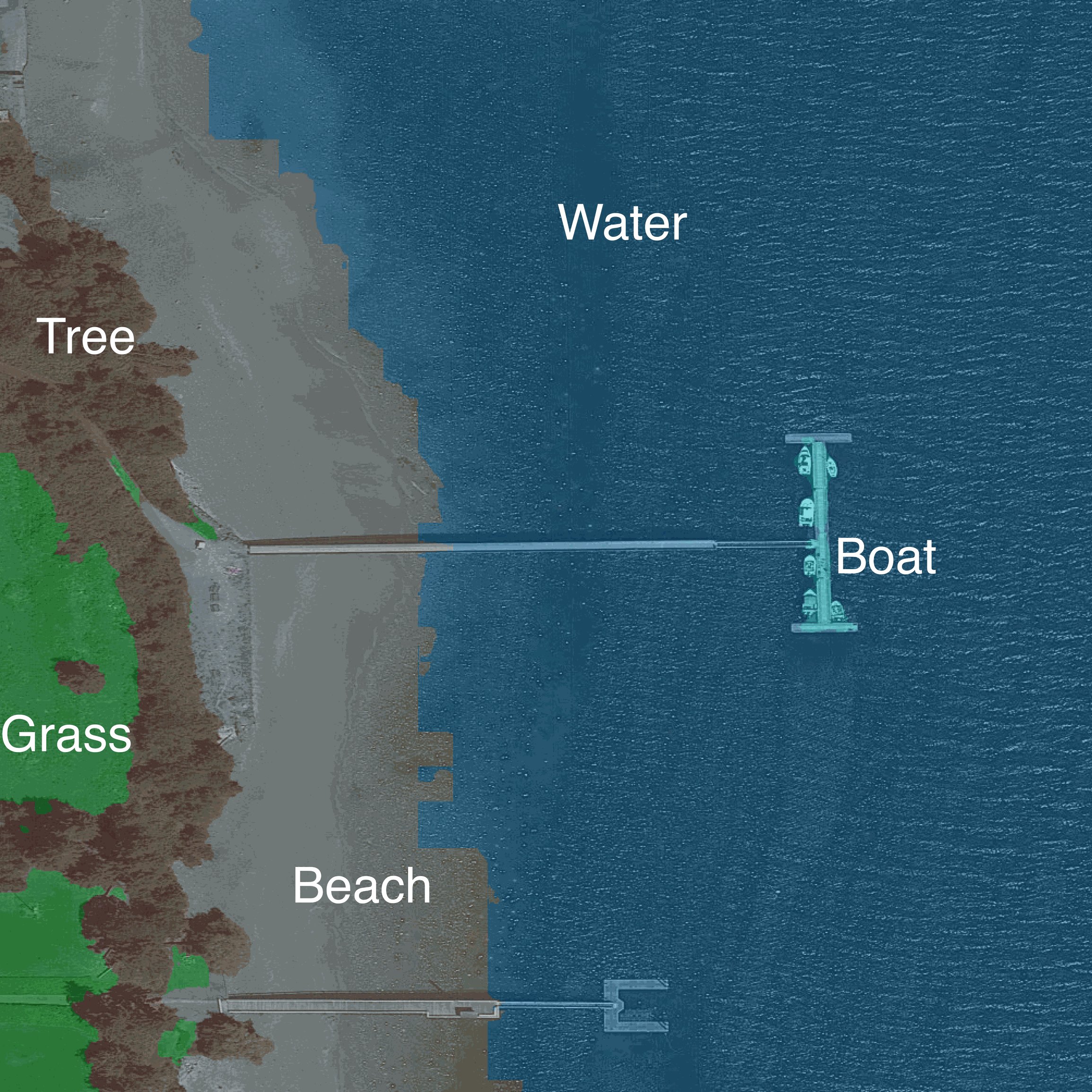}

\captionof{figure}{Open-vocabulary segmentation maps from our model, CAFe-DINO. CAFe-DINO can accurately segment remote sensing scenes with arbitrary semantic classes, despite zero training on remote sensing imagery. \vspace{1em}
}
\label{fig:teaser}
}]

\begin{abstract}
The remote sensing (RS) domain suffers from a lack of densely labeled datasets, which are costly to obtain. Thus, models that can segment RS imagery well without supervised fine-tuning are valuable, but existing solutions fall behind supervised methods. 
Recently, DINOv3 \cite{simeoni2025dinov3} surpassed SOTA RS foundation models on the GEO-bench segmentation benchmark without pre-training on RS data. Additionally, DINO.txt \cite{jose2024dinov2meetstextunified} has enabled open vocabulary semantic segmentation (OVSS) with the DINOv3 backbone. We leverage these developments to form an OVSS model for RS imagery, free of RS-domain fine-tuning. Our model, CAFe-DINO (Cost Aggregation + Feature Upsampling with DINO) exploits the strong OVSS performance of DINOv3 for RS imagery via cost aggregation and training-free upsampling of text-image similarity scores. The robust latent of the DINOv3 backbone eliminates the need for fine-tuning on RS imagery; we instead fine-tune our model on a RS-targeted subset of COCO-Stuff. CAFe-DINO achieves state-of-the-art performance on key RS segmentation datasets, outperforming OVSS methods fine-tuned on RS data. Our code and data are publicly available at https://github.com/rfaulk/DINO\_Soars.
\end{abstract}    
\section{Introduction}
\label{sec:intro}

\begin{figure*}[t]
  \centering
  \begin{subfigure}{0.33\linewidth}
    \includegraphics[width=0.9\textwidth]{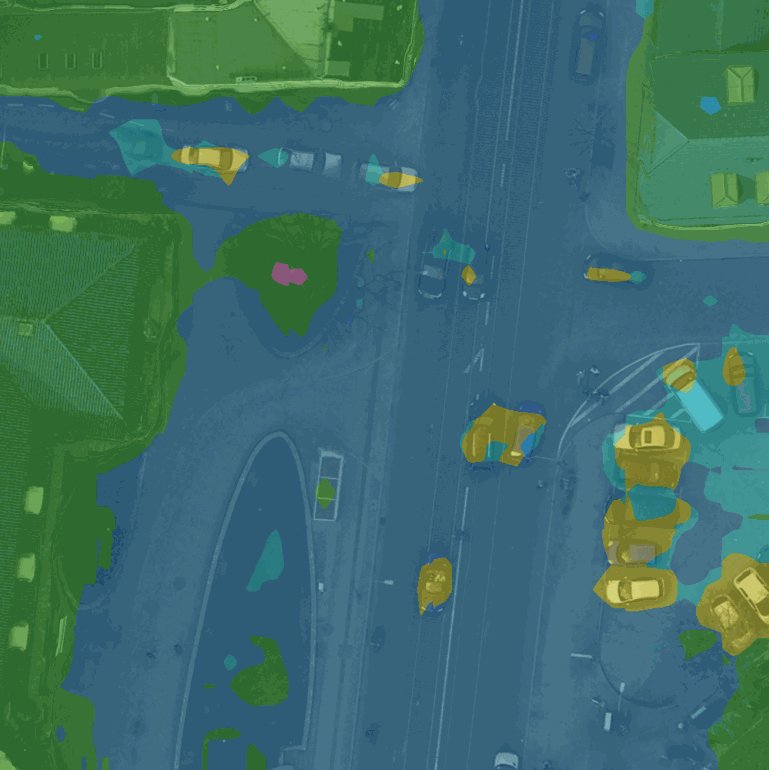}
    \caption{DINOv3.txt}
    \label{fig:seg-dino}
  \end{subfigure}
  \hfill
  \begin{subfigure}{0.33\linewidth}
    \includegraphics[width=0.9\textwidth]{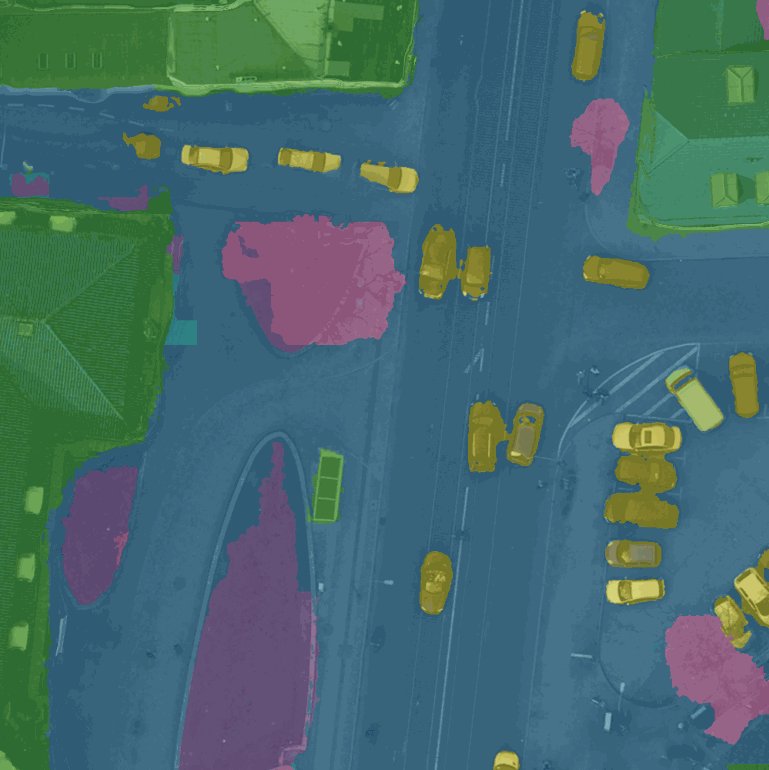}
    \caption{DINOv3.txt + CA + AnyUp (Ours)}
  \end{subfigure}
  \begin{subfigure}{0.33\linewidth}
    \includegraphics[width=0.9\textwidth]{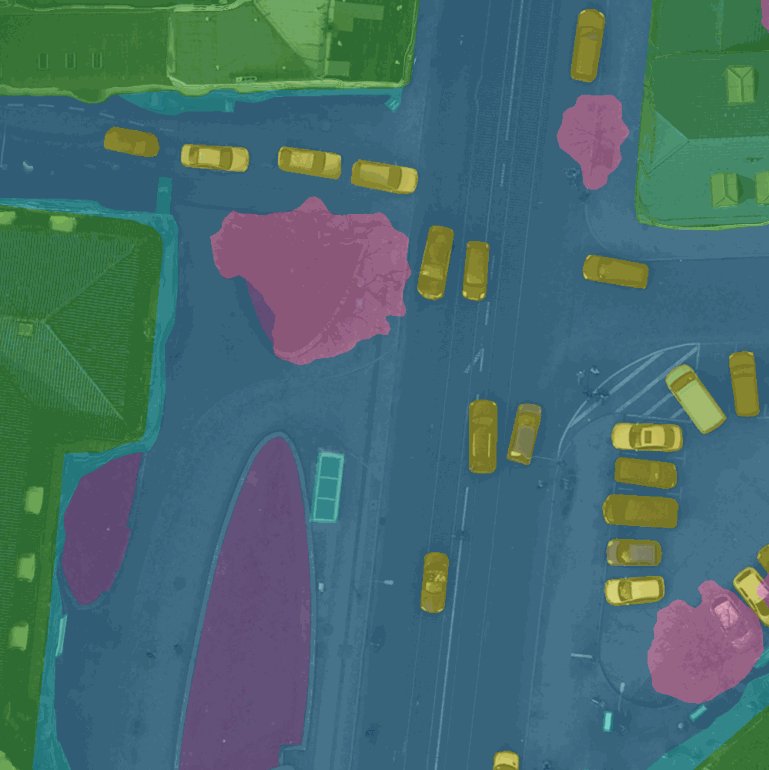}
    \caption{Ground Truth}
  \end{subfigure}
  \caption{DINOv3.txt alone is not a strong OVSS model for remote sensing}
  \label{fig:seg}
\end{figure*}

Remote sensing (RS) imagery has revolutionized our ability to observe the planet. It has equipped us with the ability to monitor our environment, optimize agriculture, and respond to natural disasters. RS imagery does not enjoy the same wealth of densely labeled datasets as natural imagery, which has led to a body of work on adapting models trained on natural imagery to fit the unique characteristics of RS imagery, with little or no fine-tuning.

For semantic segmentation tasks in RS, the latest and most successful of these models are Open Vocabulary Semantic Segmentation (OVSS) models. They leverage vision-language models (VLMs) that are pretrained to align a large corpus of image and text pairs. These models, imbued with a large ``vocabulary'' from their pretraining, can be prompted with the name of a desired class to create a similarity map between that class and an input image, enabling zero-shot segmentation of arbitrary classes.

The recent release of DINOv3 \cite{simeoni2025dinov3} included a text encoder aligned with DINOv3 image features, enabling OVSS via alignment of image features and text embeddings of semantic classes \cite{jose2024dinov2meetstextunified}. Their ``DINOv3.txt'' model surpasses OVSS state-of-the-art by a large margin on natural imagery benchmarks.
DINOv3 has also shown remarkable adaptability to RS tasks without any pre-training on RS imagery, outperforming RS-specific foundation models (e.g. Prithvi \cite{Prithvi-EO-V2-preprint}, DOFA \cite{xiong2025}), as well as a sister version of DINOv3 trained on RS imagery, on five of six GEO-Bench \cite{lacoste2023} segmentation tasks. It is noteworthy that DINOv3 achieved this while using only the RGB bands of these mostly multi-spectral datasets. This suggests that natural image training at scale can overcome RS-specific obstacles such as scale/resolution differences, sensor noise, and varying spectral characteristics. Motivated by these results, we look to DINOv3.txt as a starting point for a RS OVSS model trained without RS imagery.


Despite its pedigree on natural imagery, DINOv3.txt struggles with OVSS on RS imagery, as we show in \cref{fig:seg-dino} and \cref{tab:results}. We unlock the RS capability of DINOv3 for OVSS by adding a cost aggregation module to semantically and spatially refine the similarity maps of DINOv3.txt. Additionally, motivated by the recent success of feature upsampling \cite{fu2024featup, suri2025lift, couairon2025jafar, wimmer2025anyup}, we incorporate AnyUp \cite{wimmer2025anyup} to upsample the aggregated features directly without fine-tuning. The combination of AnyUp, which is training-free, and the cost aggregation module, which learns only to denoise similarity maps from the existing DINO vision/text encoders, creates a synergistic architecture in which RS fine-tuning is not required.

\textbf{Contributions:}

\begin{enumerate}
    \item We introduce CAFe-DINO, a cost-aggregation and feature upsampling model for the DINOv3 backbone that does not require RS fine-tuning. We demonstrate the efficacy of cost aggregation and feature upsampling for OVSS of RS imagery.
    \item Leveraging the capability of DINOv3 on RS imagery, we train CAFe-DINO on only natural imagery with an RS-targeted subset of the COCO-Stuff \cite{caesar2018cvpr} dataset.
    \item We show state of the art OVSS performance on a set of multiclass RS segmentation tasks with CAFe-DINO. 
\end{enumerate}

\begin{figure}[htbp]
  \centering
  \begin{subfigure}[b]{\linewidth}
    \centering
    \includegraphics[width=\linewidth]{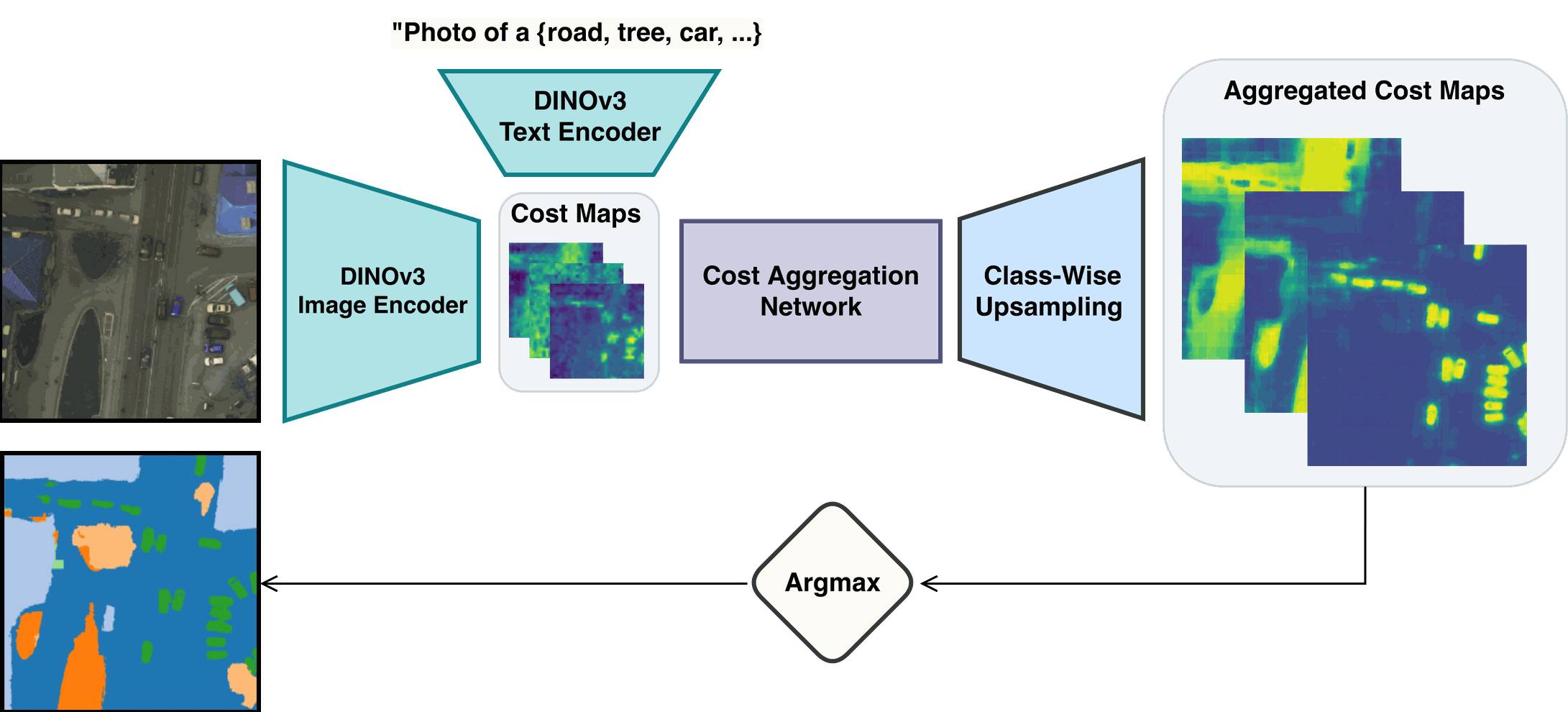} 
    \caption{An overview of our method. Class-wise similarity maps from DINOv3 are aggregated and upsampled to form a segmentation prediction.}
    \label{fig:block1}
  \end{subfigure}

  \vspace{6pt} 

  \begin{subfigure}[b]{\linewidth}
    \centering
    \includegraphics[width=\linewidth]{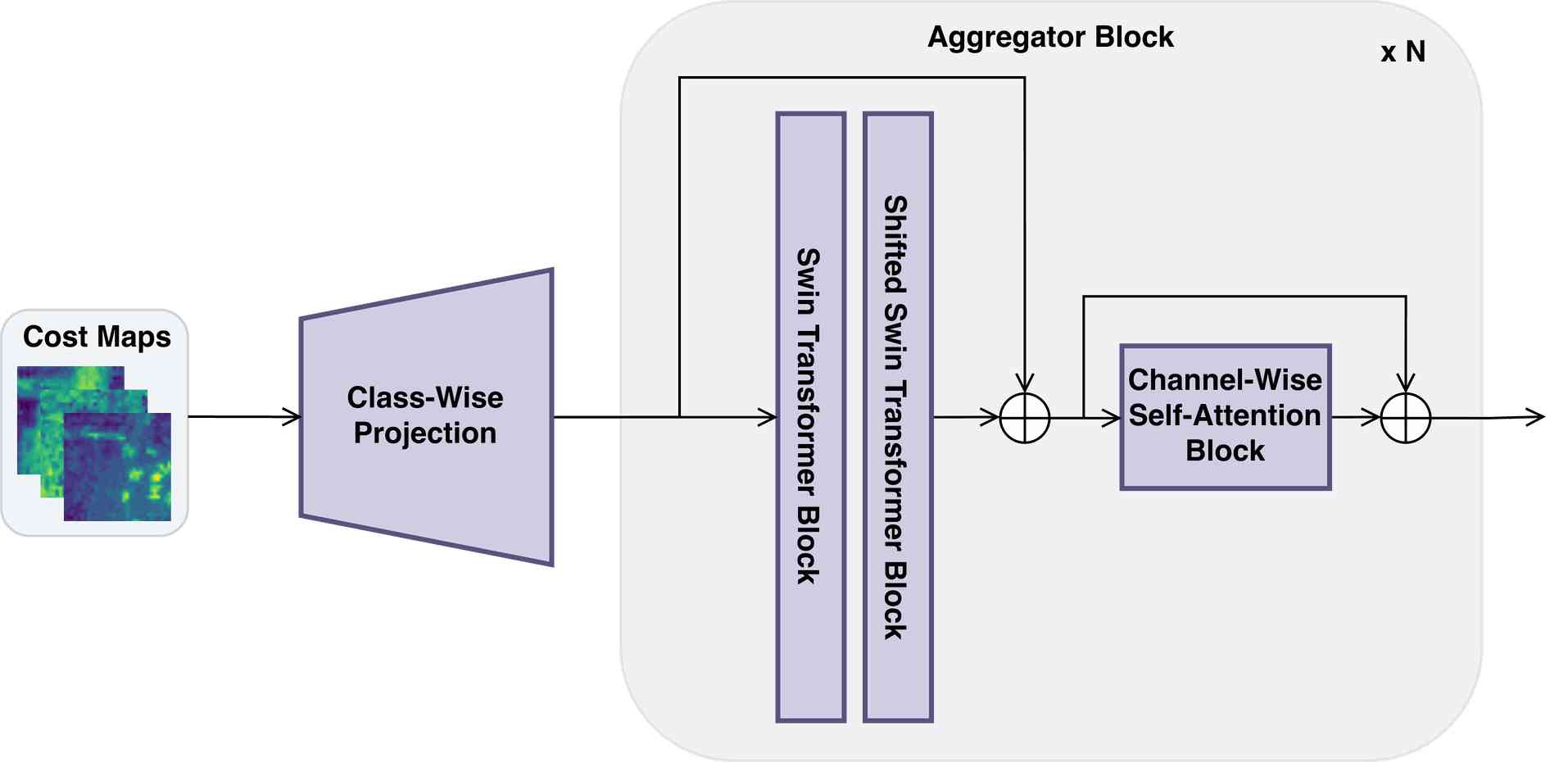} 
    \caption{An overview of the cost aggregation network. Cost maps are projected and propagated through a series of aggregation blocks.}
    \label{fig:block2}
  \end{subfigure}

  \caption{Details of the CAFe-DINO architecture.}
  \label{fig:stacked}
\end{figure}
\section{Related Work}
\label{sec:related}

\subsection{Open Vocabulary Semantic Segmentation}
Open vocabulary semantic segmentation (OVSS) \cite{10558790} has advanced significantly with vision-language models (VLMs) \cite{radford2021, li2023, cherti2023reproducible, li2022blip} that imbue an image encoder with text-conditioned reasoning. CLIP \cite{radford2021} introduced a contrastive method for aligning image and text embeddings, attaining state-of-the-art performance on zero-shot image classification in its time. CLIP has since been adapted for application to segmentation tasks and forms the backbone of most OVSS models to this day \cite{ghiasi2022, ding2022, li2022languagedriven, liu2024a, 10238837}. Early adaptations such as SegCLIP \cite{zhang2024} and PACL \cite{mukhoti2023} introduced image patches rather than a global image embedding as an alignment objective. Later work added more sophisticated techniques, such as mask proposal-based region segmentation \cite{liang2023open, barsellotti2024, kang2024, shao2025, clip_as_rnn}, and CAT-Seg \cite{cho2024a}, which incorporated cost aggregation (See \cref{sec:costagg}) of aligned CLIP features for segmentation.

Some methods have managed to eschew CLIP and still achieve good performance. One such class of methods \cite{wang2023diffusion, lan2024proxyclip} apply text-guided diffusion models to generate embeddings for arbitrary classes.

Despite these advances, most OVSS methods still struggle with fine-grained localization and domain shift, particularly when deployed outside natural image distributions such as RS.

\subsection{OVSS for Remote Sensing}
Because OVSS for remote sensing has only recently emerged, there are only a few methods published. GSNet \cite{ye2025} uses a DINO-based RS backbone as guidance for a frozen CLIP model and trains on a newly introduced large-scale semantic segmentation dataset for RS. AerOSeg \cite{dutta2025} uses SAM to spatially refine CLIP features and guide upsampling. T2ASeg \cite{11197568} uses gradient activation maps to augment CLIP embeddings. OVRS \cite{10962188} applies some RS-targeted modifications to the CAT-Seg framework, such as rotational invariance encoding and feature-guided upsampling. SegEarth-OV \cite{li2025segearthov} directly upsamples CLIP features with guidance from the input image, then subtracts the [CLS] token to extract localized features; notably, they require only self-supervised \cite{uelwer2025} training on RS data, though on a per-dataset basis. Collectively, these methods demonstrate the promise of OVSS in RS, but are limited by the need to train with RS imagery. 

\subsection{DINOv3}
DINOv3 \cite{simeoni2025dinov3} is a very large vision foundation model applicable to a variety of visual tasks. DINOv3 is based on a ViT \cite{dosovitskiy2020} scaled to 7B parameters, trained via self-supervised learning (SSL) on a pool of over 1 billion unlabeled images. The 7B-parameter model is distilled into versions matching the usual ViT sizes (ViT-H, ViT-B, etc).

DINOv3 outperforms established geospatial foundation models (GFMs) \cite{xiong2024neural, tolan2024, Prithvi-EO-V2-preprint} on both classification and segmentation tracks of the GEO-Bench \cite{lacoste2023} benchmark, despite using only RGB channels, whereas GFMs use all available bands. In fact, the authors additionally pre-train a DINOv3 model on satellite imagery and find that it does not match the performance of the natural-imagery-trained version on GEO-Bench. This result suggests that general-purpose SSL on natural imagery can surpass domain-specific training. We capitalize on this property by leveraging DINOv3 as the backbone for our OVSS model in an RS-free training regime.

\subsection{DINO.txt}

DINO.txt \cite{jose2024dinov2meetstextunified} was originally released for DINOv2, but has since been integrated into DINOv3 without modification (we distinguish this version with the name ``DINOv3.txt''). This work equipped the DINO backbone with an aligned text encoder to enable vision-language tasks in the style of CLIP. Unlike CLIP, DINO.txt freezes DINO and trains only the text encoder with a similar contrastive objective as CLIP (known as ``locked-image tuning'' \cite{zhai2022}). Whereas CLIP aligns a global image representation ([CLS] token), DINO.txt concatenates average-pooled embeddings of image patches with the [CLS] token, allowing gradients to propagate to specific patches and improving dense feature representation. Its RS efficacy and segmentation-guided design make DINOv3.txt a more optimal backbone than CLIP for RS OVSS tasks.

\subsection{Cost Aggregation}
\label{sec:costagg}
Cost aggregation was originally developed for semantic correspondence tasks \cite{chang2018, 9933865, guo2019} as a mechanism for processing for similarity maps between e.g. images. The modern transformer-based standard for cost aggregation is CATs++ \cite{9933865}, which aggregates individual cost maps spatially and models dependencies between cost maps with separate transformer modules. 

CAT-Seg \cite{cho2024a} introduced cost aggregation for OVSS by framing similarity scores between CLIP text and image embeddings as cost maps. Following the outline of CATs++, CAT-Seg creates a cost map from CLIP for each semantic class of interest, then aggregates, alternating between spatially refining cost maps individually and inter-class feature dependencies. This aggregation refines the raw cost maps into per-class probability maps. After upsampling, an argmax operation across classes forms the resulting prediction. 

\subsection{Feature Upsampling}

Recently, methods have been developed to upsample the deep latent features of foundation models to higher resolutions with high fidelity \cite{fu2024featup, suri2025lift, couairon2025jafar, wimmer2025anyup}. These methods use the input image as guidance to upsample low-resolution latent features. Feature upsamplers are immediately applicable to CLIP-based OVSS methods \cite{li2025segearthov}, because image-text similarity can be computed after image feature upsampling, providing full-resolution similarity scores that can be directly converted to segmentation predictions. Of particular interest is AnyUp \cite{wimmer2025anyup}, an upsampler that is feature-agnostic, i.e. performs well on features from an unseen backbone without fine-tuning. Prior to AnyUp, cost aggregation required trainable upsampling, which fit the training domain but did not generalize to unseen domains without further fine-tuning. By using AnyUp, we avoid training a domain-constrained upsampler and enable RS-data-free fine-tuning.


\section{Method}
\label{sec:method}

\begin{figure*}[t]
    \centering
    \setlength{\tabcolsep}{1.5pt}

    \begin{tabular}{cccccc}
        DINOv3.txt & OVRS & GSNet & SegEarth-OV & CAFe-DINO (Ours)  & Ground Truth \\

        \includegraphics[width=0.16\textwidth]{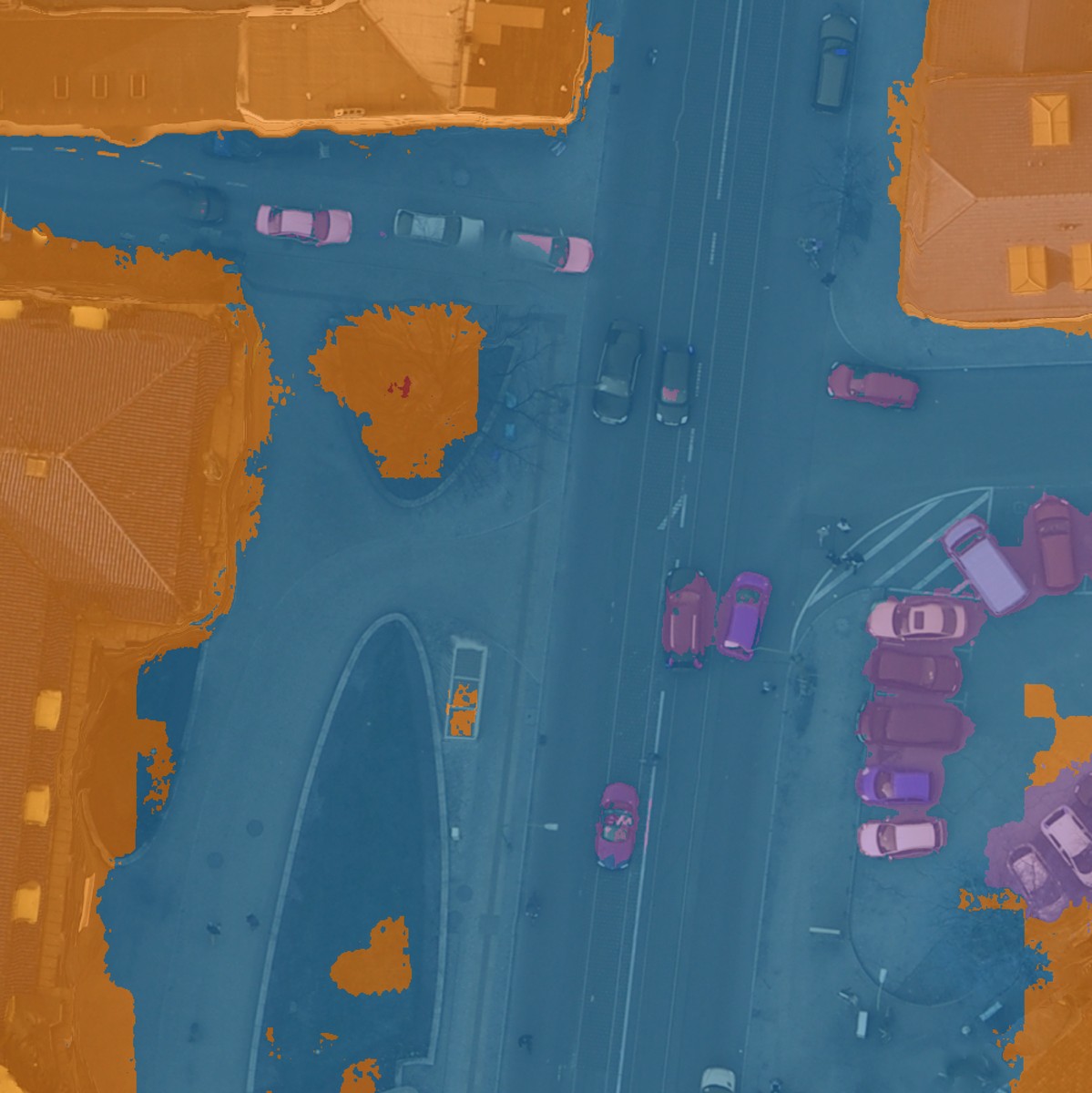} &
        \includegraphics[width=0.16\textwidth]{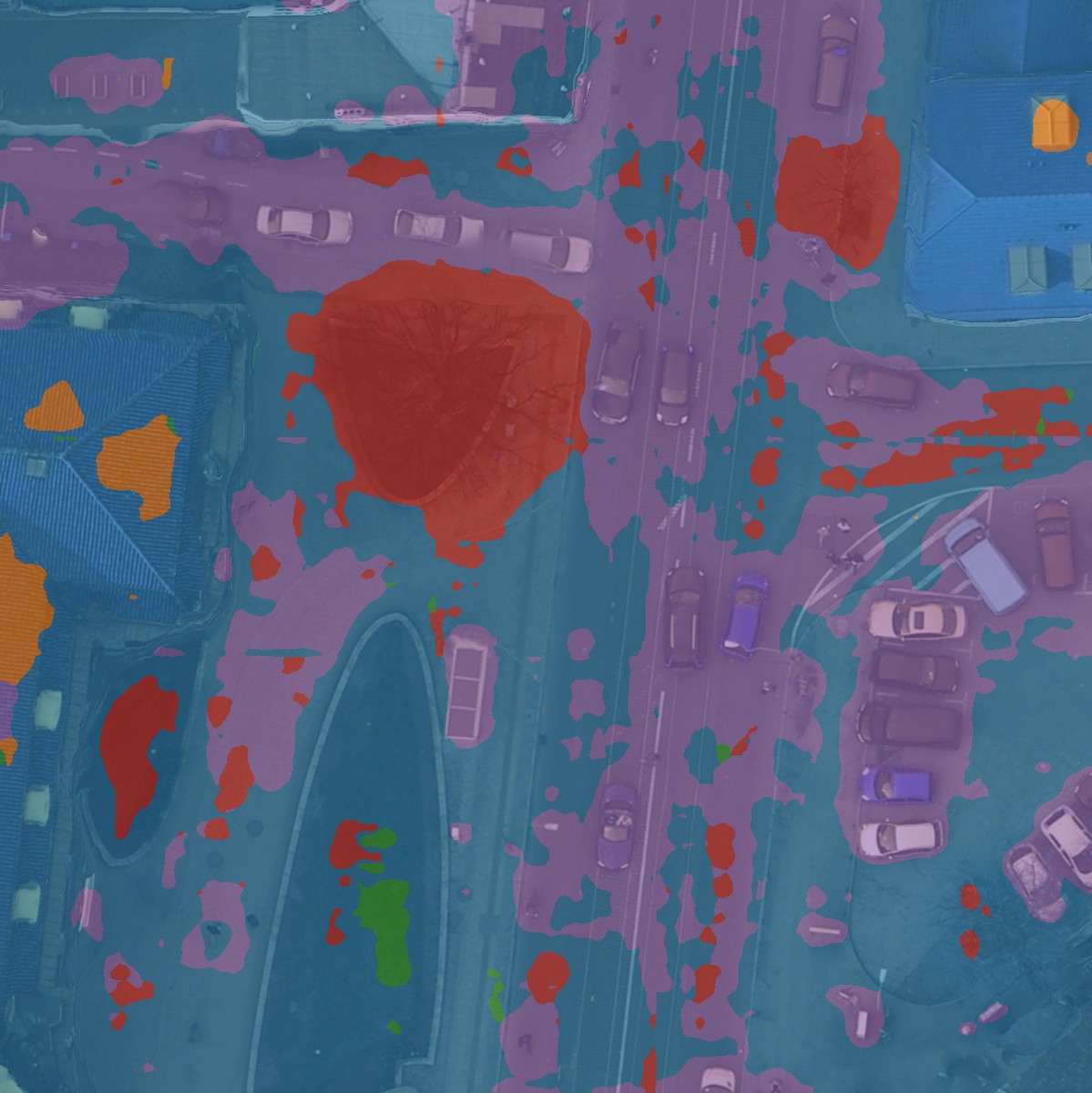}&
        \includegraphics[width=0.16\textwidth]{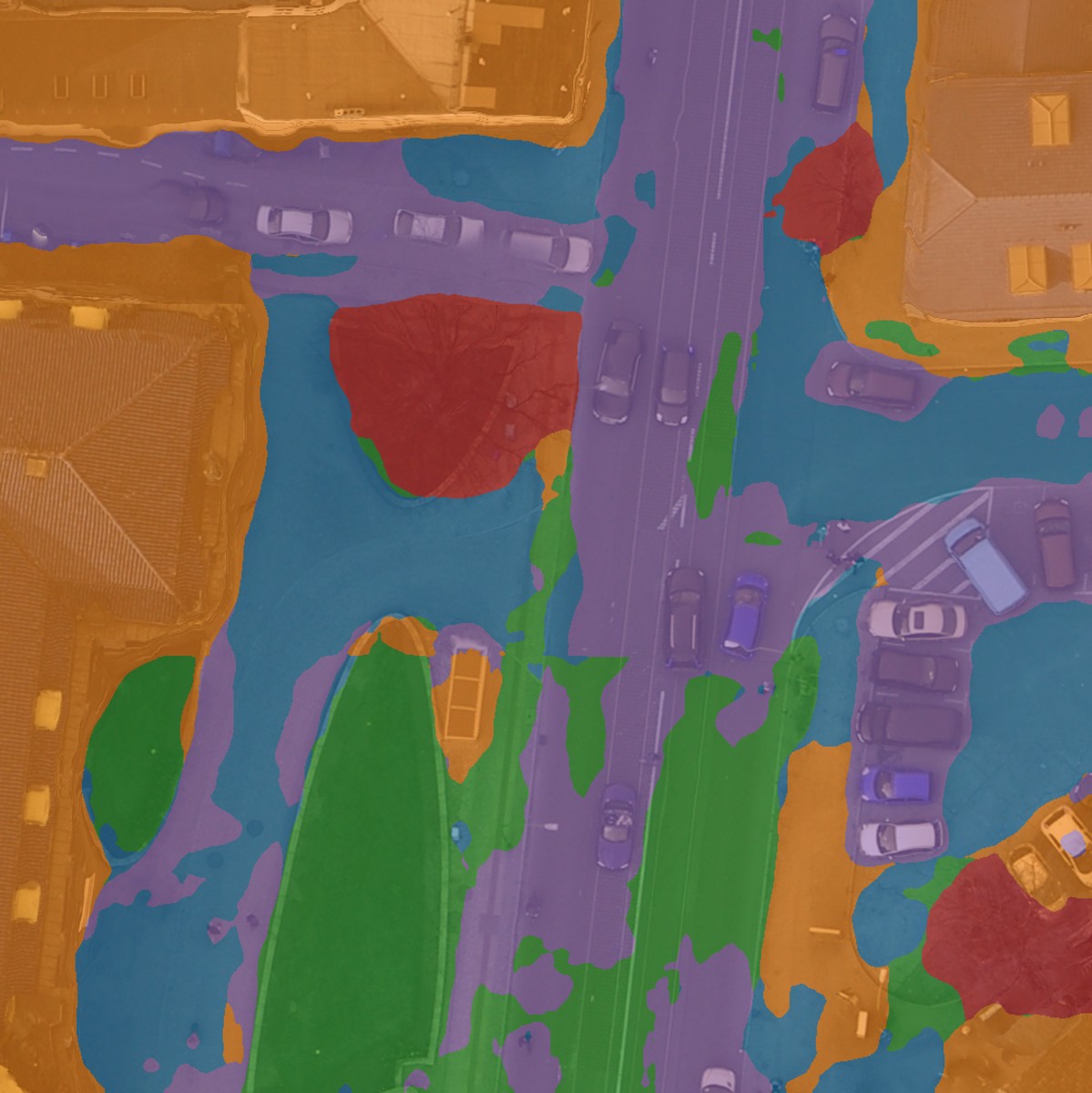}&
        \includegraphics[width=0.16\textwidth]{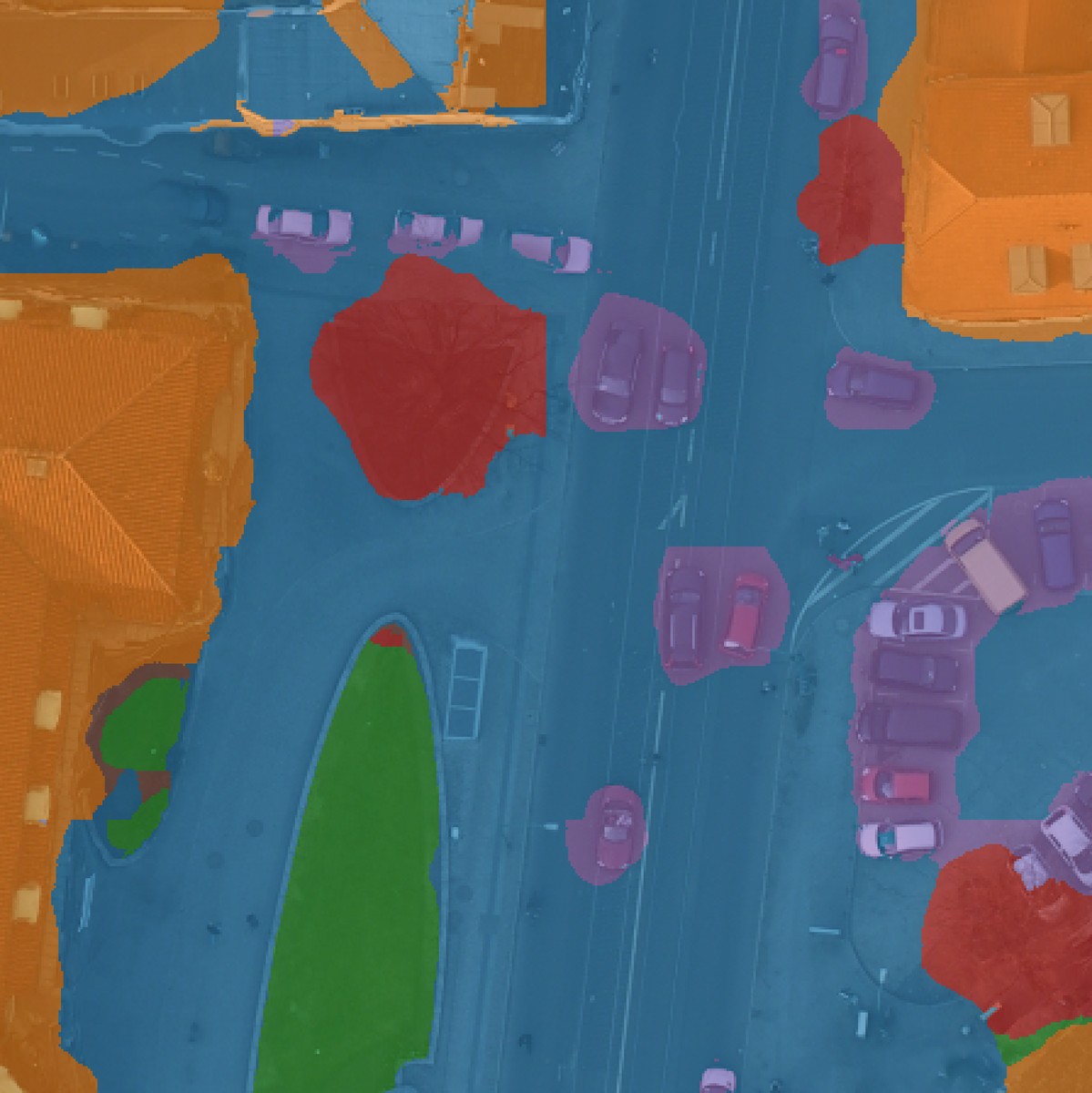}&
        \includegraphics[width=0.16\textwidth]{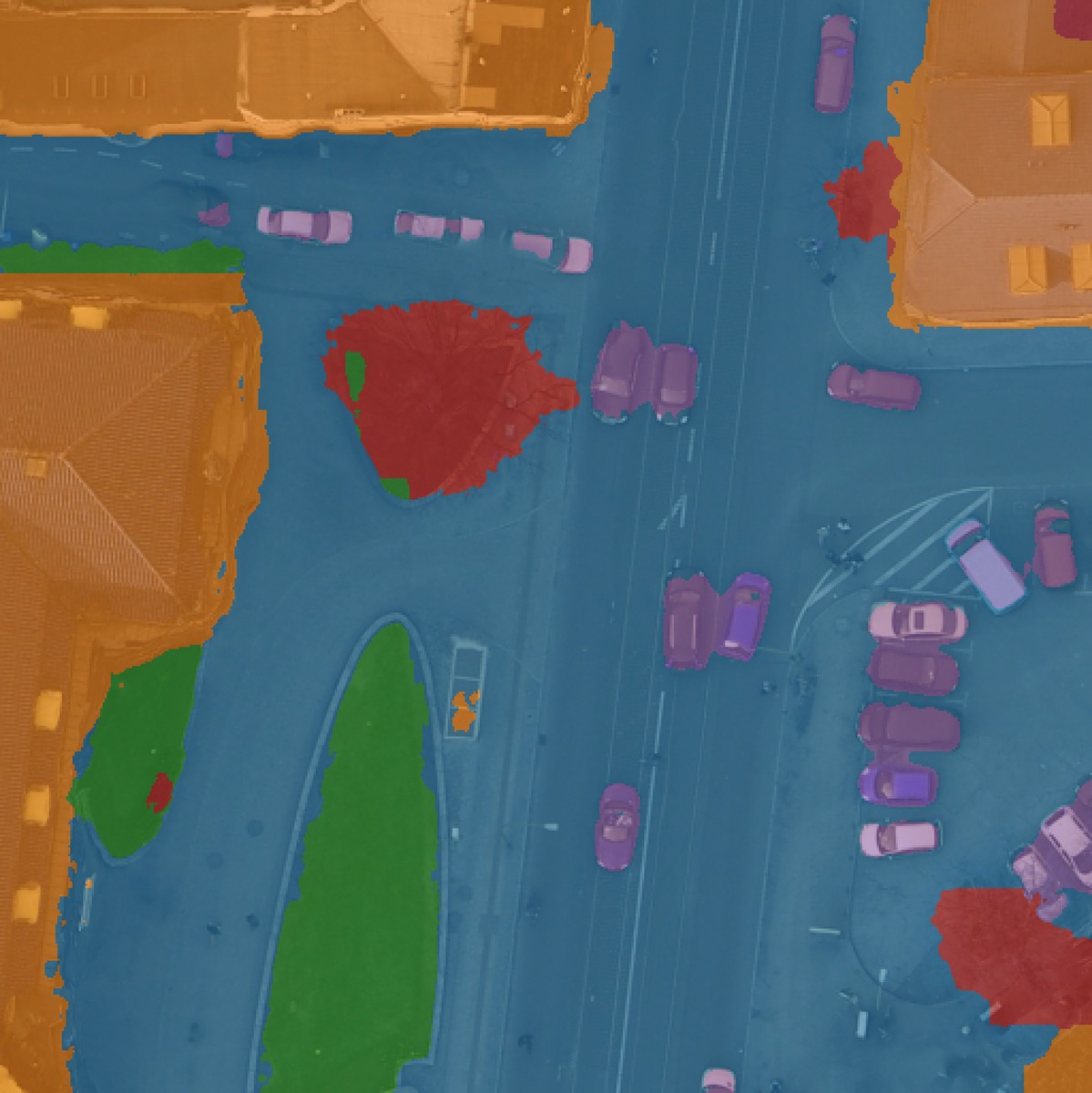}&
        \includegraphics[width=0.16\textwidth]{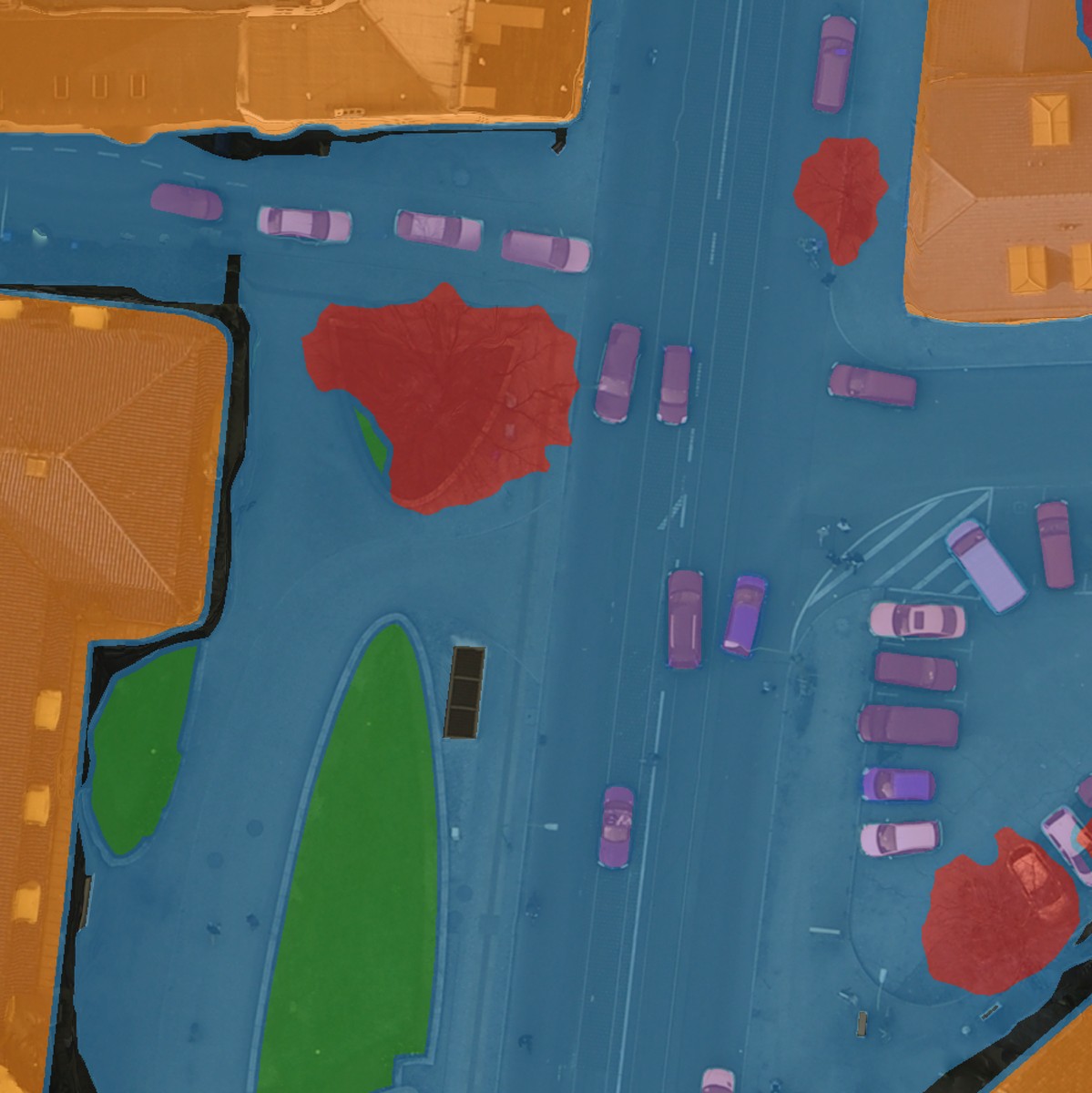} \\[1pt]

        \includegraphics[width=0.16\textwidth]{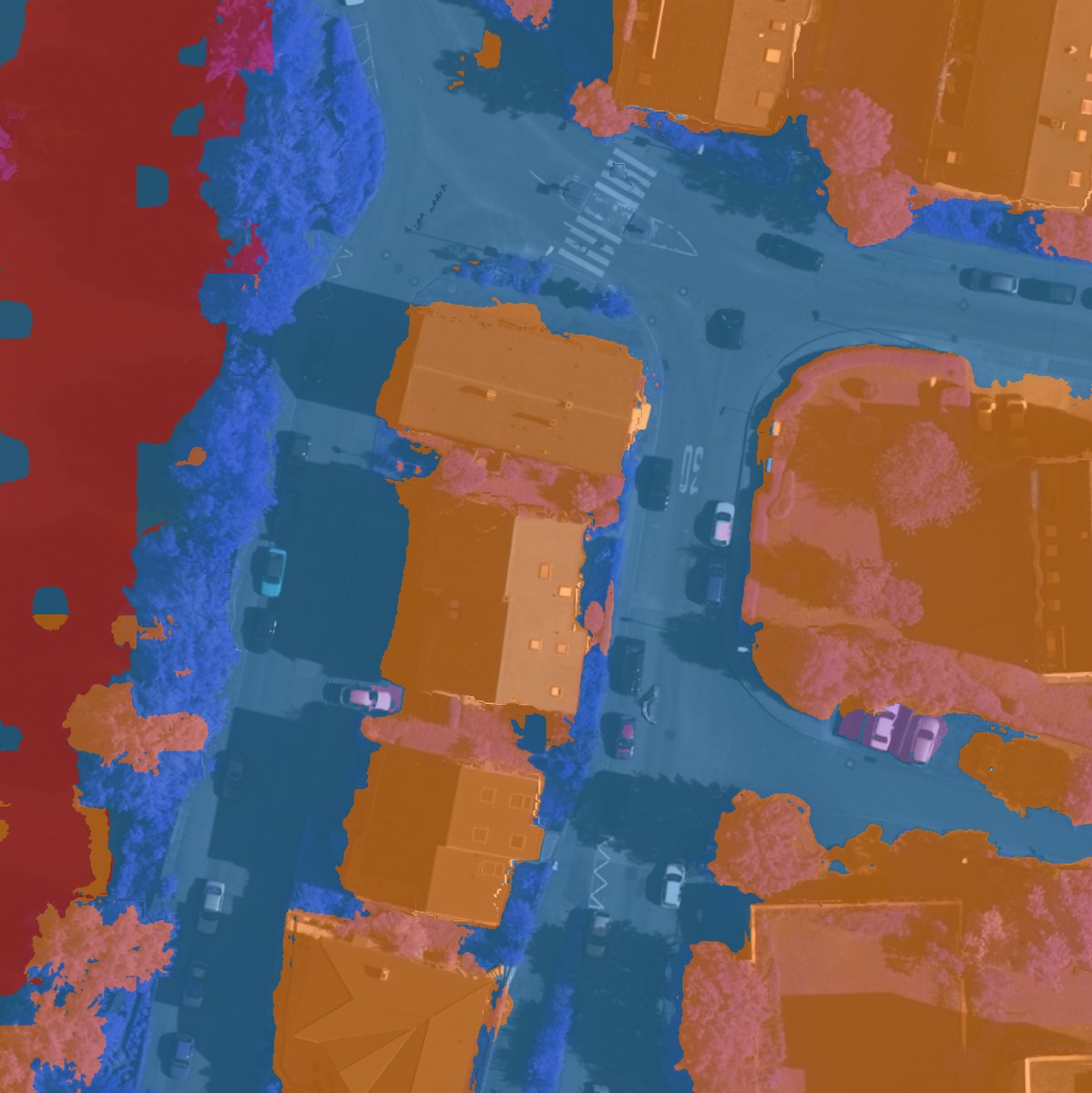} &
        \includegraphics[width=0.16\textwidth]{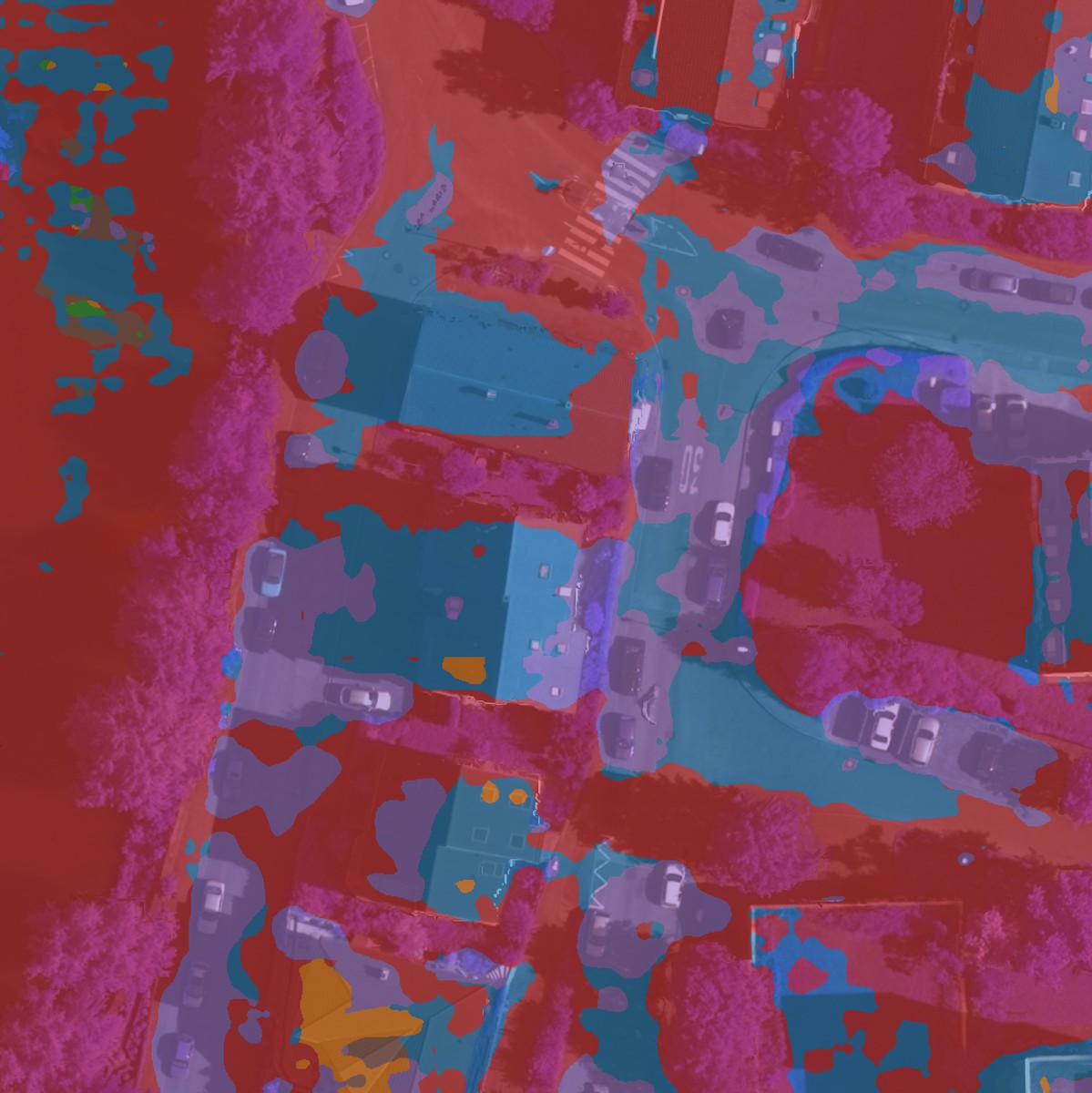}&
        \includegraphics[width=0.16\textwidth]{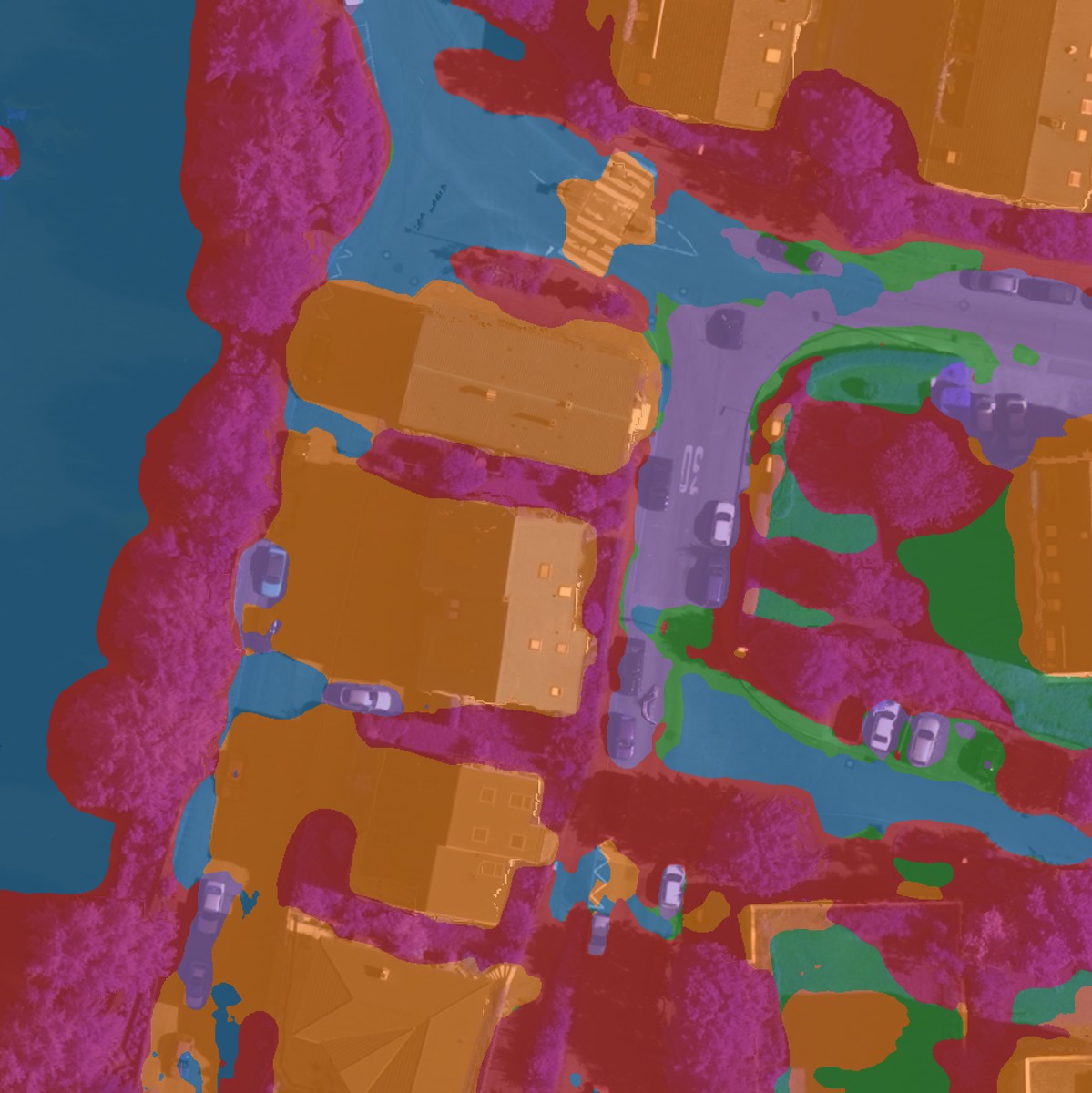}&
        \includegraphics[width=0.16\textwidth]{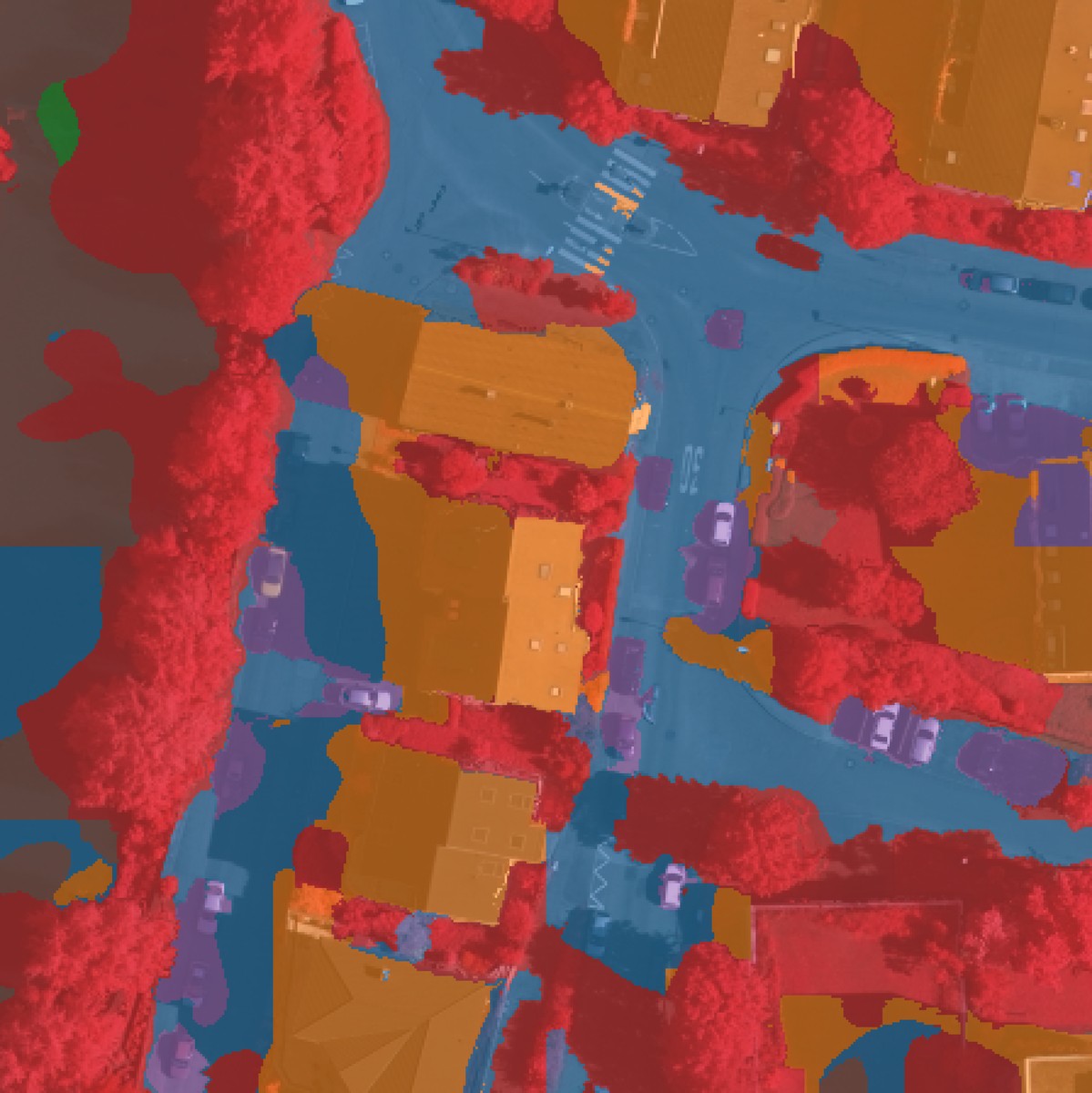}&
        \includegraphics[width=0.16\textwidth]{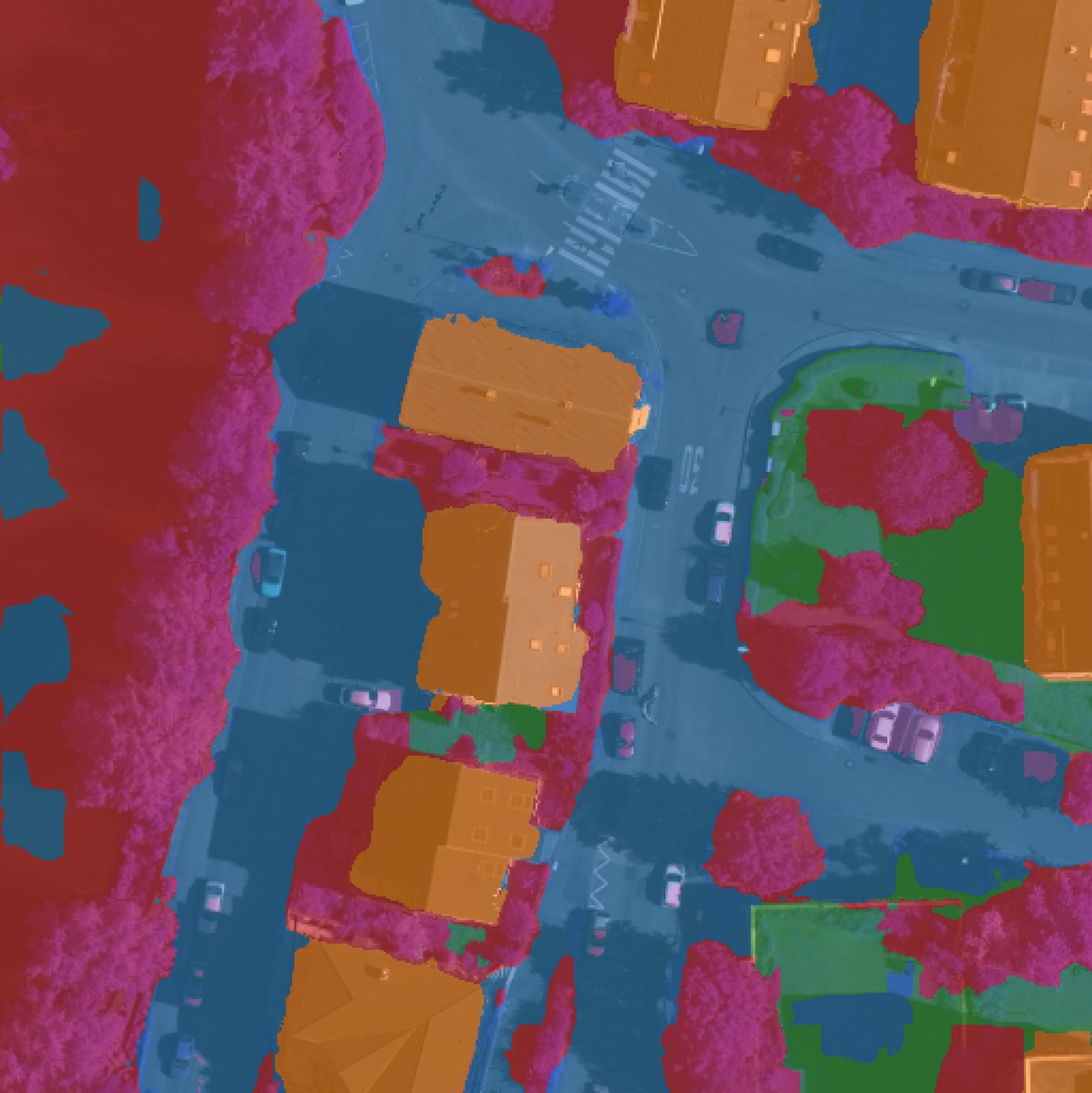}&
        \includegraphics[width=0.16\textwidth]{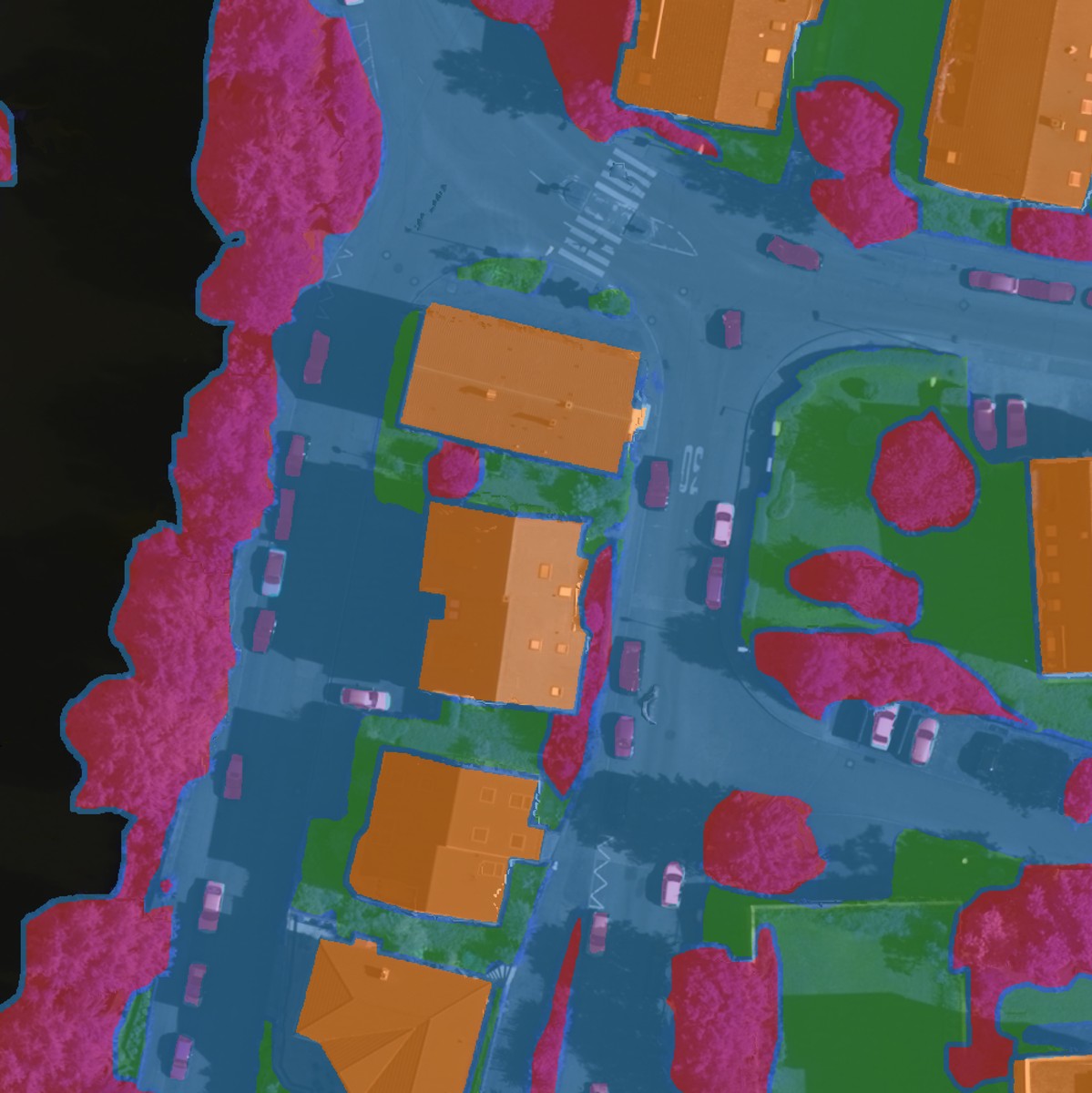} \\[1pt]

        \includegraphics[width=0.16\textwidth]{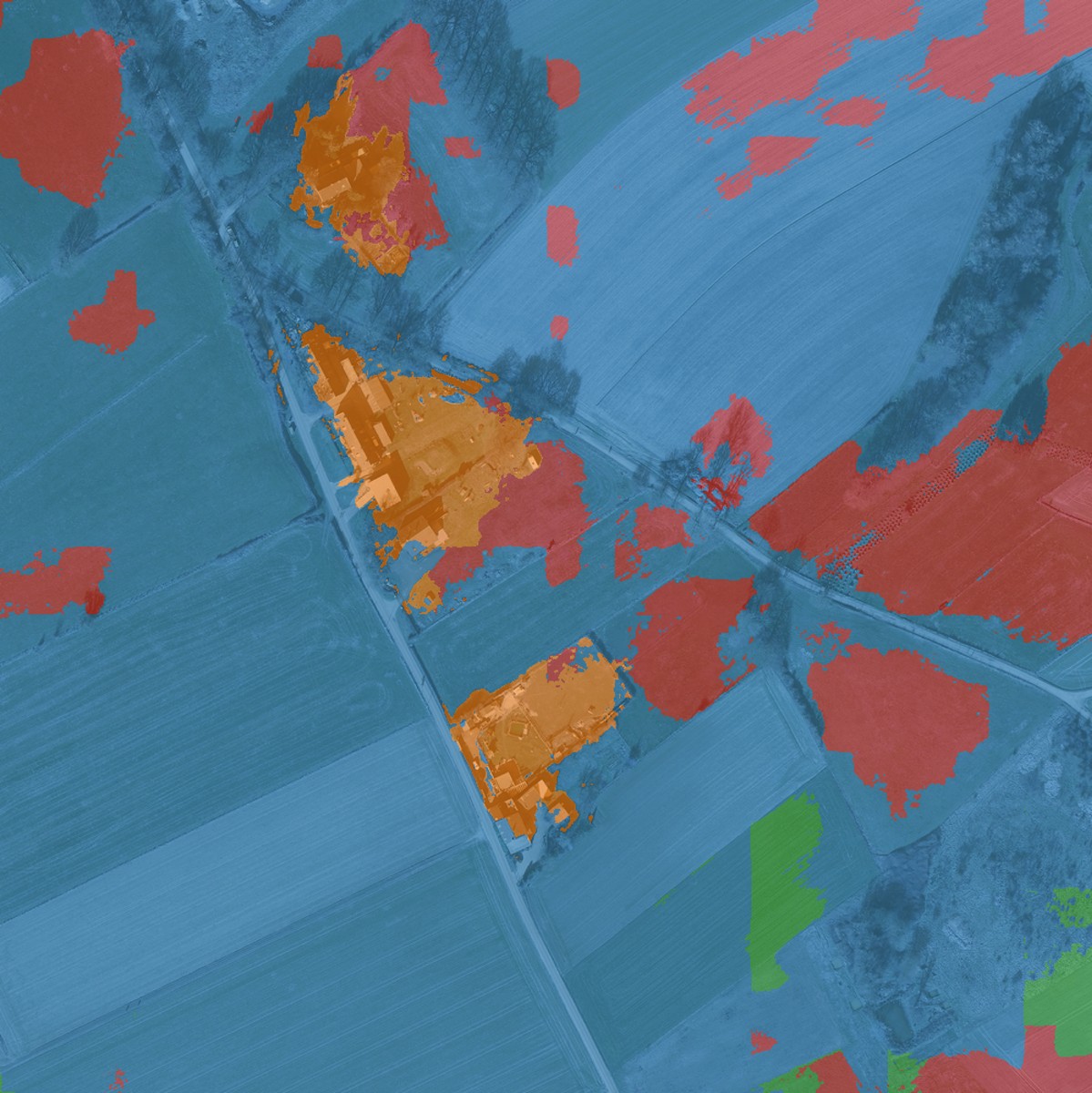} &
        \includegraphics[width=0.16\textwidth]{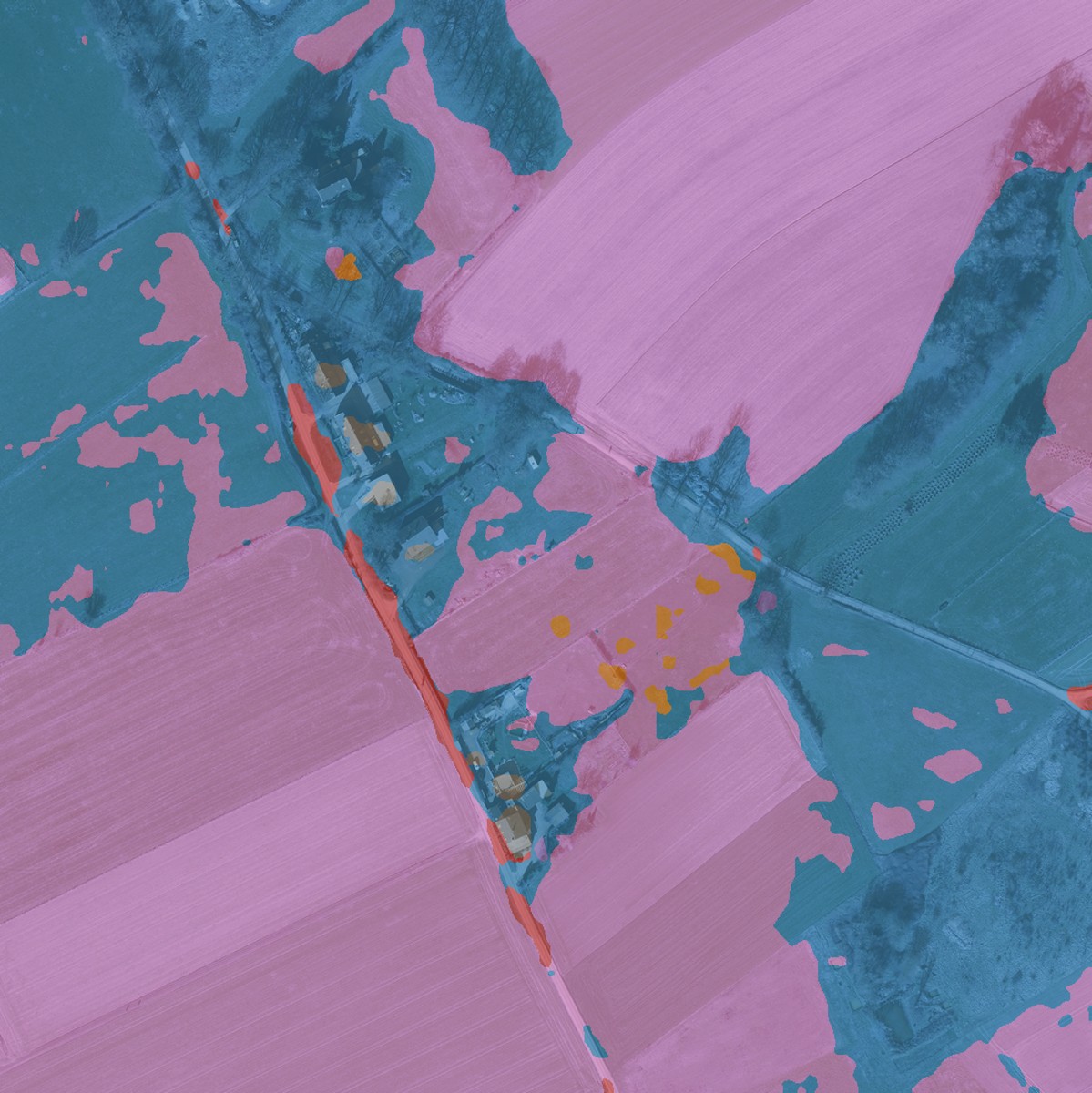}&
        \includegraphics[width=0.16\textwidth]{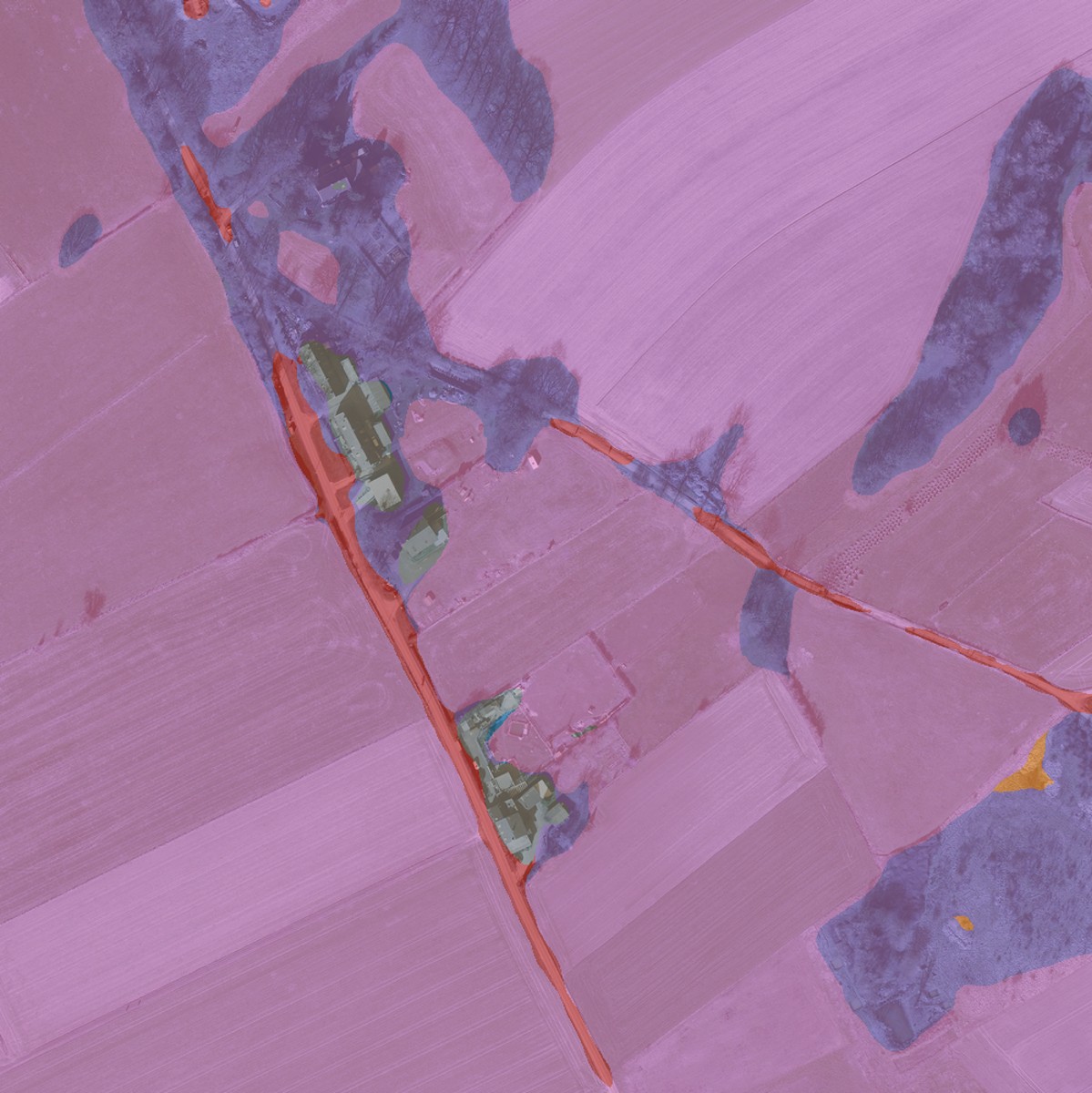}&
        \includegraphics[width=0.16\textwidth]{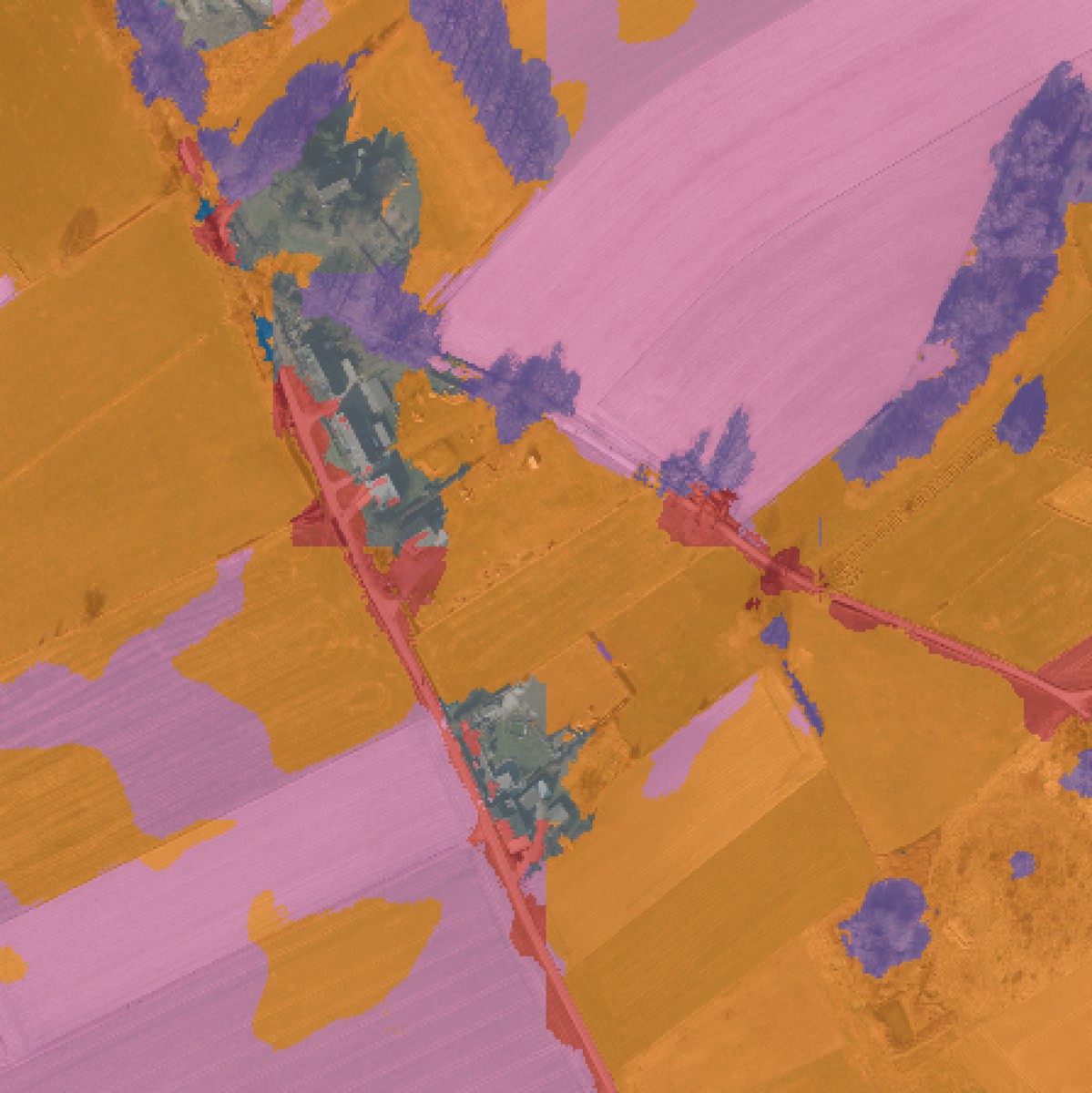}&
        \includegraphics[width=0.16\textwidth]{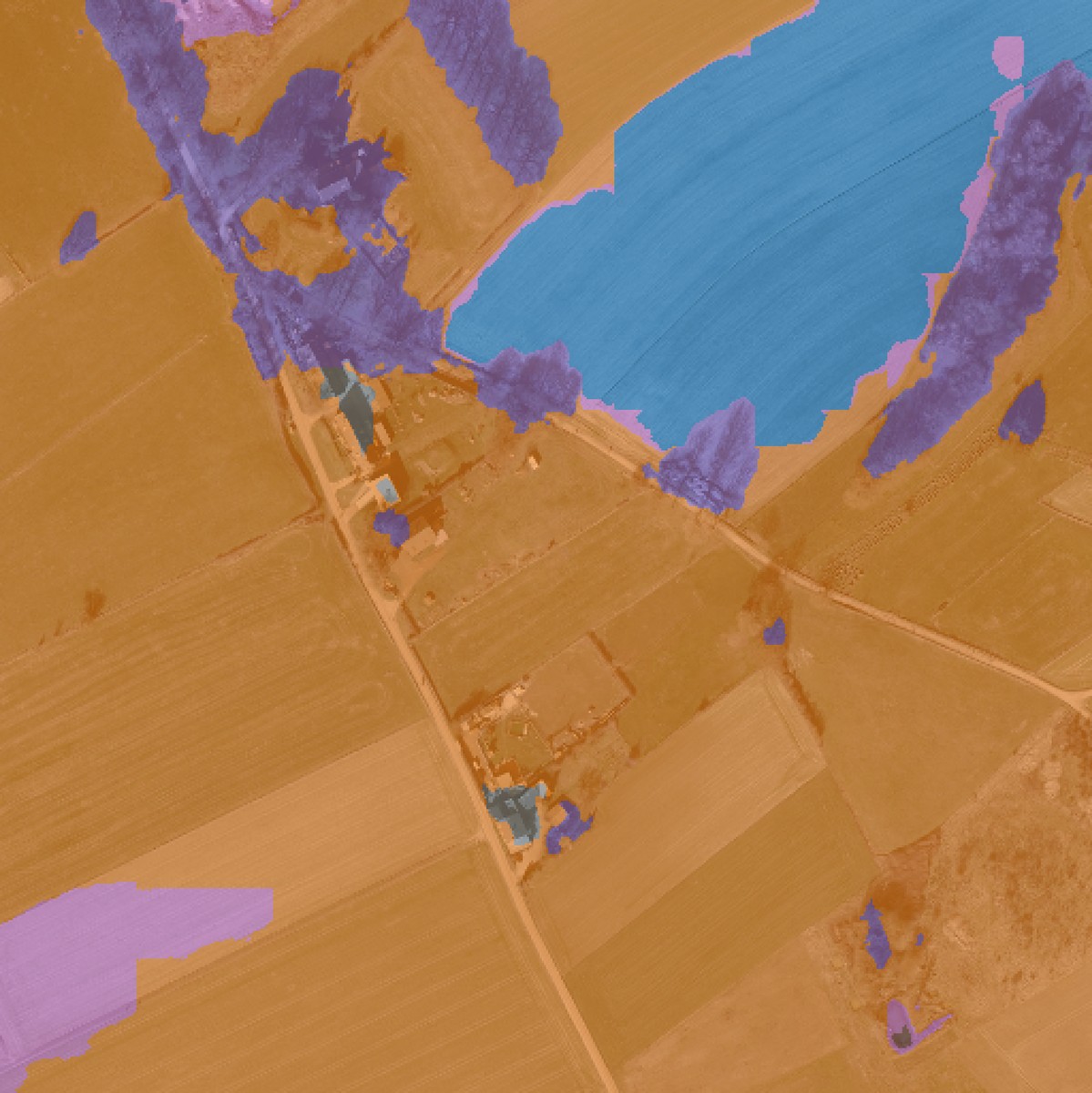}&
        \includegraphics[width=0.16\textwidth]{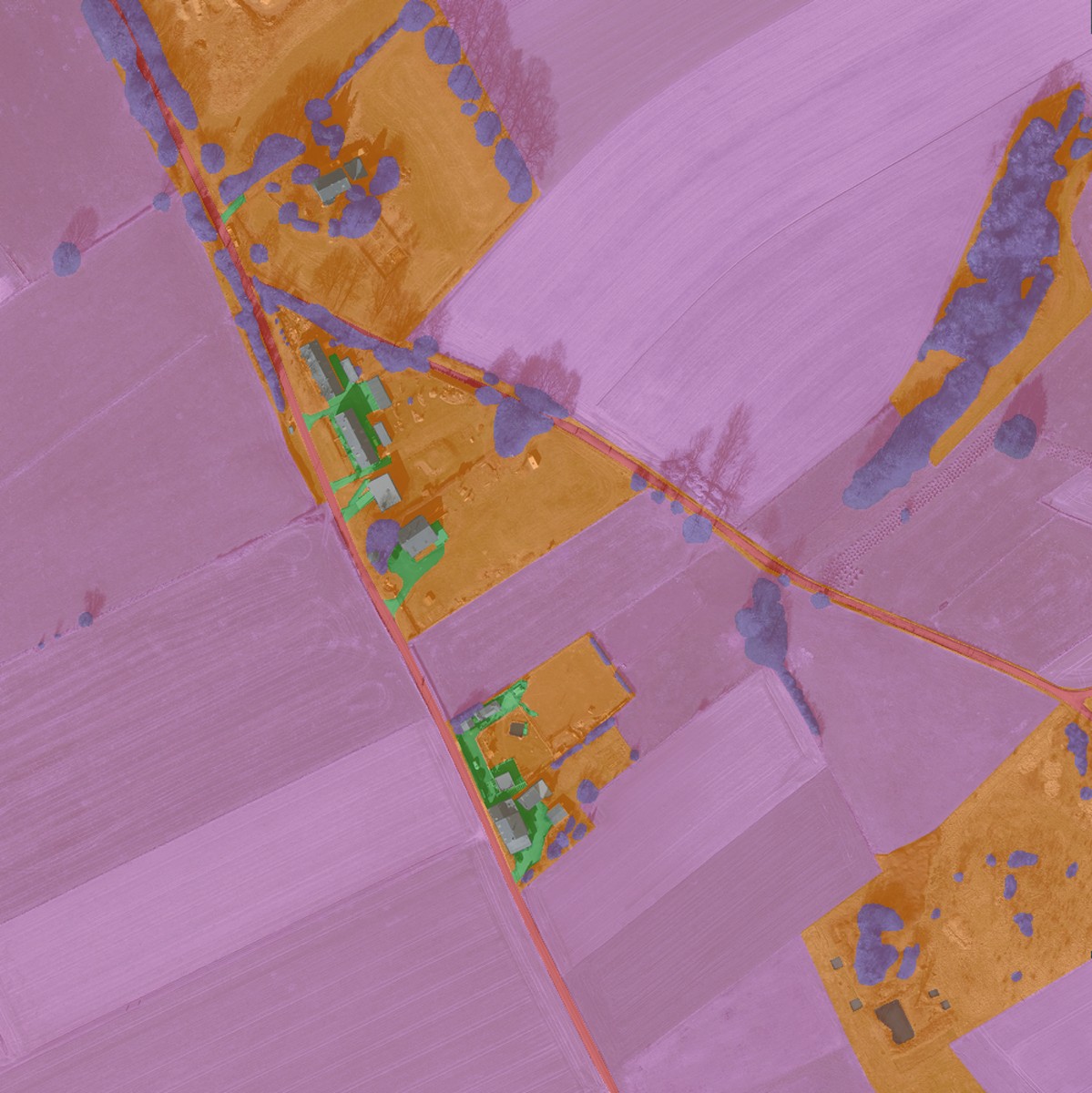} \\[1pt]

        \includegraphics[width=0.16\textwidth]{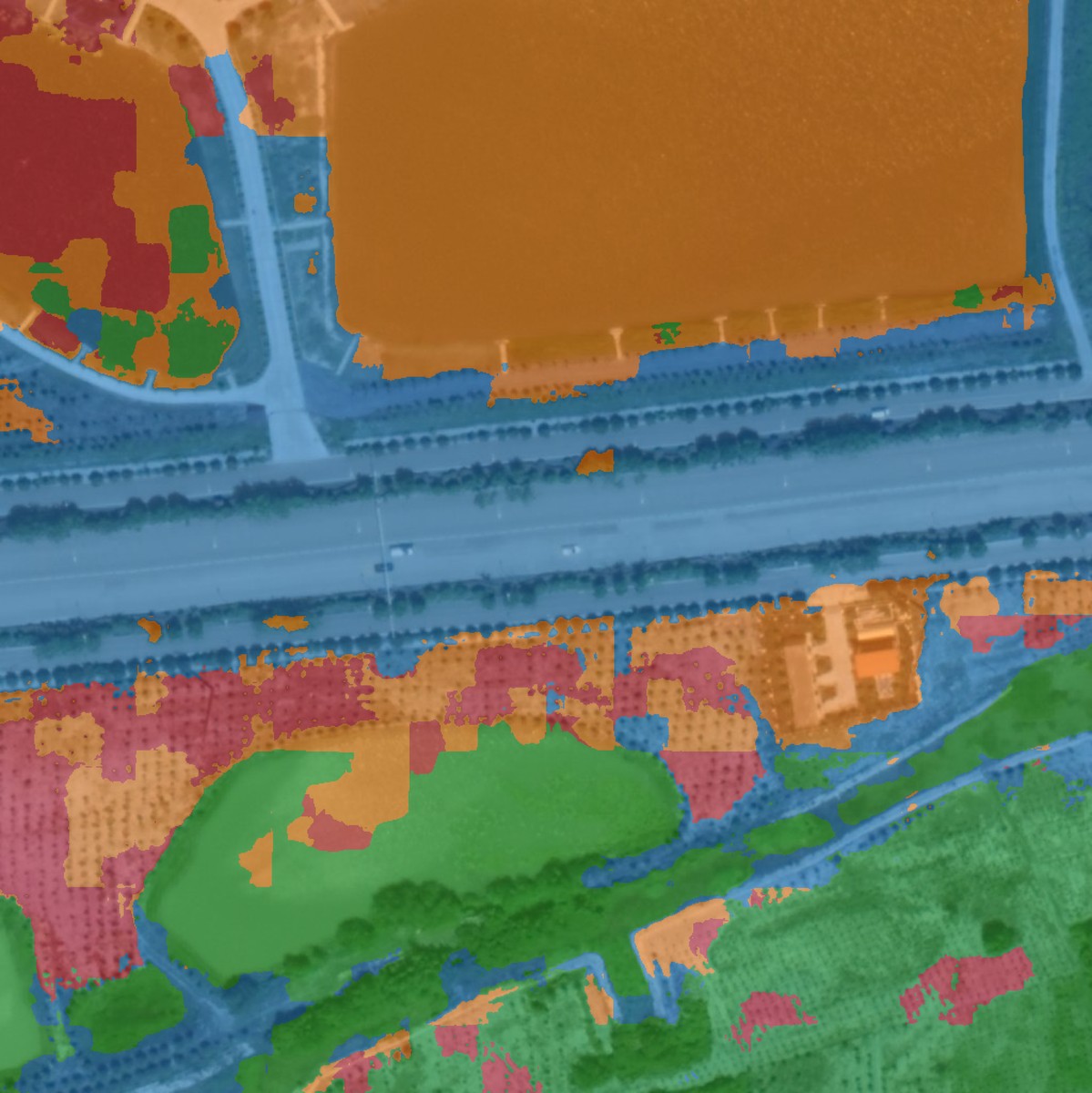} &
        \includegraphics[width=0.16\textwidth]{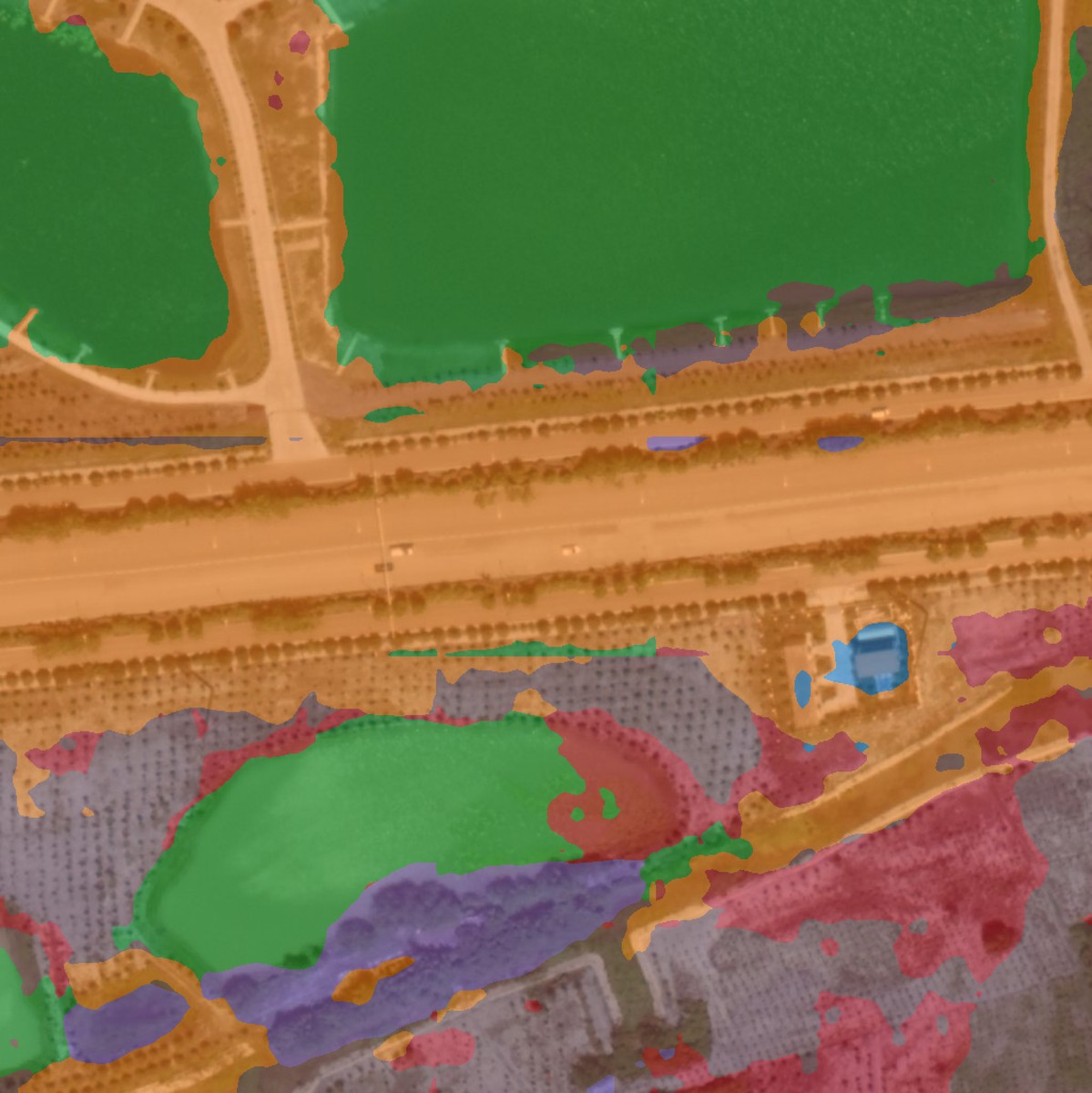}&
        \includegraphics[width=0.16\textwidth]{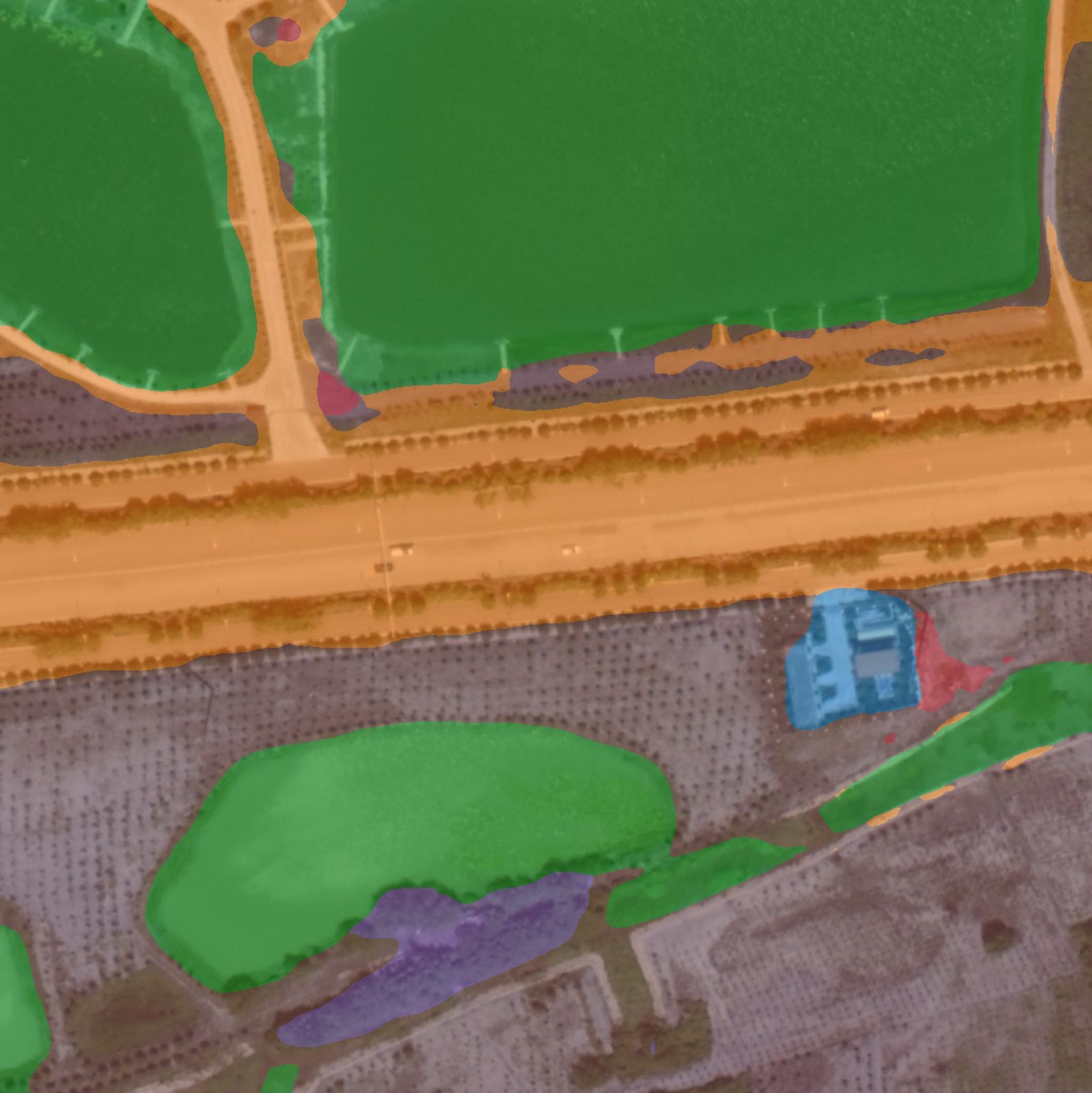}&
        \includegraphics[width=0.16\textwidth]{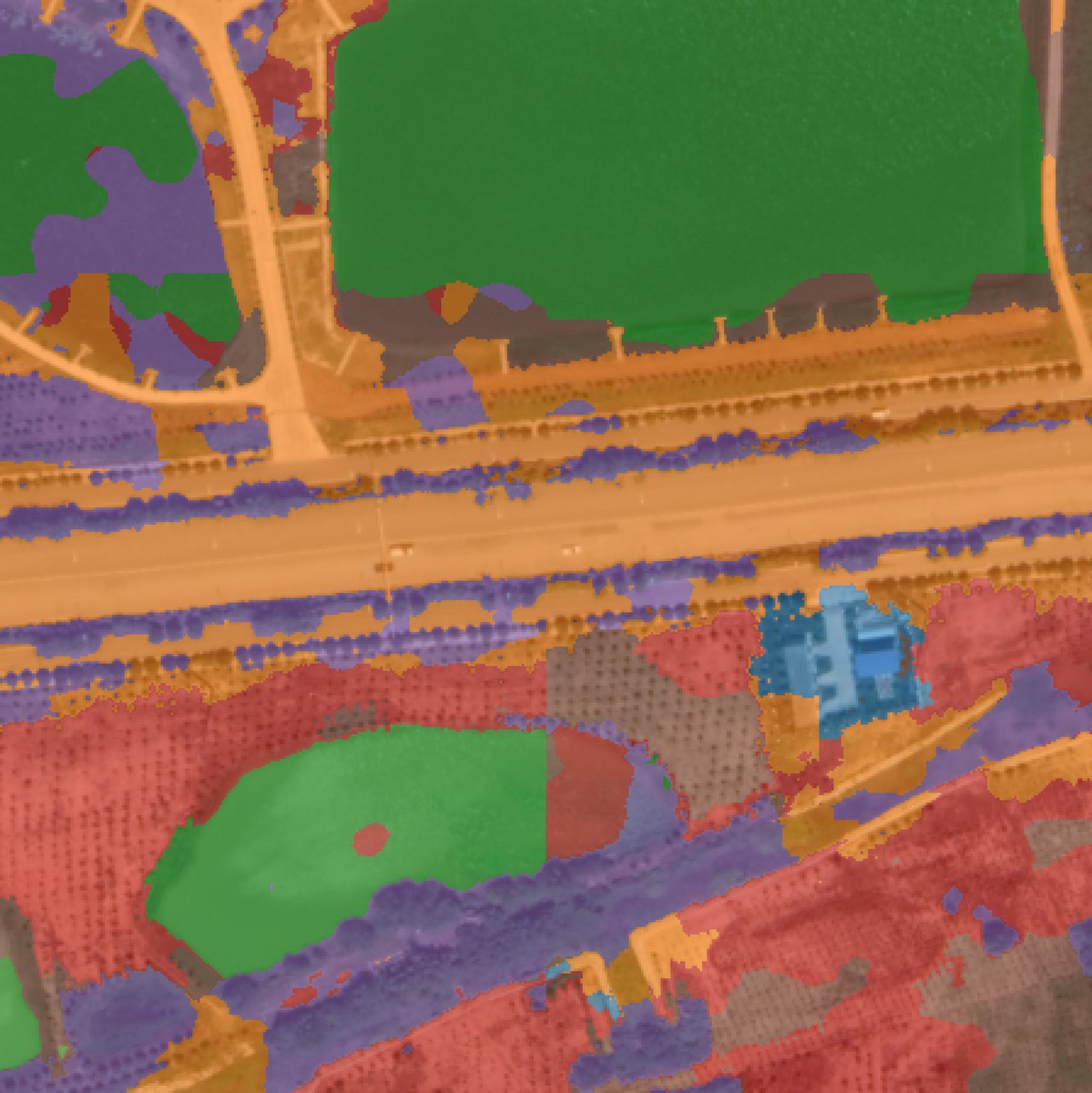}&
        \includegraphics[width=0.16\textwidth]{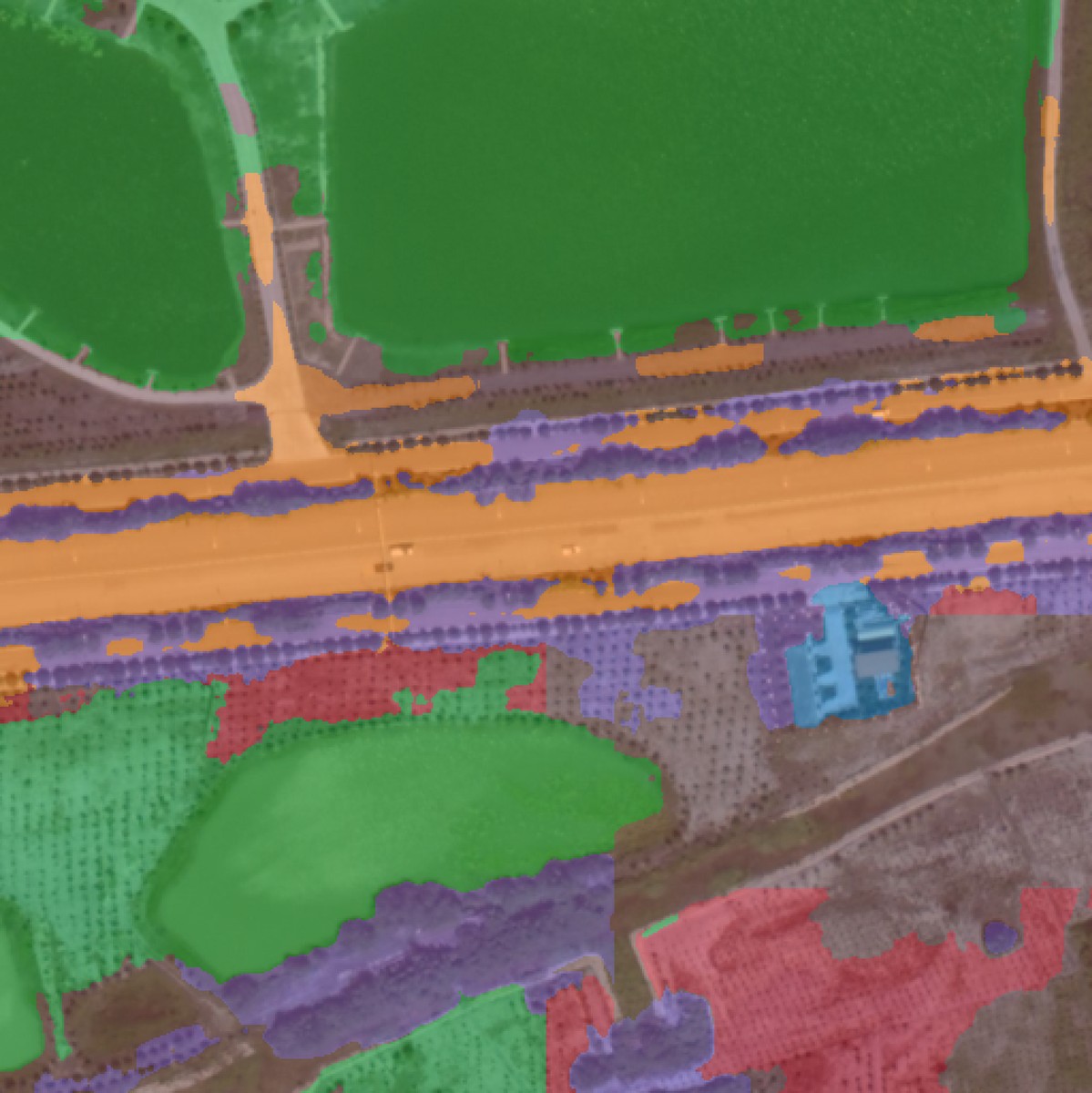}&
        \includegraphics[width=0.16\textwidth]{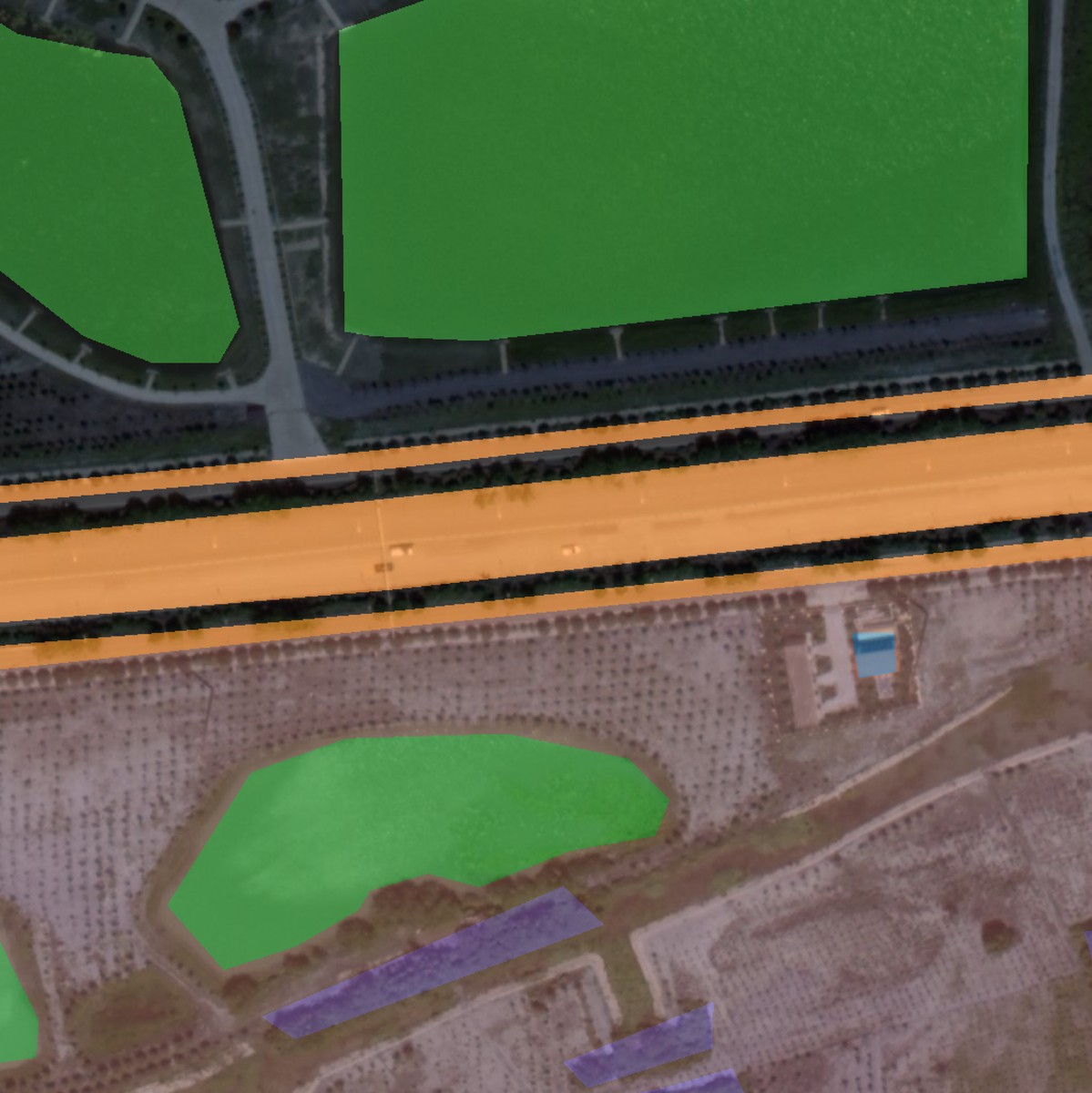} \\[1pt]

    \end{tabular}

    \caption{Quantitative results for CAFe-DINO and other OVSS methods. Our method is remarkably accurate on urban scenes, but sometimes struggles to distinguish fine-grained textures in rural scenes.}
    \label{fig:quant}
\end{figure*}

A graphical overview of our method is shown in \cref{fig:block1}. In short, a cost volume from DINOv3.txt is cost-aggregated and upsampled to form pixel-wise probabilities for each class. The argmax of this volume across the class dimension forms a segmentation prediction.

\subsection{Preliminaries}
\subsubsection{DINOv3.txt}

DINOv3.txt is composed of a DINOv3 vision backbone $\phi_V(\cdot)$ and an aligned text encoder $\phi_T(\cdot)$.

Given an input image $I \in \mathbb{R}^{3 \times H \times W}$, the encoder partitions $I$ into a 1-D sequence of $N$ patch tokens plus a learnable global image representation token (a.k.a. \texttt{[CLS]} token) and projects them to a higher dimension $D$: $\phi_V(I) = [\texttt{[CLS]}, f_1, ..., f_N] \in \mathbb{R}^{(N + 1) \times D}$, where $f_p \in \mathbb{R}^D$ is the embedded patch $p$. To provide both global and dense feature representations, the actual encoder output is a concatenation of the \texttt{[CLS]} token and average-pooled patches:
\begin{equation}
    \mathbf{g} = [c; \sigma([f_1, ..., f_N])] \in \mathbb{R}^{2D},
    \label{eq:vision}
\end{equation}
where $\sigma$ denotes average pooling and $;$ denotes concatenation. 

The text encoder functions similarly, encoding an arbitrary text input $q$ into the same fixed dimension: 
\begin{equation}
    T_q = \phi_T(q) \in \mathbb{R}^{2D}.
\end{equation}

A similarity score between an image and semantic class is formed by the cosine similarity of the text embedding for semantic class $t_i$ and the output of \cref{eq:vision} for input image $I$: $\mathbf{g}_I$. For a set of $M$ semantic classes, a cost volume is formed by concatenating the similarity scores for each class $i$:
\begin{equation}
    V = [\text{sim}(\mathbf{g}_I, T_{t_1}); ... ;\text{sim}(\mathbf{g}_I, T_{t_M})] \in \mathbb{R}^{h \times w \times M},
    \label{eq:costvol}
\end{equation}
where $h \times w$ is the dimension of the patch sequence after spatial reconstruction.

DINOv3.txt forms a segmentation prediction for a set of $M$ semantic classes by applying an argmax operation across $V$. CAFe-DINO uses $V$ as the input to the cost aggregation network.

\subsubsection{AnyUp}

Given an input image $I \in \mathbb{R}^{3 \times H \times W}$ and the low-resolution encoded representation of that image from an arbitrary encoder, $p = \phi(I) \in \mathbb{R}^{h \times w \times C}$, the AnyUp model uses $I$ as guidance to upsample $p$ to the image resolution $H\times W$:
\begin{equation}
    q = \mathcal{U}(I, p) \in \mathbb{R}^{H \times W \times C},
\end{equation}
where $\mathcal{U}$ is the AnyUp network and $q$ is the full-resolution feature map.


\subsection{Cost Aggregation Network}

As with other remote sensing works that employ cost aggregation \cite{10962188}, our model closely follows the CAT-Seg \cite{cho2024a} implementation. However, we replace the linear transformer blocks used by CAT-Seg with channel attention blocks to improve representational capacity, and we dispose of the convolutional upsampler, opting instead for feature upsampling, as described in \cref{sec:featup}. We surmise that this choice will mitigate overfitting to the natural image domain during fine-tuning.

The cost aggregation network is displayed graphically in \cref{fig:block2}. We begin by forming a cost volume from an image $I \in \mathbb{R}^{3 \times H \times W}$ and a set of captions for each of $M$ semantic classes of interest $\mathcal{Q} = \{q_1, ..., q_M\}$. The resulting cost volume $V$ is given by equation \cref{eq:costvol}.

The Class-Wise Projection layer projects each cost map to a higher dimension $D_{agg}$ for cost aggregation with a small convolutional net. Weights are shared across cost maps so that the same features are extracted from each:
\begin{equation}
    V_{agg} = \phi_{proj}(V) \in \mathbb{R}^{h \times w \times M \times D_{agg}}.
\end{equation}

The projected cost volume is passed to a set of aggregation blocks. Following CAT-Seg, we use 6 blocks in practice. Each block first aggregates cost slices independently of each other with a Swin Transformer \cite{liu2021} block pair,
\begin{equation}
    V'_{agg} = \phi_{swin}(V_{agg}(:,i)) + V_{agg}(:,i).
\end{equation}
$V_{agg}(:, i)$ indicates the $i_\text{th}$ cost slice of $V_{agg}$.

Next, the spatially aggregated cost volume is input to a channel attention block. This block computes self-attention across classes on a per-pixel basis, independent of spatial context.
\begin{equation}
    V''_{agg} = \phi_{chan}(V'_{agg}(:, j, k)) + V'_{agg}(:, j, k).
\end{equation}
$V'_{agg}(:, j, k)$ indicates pixel $(j, k)$ of the cost volume.

Residual connections between all aggregation block modules are added to encourage training stability with increasing depth.

\subsection{Class-Wise Upsampling}
\label{sec:classup}
\label{sec:featup}

Given a projected, aggregated cost volume with $M$ classes, $V''_{agg} \in \mathbb{R}{h \times w \times M \times D_{agg}}$, we apply an upsampling and channel-reduction procedure for each cost map $V''{agg}(:, i)$, $i \in [1, ..., M]$.

The cost map features are upsampled to the resolution of the original image $I$ with AnyUp:
\begin{equation}
    V_{up}(i) = \mathcal{U}(I, V''_{agg}(:, i)) \in \mathbb{R}^{H \times W \times D_{agg}}.
\end{equation}
We then apply a network of $1 \times 1$ convolutional layers to reduce the upsampled cost map $V_{up}$ from $D_{agg}$ channels to a single channel of pixel-wise probabilities. As with projection, the same convolutional kernels are applied to each cost slice for channel reduction.

The upsample/reduce procedure is applied for every cost map of the volume in parallel. The segmentation prediction $Y$ is formed by taking the argmax of the upsampled and reduced cost maps:
\begin{equation}
    Y = \text{argmax}[\phi_{red}(V_{up}(1)); ...; \phi_{red}(V_{up}(M))]
\end{equation}
$\phi_{red}(\cdot)$ denotes channel reduction, and $;$ denotes concatenation.

A key benefit of the feature upsampler is that it does not require fine-tuning, so it will not overfit to the natural image domain during training.
\section{Experiments}
\label{sec:experiments}

\begin{figure*}
    \centering
    \setlength{\tabcolsep}{1.5pt}

    \begin{tabular}{c*{7}{c}}
        & Road & Building & Low Veg. & Tree & Car  & Prediction \\
        
        \raisebox{26pt}{DINOv3.txt} &
        \includegraphics[width=0.14\textwidth]{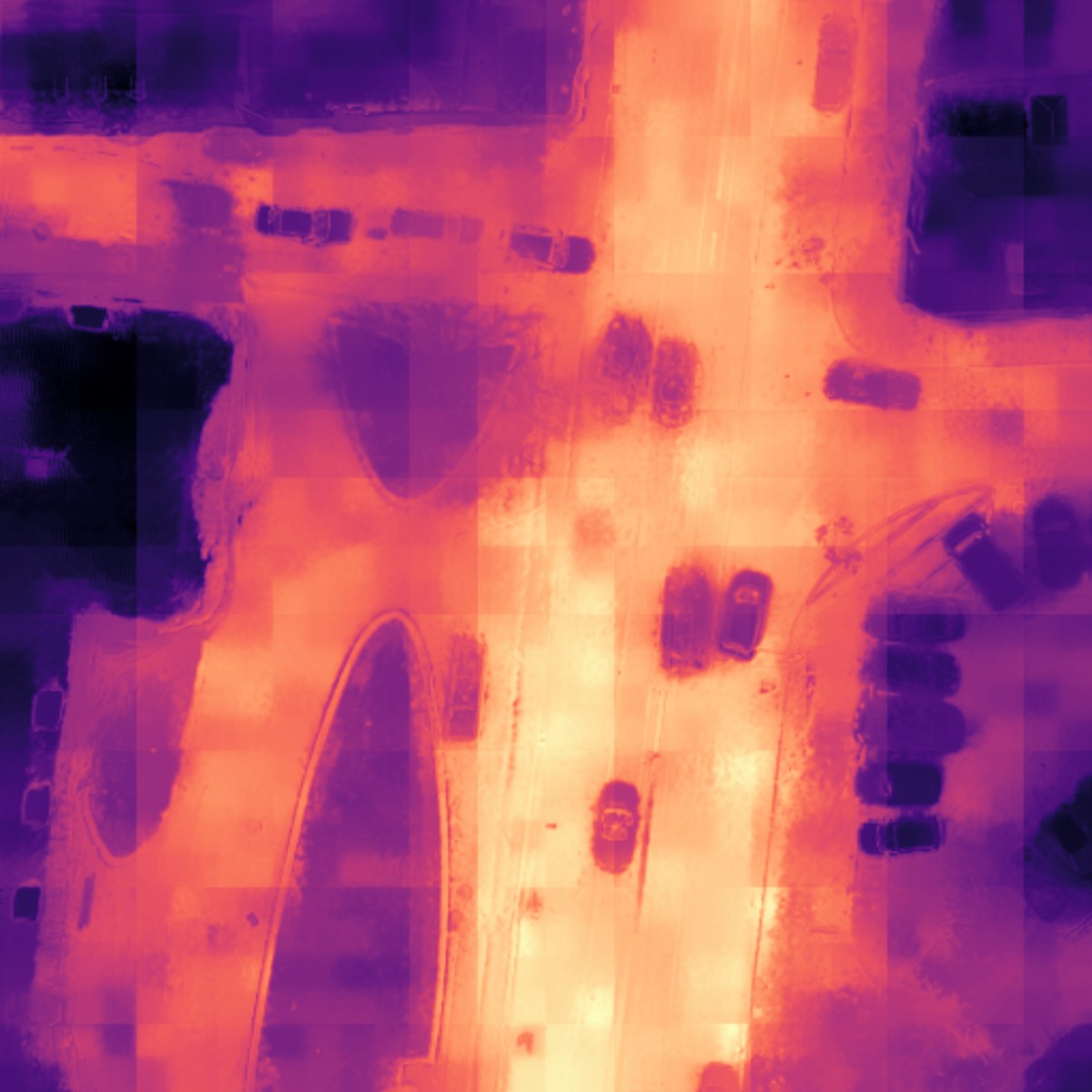} &
        \includegraphics[width=0.14\textwidth]{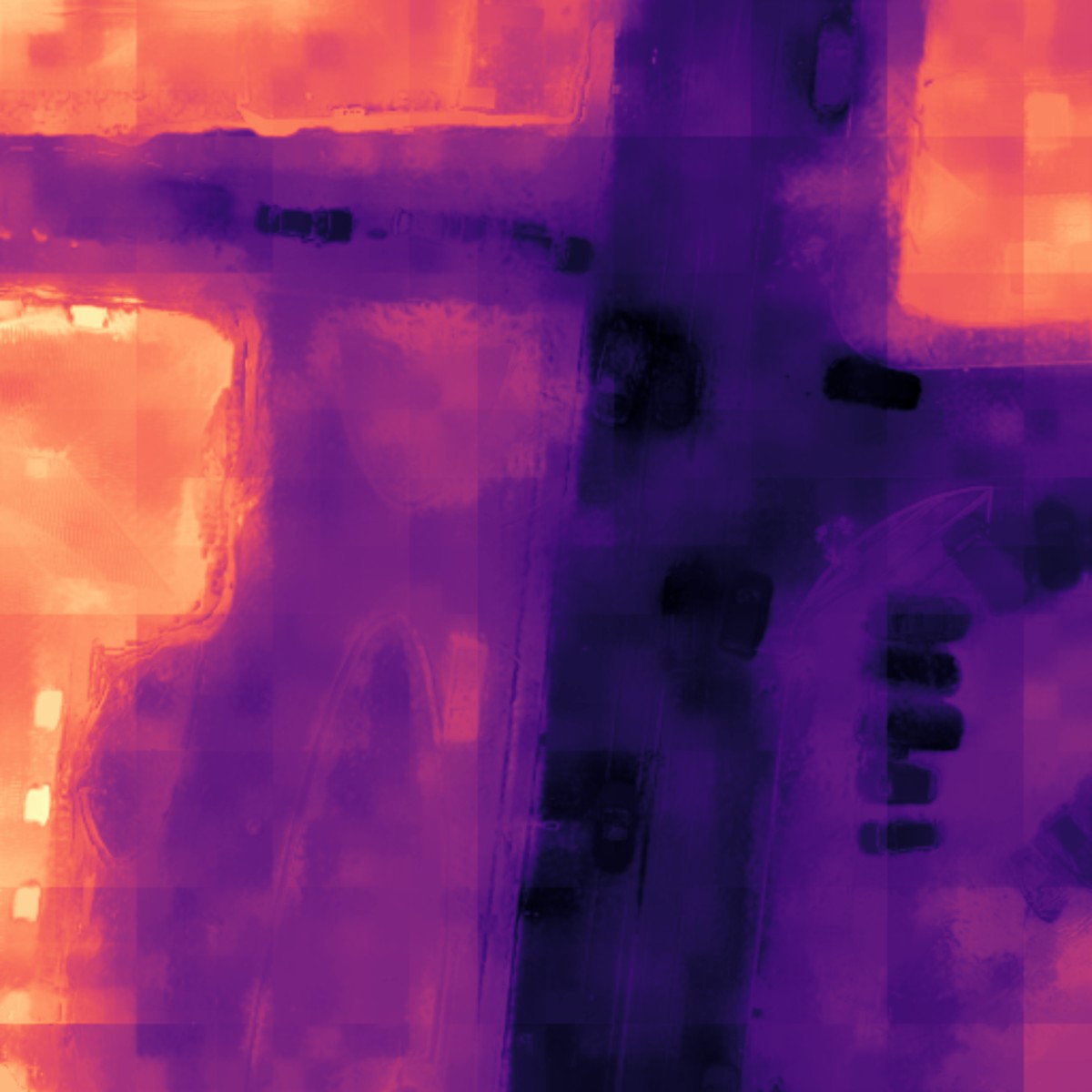}&
        \includegraphics[width=0.14\textwidth]{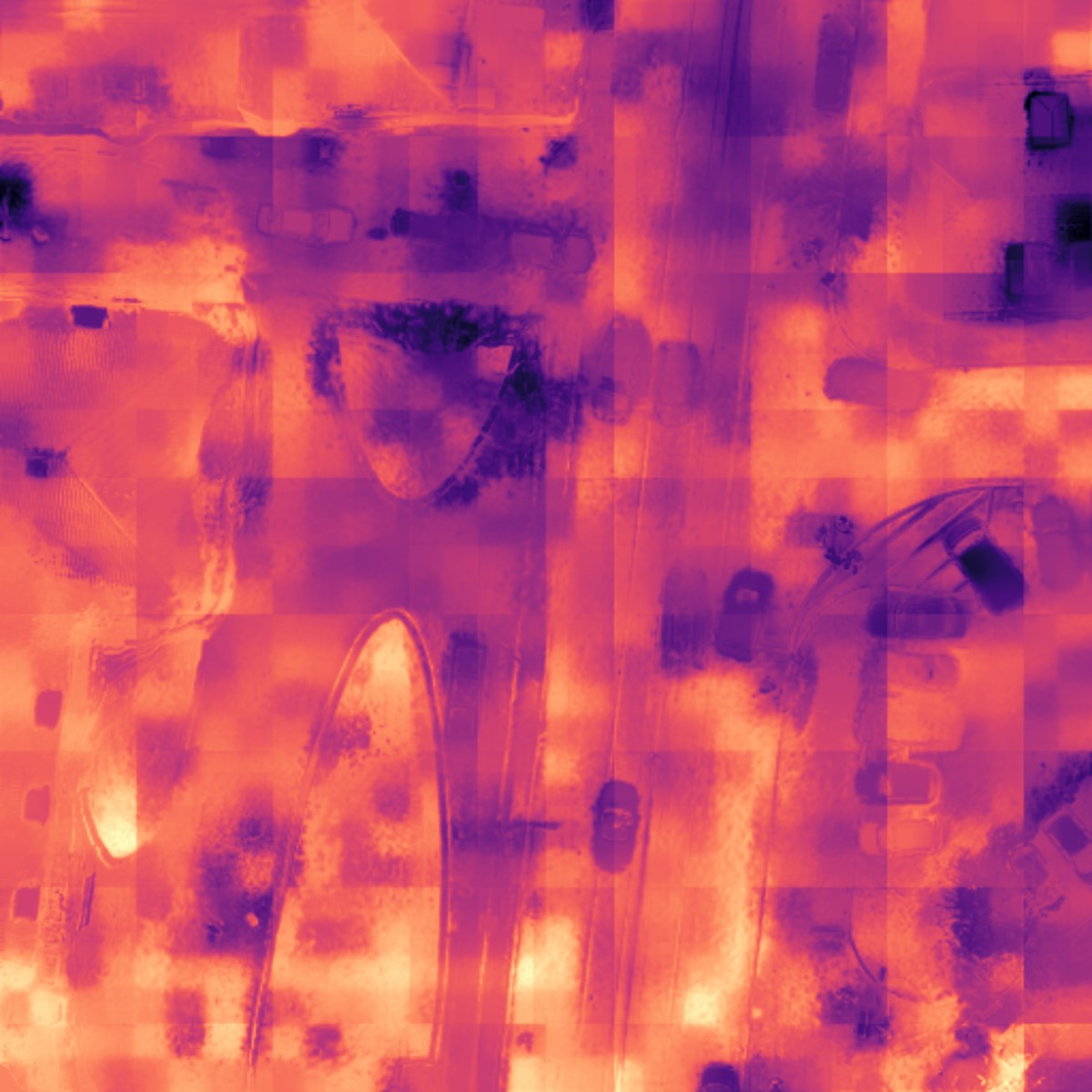}&
        \includegraphics[width=0.14\textwidth]{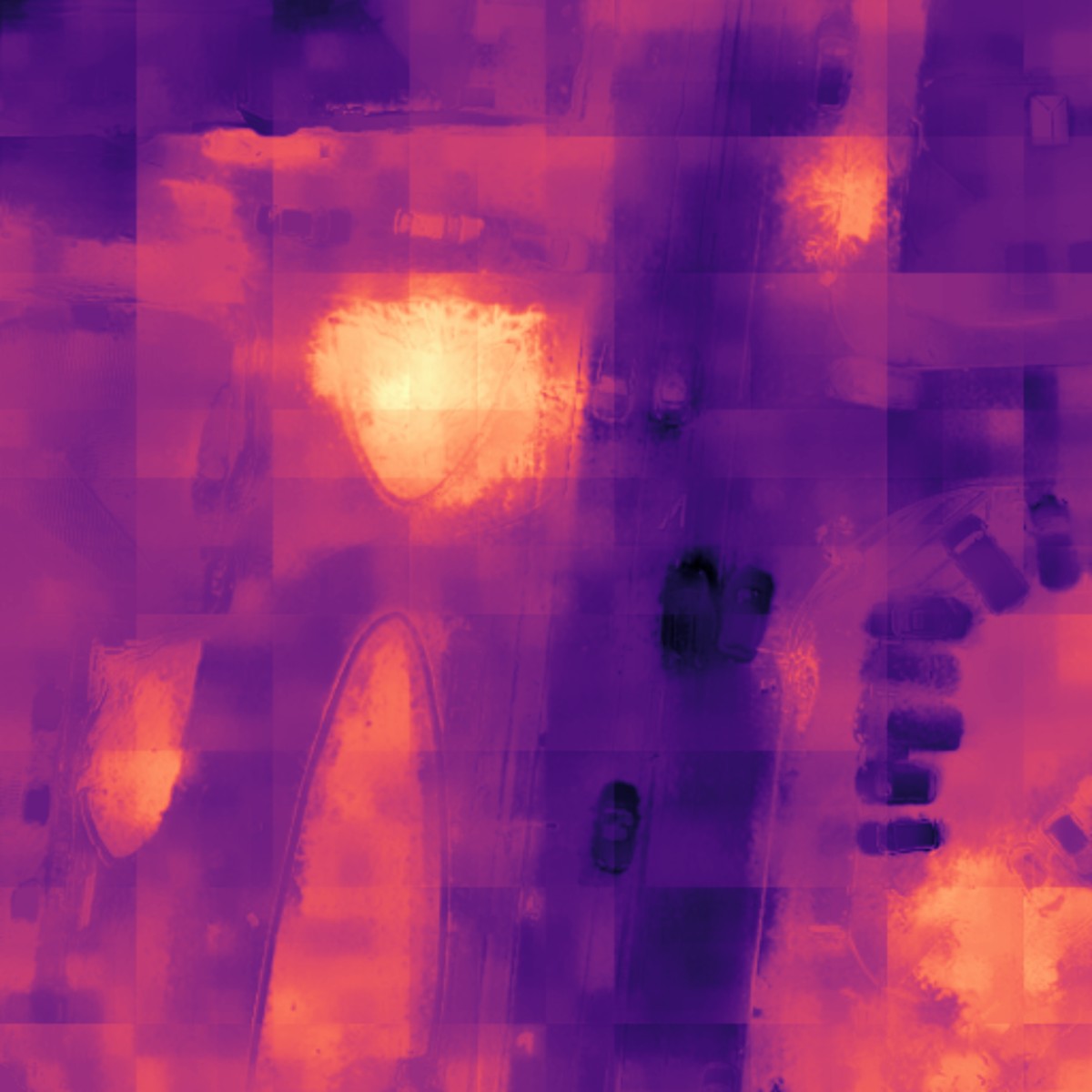}&
        \includegraphics[width=0.14\textwidth]{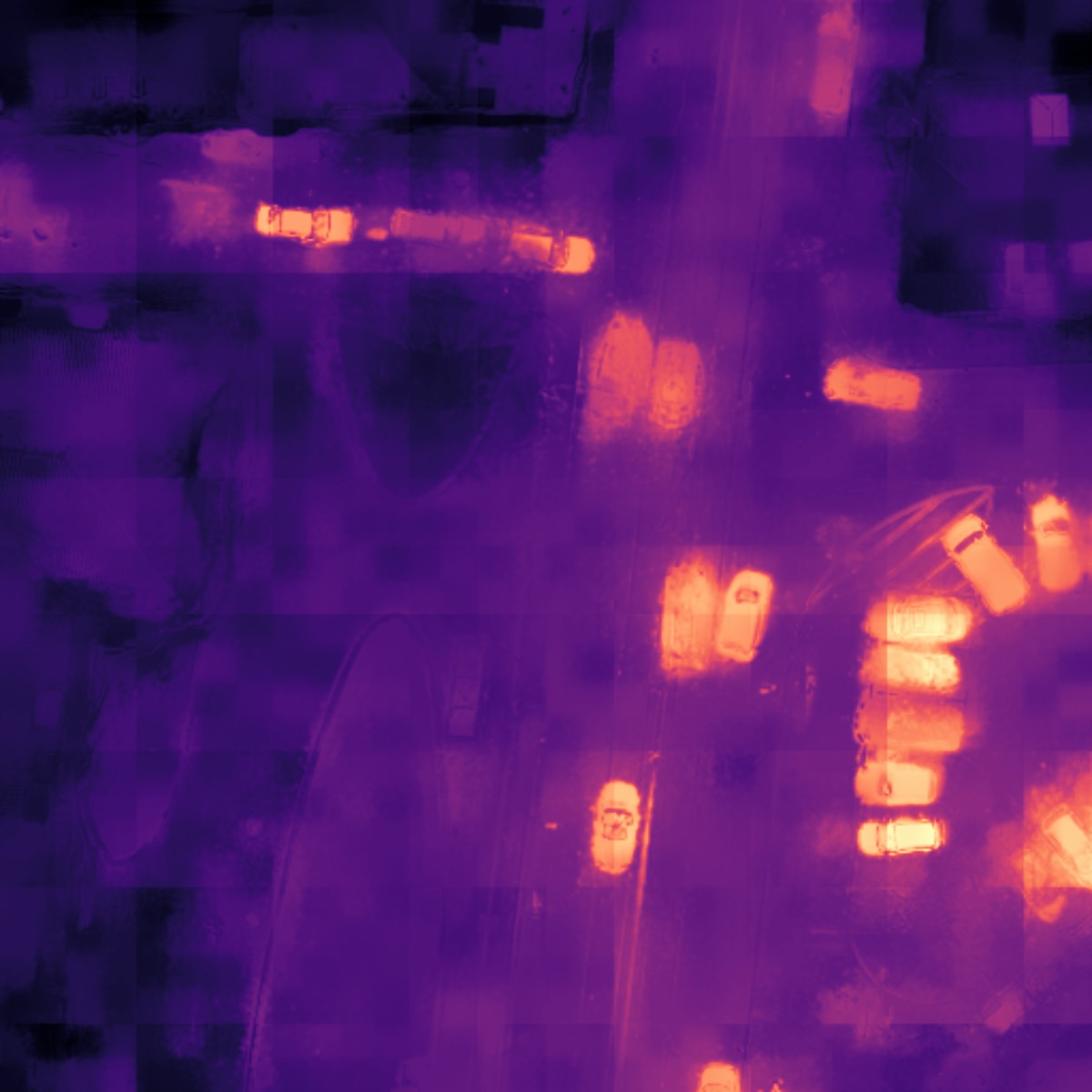}&
        \includegraphics[width=0.14\textwidth]{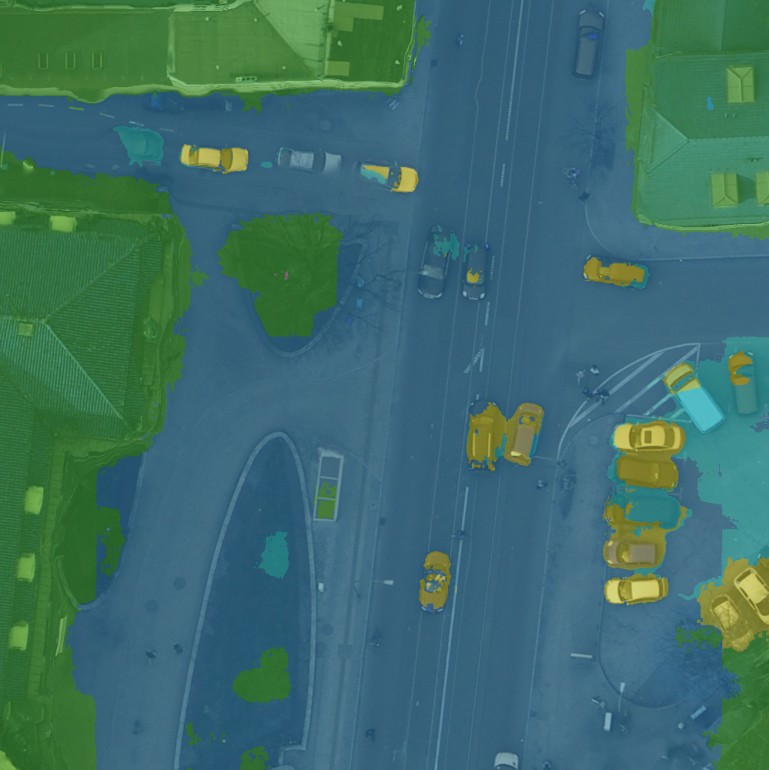} \\[1pt]

        \raisebox{26pt}{CAFe-DINO} &
        \includegraphics[width=0.14\textwidth]{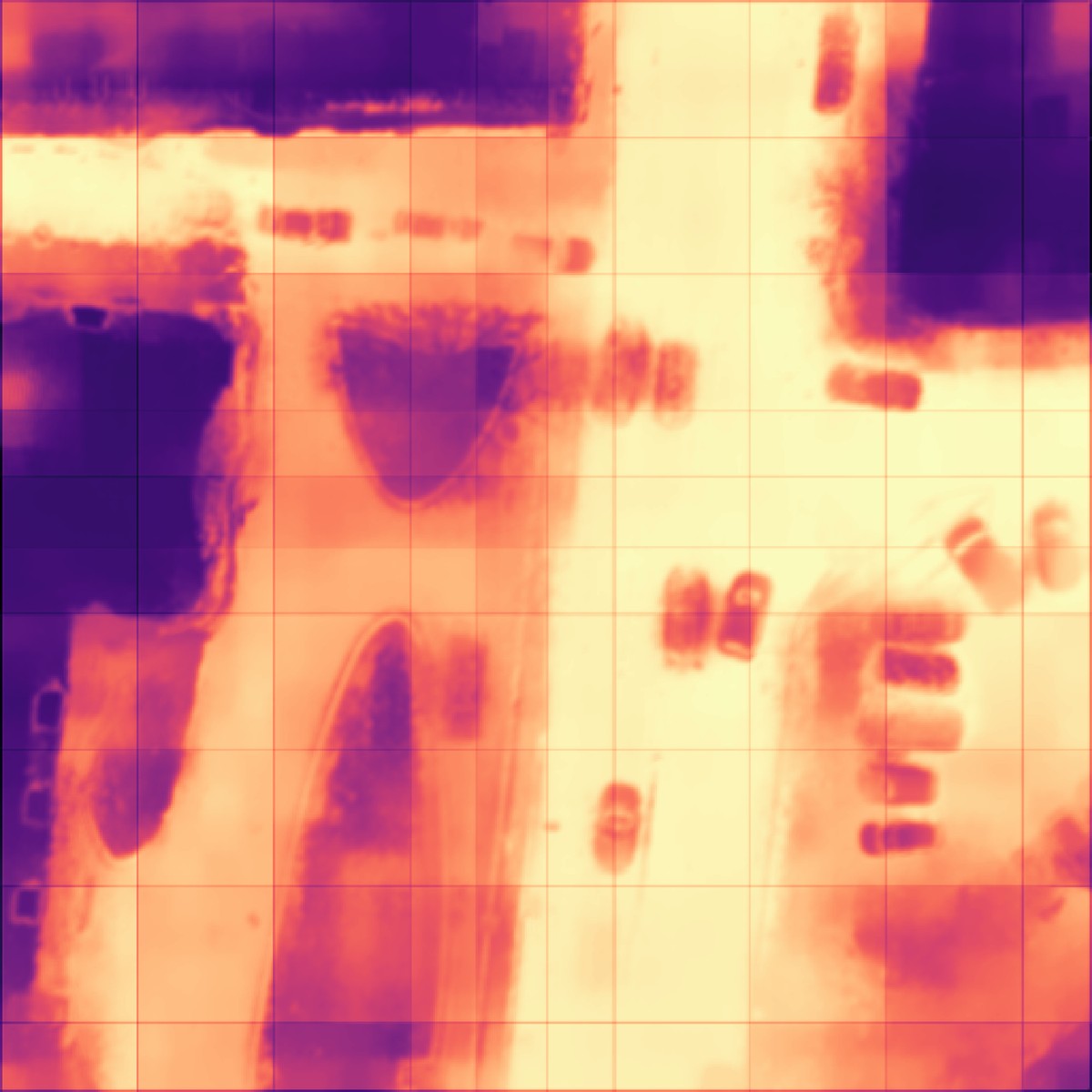} &
        \includegraphics[width=0.14\textwidth]{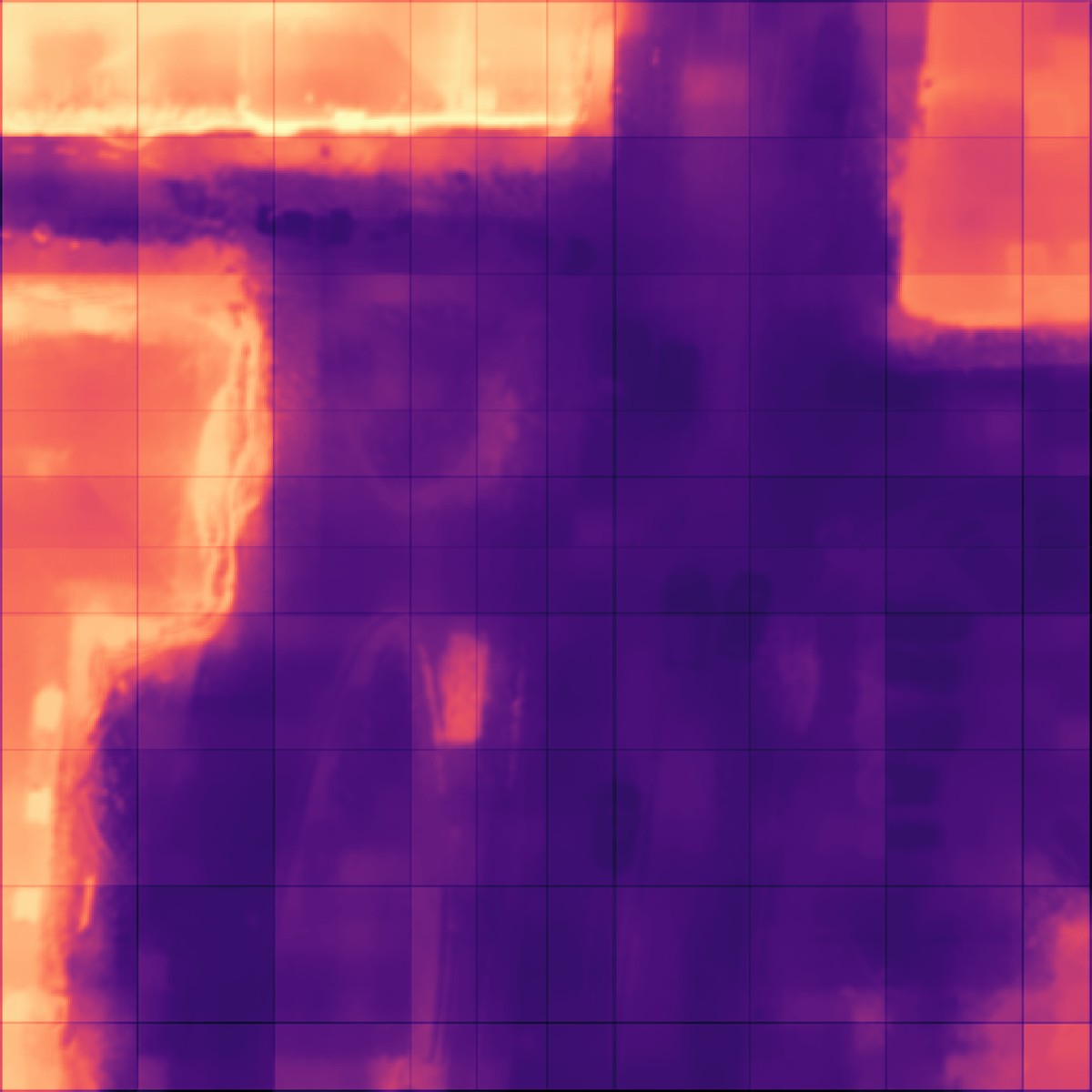}&
        \includegraphics[width=0.14\textwidth]{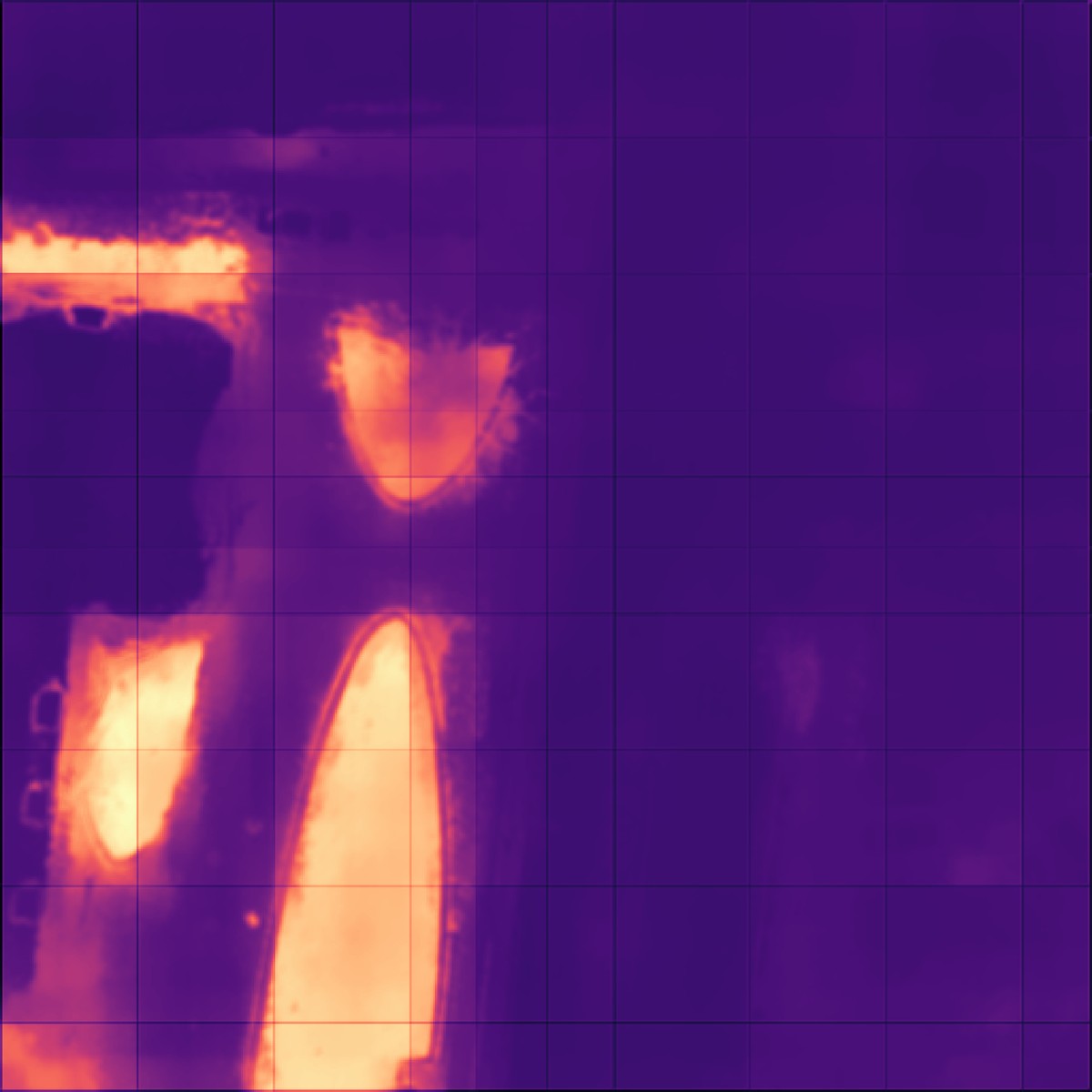}&
        \includegraphics[width=0.14\textwidth]{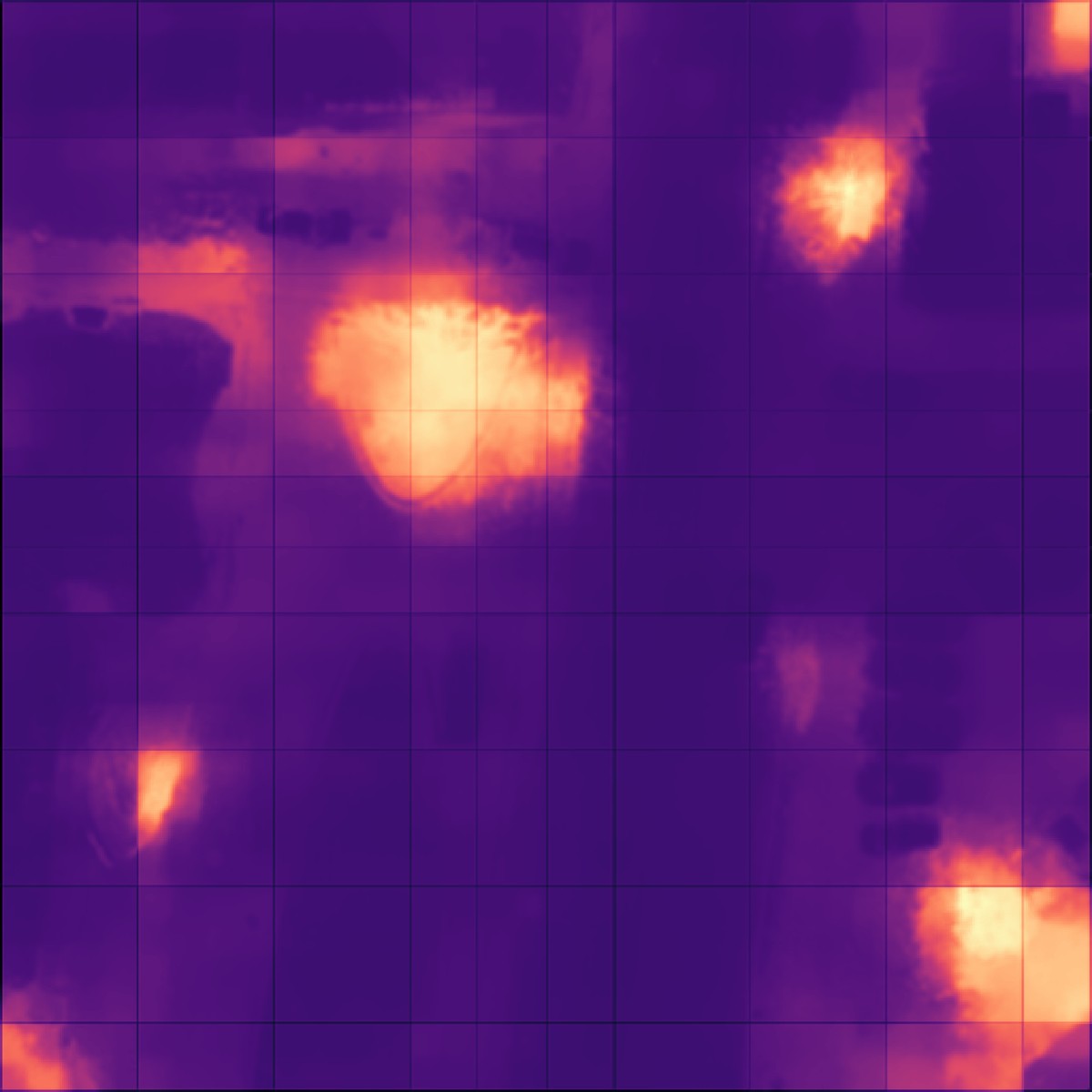}&
        \includegraphics[width=0.14\textwidth]{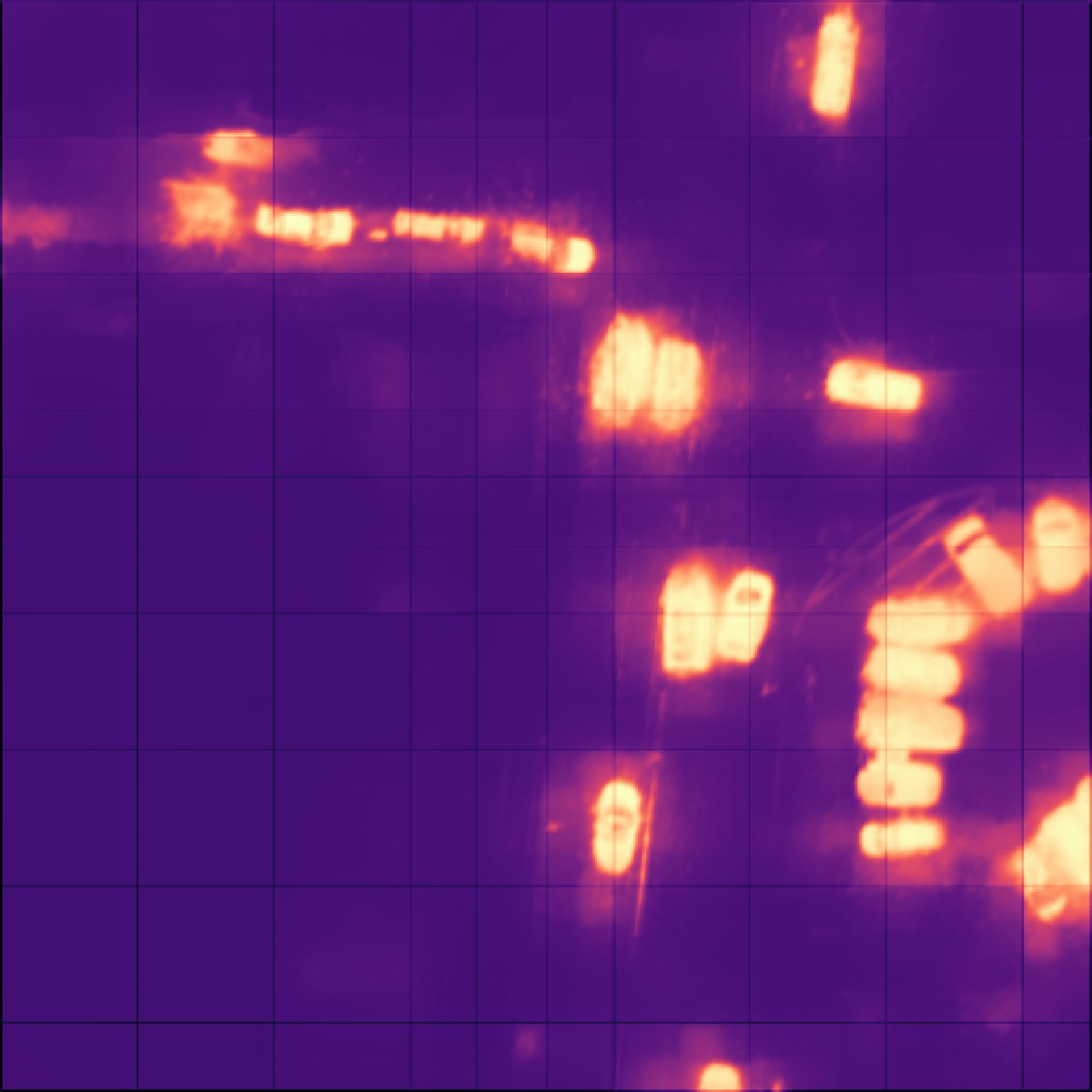}&
        \includegraphics[width=0.14\textwidth]{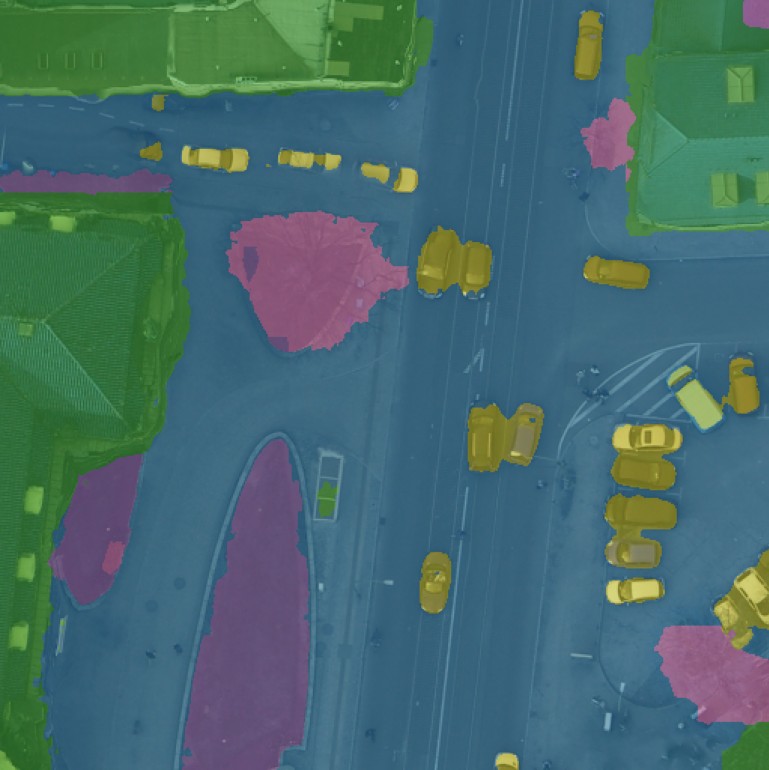} \\
    \end{tabular}

    \caption{DINOv3.txt cost maps for each of the Potsdam classes before (top row) and after (bottom row) CAFe-DINO aggregation. The segmentation prediction is an argmax over all cost maps. Cost maps for each dataset are shown in Sec. 7 of the supplementary material.}
    \label{fig:grid}
\end{figure*}

\begin{table*}
\centering
\caption{mIoU scores of our RS-targeted subset of COCO-Stuff vs. 5 random samples. Scores are reported in mIoU (\%)}
\label{tab:coco}
\begin{tabular}{c|cccc}
\toprule
Method                  &Potsdam&Vaihingen&OEM&LoveDA \\
\midrule
RS-targeted Subset              &\textbf{65.5} & \textbf{56.5} & \textbf{38.8} & \textbf{59.3}\\
Random Subsets        & 4.8 (1.4) & 4.3 (0.9) & 3.8 (1.9) & 3.5 (2.5) \\
\bottomrule
\end{tabular}
\end{table*}

\begin{table*}
\centering
\caption{Comparison of OVSS methods across multiple datasets with mIoU (\%). \textsuperscript{\textdagger}Method was trained on this dataset.}
\label{tab:results}
\begin{tabular}{cc|c|cccc|c}
\toprule
Method &&
RS Training&Potsdam&Vaihingen&OEM&LoveDA&Average\\
\midrule
DINOv3.txt\cite{simeoni2025dinov3} & ~\textsubscript{\scriptsize FAIR'25}
    & No & 27.0 & 20.1 & 24.2 & 43.7 & 28.8 \\
OVRS\cite{10962188} & ~\textsubscript{\scriptsize TGRS'25}
    & Yes & 28.6 & 22.0 & 21.7 & 47.6 & 30.0 \\
GSNet\cite{ye2025} & ~\textsubscript{\scriptsize AAAI'25}
    & Yes & 43.2 & 34.1 & 36.8 & \textcolor{gray}{78.2}\textsuperscript{\textdagger} & 38.0 \\
SegEarth-OV\cite{li2025segearthov} & ~\textsubscript{\scriptsize CVPR'25}
    & Self-Supervised & 52.0 & 27.2 & 35.6 & 53.5 & 48.0  \\
    \midrule
CAFe-DINO & ~\textsubscript{\scriptsize Ours}
    & No & \textbf{66.8} & \textbf{54.4} & \textbf{39.6} & \textbf{65.3} & \textbf{56.5} \\
\bottomrule
\end{tabular}
\end{table*}

\begin{table*}
\centering
\caption{The same evaluation as \cref{tab:results}, but with the background class included. \textsuperscript{\textdagger}Method was trained on this dataset.}
\label{tab:results_wbackground}
\begin{tabular}{cc|c|cccc|c}
\toprule
Method &&
RS Training&Potsdam&Vaihingen&OEM&LoveDA&Average\\
\midrule
DINOv3.txt\cite{simeoni2025dinov3} & ~\textsubscript{\scriptsize FAIR'25}
    & No & 23.8 & 17.3 & 21.1 & 26.3 & 22.1 \\
OVRS\cite{10962188} & ~\textsubscript{\scriptsize TGRS'25}
    & Yes & 17.9 & 19.0 & 9.6 & 10.2 & 14.2 \\
GSNet\cite{ye2025} & ~\textsubscript{\scriptsize AAAI'25}
    & Yes & 34.9 & 28.2 & 31.6 & \textcolor{gray}{43.8}\textsuperscript{\textdagger} & 31.6 \\
SegEarth-OV\cite{li2025segearthov} & ~\textsubscript{\scriptsize CVPR'25}
    & Self-Supervised & 46.2 & 24.3 & \textbf{39.8} & 36.9 & 36.8 \\
    \midrule
CAFe-DINO & ~\textsubscript{\scriptsize Ours}
    & No & \textbf{56.4} & \textbf{47.1} & 35.5 & \textbf{42.5} & \textbf{45.4} \\
\bottomrule
\end{tabular}
\end{table*}

\begin{table*}
\centering
\caption{The effect of freezing parts of CAFe-DINO during training on multiple datasets. Scores are reported in mIoU (\%)}
\label{tab:freezing}
\begin{tabular}{c|cccc|c}
\toprule
Method                  &Potsdam&Vaihingen&OEM&LoveDA& Average\\
\midrule

Vision Blocks Trainable       & \textbf{66.8} & 54.4 & \textbf{39.6} & \textbf{65.3} & \textbf{56.5}  \\
Text Blocks Trainable    & 61.5 & 52.5 & 36.6 & 62.2 & 53.2  \\
Both Trainable              &66.7 & \textbf{56.5} & 37.1 & 64.9 & 56.3 \\
Neither Trainable             & 50.9 & 43.5 & 32.9 & 51.1 & 44.6 \\
\bottomrule
\end{tabular}
\end{table*}

\begin{table*}
\centering
\caption{The effect of reducing the aggregation dimension before upsampling. Providing AnyUp with the full feature dimension of the aggregated cost maps has a pronounced effect on overall performance. Scores are reported in mIoU (\%)}
\label{tab:reduce}
\begin{tabular}{c|cccc}
\toprule
Method                  &Potsdam&Vaihingen&OEM&LoveDA \\
\midrule
Base Model              &\textbf{66.8} & \textbf{54.4} & \textbf{39.6} & \textbf{65.3} \\
Reduce Before Up        & 52.4 & 40.0 & 33.6 & 41.8\\
\bottomrule
\end{tabular}
\end{table*}

\begin{table}
\centering
\caption{Linear attention vs. full attention. Scores are reported in mIoU (\%)}
\label{tab:linear}
\begin{tabular}{c|cccc}
\toprule
Method                  &Pots.&Vaihin.&OEM&LoveDA \\
\midrule
Full Attn.              & \textbf{66.8} & \textbf{54.4} & \textbf{39.6} & \textbf{65.3} \\
Linear Attn.        & 63.9 & 53.8 & 34.9 & 58.6 \\
\bottomrule
\end{tabular}
\end{table}

\subsection{Remote-Sensing-Targeted Natural Image Training}

We train CAFe-DINO on natural imagery with the goal of introducing as many relevant semantic classes as possible, noting that a key drawback of existing RS datasets is a lack of dense labeling with a large variety of classes. Thus, we train CAFe-DINO on the COCO-Stuff \cite{caesar2018cvpr} dataset. We reduce the COCO-Stuff dataset to a subset of 41 classes relevant to RS. Though semantically smaller than datasets used for natural imagery, this dataset is rich in the context of RS, which tends to offer datasets with fewer semantic classes. Labels of other classes are ignored during training; this has the effect of hastening convergence time and keeping the procedure tractable, as GPU memory grows rapidly with increasing cost maps.

The classes that form our subset are listed in Sec. 8 of the supplementary material. We provide additional justification for this choice by comparing our curated subset with 5 randomly sampled subsets of COCO-Stuff, shown in \cref{tab:coco}. Our curated subset outperforms the random trials by a large margin.

\subsection{Implementation Details}

We use the ViT-L variant of the DINOv3.txt model and load it with the pretrained weights provided by the authors. We keep the cost aggregation network fully trainable, and the upsampler fully frozen.

For contrastive training of DINOv3.txt, its authors added two ViT blocks to the base DINOv3 backbone. The original pre-training of DINOv3.txt left only those two blocks and the text encoder trainable. We ablate training different parts of the DINOv3 backbone in our own regime in \cref{tab:freezing}. 

Semantic classes are passed to the text encoder as part of a prompt, e.g. ``A photo of a \{\}'', ``An image of a \{\}''. Following common practice \cite{cho2024a, li2025segearthov}, we ensemble a set of prompts; that is, we pass a single class through the text encoder as part of many prompts and take the average. We use the prompt template set established by \cite{li2025segearthov}, listed in Sec. 6 of the supplementary material. We resize images to $224 \times 224$ for efficient training. For testing, we resize images to $512 \times 512$ and perform sliding window inference with a window size of $224 \times 224$ and a stride of $112$. We train CAFe-DINO on a single NVIDIA Ada 6000 GPU with a batch size of 4 for 45,000 iterations.

\subsection{Results}

\subsubsection{Datasets}

We evaluate CAFe-DINO on 4 multiclass RS datasets: ISPRS Potsdam and Vaihingen \cite{zotero-item-218}, OpenEarthMap \cite{xia2023}, and LoveDA \cite{wang}. These provide a healthy mix of urban vs. rural scenes, various resolutions/ground sample distances, and a collection of both object and land cover classes. For each dataset, we serve the names of the semantic classes as prompts to CAFe-DINO. Sometimes, we alter the given semantic name to improve model recognition (e.g., we will use ``farm'' instead of ``agricultural'' in LoveDA). A full list of our semantic prompts for each dataset is provided in Sec. 6 of the supplementary material. 

\subsubsection{Comparison to State-of-the-Art}

The wave of RS-specific OVSS methods recently published has decisively eclipsed OVSS methods from the natural imagery domain \cite{ye2025, li2025segearthov}, therefore we compare against only RS-specific baselines. For performance comparisons of natural image OVSS methods, refer to the respective results of these baselines.

We additionally compare against vanilla DINOv3.txt in order to demonstrate the efficacy of our cost aggregation network in unlocking the potential of DINOv3 on RS imagery.

Our results are shown in \cref{tab:results}. CAFe-DINO outperforms the existing state-of-the-art on all evaluated datasets without any RS training, supervised or unsupervised. Our largest margins of success are on the Potsdam and Vaihingen datasets. This is not surprising, as the semantic classes of Potsdam/Vaihingen are common in natural imagery, and it is a relatively high-resolution dataset.

The lower mIoU score and slimmer margin on OEM suggest that CAFe-DINO is weaker on rural scenes than urban. This is corroborated by row 3 of \cref{fig:quant}; we see that CAFe-DINO has mis-segmented a crop as a barren patch in one region and a field of grass in another. Interestingly, a similar mistake was made by SegEarth-OV, which was not fine-tuned on RS data (only self-supervised), that was avoided by the other two RS-trained baselines. This suggests an inherent limitation of natural-imagery-trained models for distinguishing specific textures in RS land cover, as there is no such distinction to be learned in natural imagery.

\subsubsection{Comparison to DINOv3}

DINOv3.txt falls short of its natural imagery performance when applied to RS. We provide an explanatory diagram of the strengths and weaknesses of DINOv3.txt alone on RS data, and how our method improves it in \cref{fig:grid}. This figure shows raw similarity maps of DINOv3.txt for each class against the aggregated versions produced by CAFe-DINO in an urban scene. Familiar forms from natural imagery are well-activated, but the ``Low Vegetation'' class is not. In spite of this, the cost aggregation network recovers a legitimate segmentation prediction. It is noteworthy that Low Vegetation was not a part of the text corpus for fine-tuning the cost aggregator, meaning that the cost aggregation network produced this probability score via cross-class reasoning against more well-formed similarity maps.

\subsubsection{Inclusion of the Background Class}

Prior art in RS OVSS varies on the evaluation choice of whether to include the background/clutter/unlabeled classes of certain datasets as a semantic class, or ignoring it. The background class is relevant for certain evaluation scenarios, but because the background is an ill-defined category whose semantic meaning varies across datasets, we argue that ignoring the background is a better way to measure OVSS methods against one another. We note in particular the large proportion of background labels (>20\%) in the LoveDA dataset, which could skew results wildly between methods.

For completeness, we provide an evaluation that includes backgrounds in \cref{tab:results_wbackground}, though we do not consider these our primary results. Performance is generally worse across all datasets and methods given the increased difficulty of background segmentation. Our method is still dominant in all domains except OEM under this regime.

\subsection{Ablations}

\subsubsection{Fine-tuning DINOv3}

In the proposed architecture, the cost aggregation network is fine-tuned and the upsampling network is necessarily left frozen, but there are some options available for partially fine-tuning the DINOv3 backbone.

We ablate fine-tuning the DINOv3 text encoder and fine-tuning the final 2 blocks of the DINOv3 vision encoder. The last 2 blocks are unique, as they were added post-hoc to the vision backbone and fine-tuned with the text encoder on image-text tasks in the original DINOv3 work.

Results are shown in \cref{tab:freezing}. Incorporating fine-tuning of either DINOv3 backbone component (or both at once) is necessary for competitive performance, yielding, in the worst case, an 11.9\% mIoU gain compared to freezing the backbone completely. Since cost maps are generated from the similarity score between embeddings of both the vision and text encoders, fine-tuning either component can influence the initial cost maps in a similar (though distinct) way. It follows intuition that fine-tuning the vision blocks is more effective, as the domain gap between natural and RS imagery is great, whereas most text labels used in RS will have been well-represented in DINOv3's pretraining. 

CAFe-DINO reduces the aggregated cost map channel dimension after upsampling (see \cref{sec:classup}). We explore a more runtime-efficient approach, in which the aggregator feature dimension is reduced to a single channel before upsampling. As a result, the prediction maps are being directly upsampled, rather than projected cost slices. Results are shown in \cref{tab:reduce}. Retaining the deep aggregator feature dimension is critical for good performance with AnyUp; we note that AnyUp was pre-trained on embedded features with high channel counts, not single channel inputs.

\section{Conclusion}
\label{sec:conclusion}

In this work, we introduce CAFe-DINO, an OVSS method for RS imagery that eschews fine-tuning on geospatial data. We have demonstrated strong OVSS improvements with large margins on several datasets. We look forward to additional research leveraging the representational power of DINOv3 for RS tasks. 

\subsection{Limitations}
While CAFe-DINO achieves strong OVSS performance without RS supervision, several limitations remain:

\subsubsection*{Scaling with Increasing Classes}
Cost aggregation uses a cost volume whose memory usage grows linearly with the number of semantic classes. As the cost volume grows, memory increases linearly, which may limit applicability in large open-vocabulary settings. Runtime is not as badly impacted due to parallelization of cost aggregation operations.

\subsubsection*{Sensitivity to Pretraining Classes}
We employ an RS-focused subset of COCO-Stuff for training, and show that it is necessary for good performance (vs. random subsets). Though we logically form our subset around RS-related classes, we do not claim an optimal semantic set. The stark performance difference between training with our subset and random subsets may warrant further study of which classes are relevant and which are harmful. 

\subsubsection*{Performance on Rural Scenes}
CAFe-DINO performs strongly on urban datasets such as Potsdam and Vaihingen, but results on OpenEarthMap show poor discrimination between visually similar land-cover textures (e.g., grass and crops, see Fig. 9 of supplementary material). This is likely a limitation of large-scale natural image pretraining, where such distinctions are underrepresented. Incorporating texture-aware representations or multi-spectral information could alleviate this issue.

\section{Acknowledgment}
This work was supported by NASA under Grant 80NSSC22K1163.

\onecolumn
\twocolumn[
]
{
    \small
    \bibliographystyle{ieeenat_fullname}
    \bibliography{main}
}

\clearpage
\setcounter{page}{1}
\maketitlesupplementary

\section{Prompt Details}
\label{sup:prompt}

Every dataset contains a named set of classes. Instead of using these names directly, we have found it beneficial to make some logical modifications to the existing names, which we list here in \cref{tab:classes} for reproducibility. Our heuristic in making substitutions is to target classes with complex or abstract semantic names (e.g. ``agriculture'') and replace them with something simple and concrete (e.g. ``farm''). There may be more gains to be realized with sophisticated prompt tuning, but this is beyond the scope of our work.
Note that the semantic names used for Potsdam and Vaihingen are left unmodified. We believed that the model would struggle to apply the ``low vegetation'' class, so we attempted some simple substitutions (grass, greenery, etc.) only to find that using low vegetation gave the best performance. For LoveDA, we realized small performance gains by substituting ``forest'' and ``agriculture'' with ``tree'' and ``farm'', respectively. For OpenEarthMap, we substituted ``bareland'', ``rangeland'', ``developed space'', and ``agriculture land'' with ``barren'', ``grass'', ``pavement'', and ``cropland'', respectively.

Besides class names, the other variable in forming prompts is the choice of surrounding sentence. Ensembling is a common approach, in which a group of ``wrapper'' prompts is submitted to the text encoder for each semantic class, and the mean of all embeddings is used downstream. We use the set of wrappers defined by \cite{li2025segearthov} in their code implementation, and repeat them here in \cref{tab:prompts} for easy reference.

\begin{table}[]
\centering
\caption{The prompt class name of the evaluation datasets.}
\begin{tabular}{l p{0.5\columnwidth}}
    \hline
    \textbf{Dataset} & \textbf{Class Name} \\
    \hline
    OpenEarthMap &
    barren, grass, pavement, road, tree, water, cropland, building \\
    
    LoveDA &
    building, road, water, barren, tree, farm \\
    
    Potsdam, Vaihingen &
    road, building, low vegetation, tree, car \\
    \hline
\label{tab:classes}
\end{tabular}
\end{table}

We show aggregated cost maps for Potsdam, Vaihingen, LoveDA, and OEM in \cref{fig:grid1}, \cref{fig:grid2} ,\cref{fig:grid3}, \cref{fig:grid4}, respectively, for both vanilla DINOv3 and CAFe-DINO. Aggregated cost maps are effectively a per-class probability score, so they are a good empirical heuristic of model uncertainty and identifying problem classes. For example, the ``car'' cost map for Potsdam is much sharper than the one for Vaihingen for both models, and therefore the predictions are weaker on the Vaihingen image (we attribute this particular phenomenon to the non-RGB spectra of the Vaihingen dataset). A core limitation of CAFe-DINO (and RS-training-free methods in general) can be gleaned from the cost maps of OEM, particularly the near-identical Grass and Cropland maps. We conjecture that the difference between texture-characterized classes such as grass and crops in a satellite image is too fine-grained for a model trained on natural imagery to distinguish, though more shape-characterized greenery such as trees are more easily identified.

CAFe-DINO cost maps are generally sharper and higher-contrast than DINOv3 cost maps as a result of cost aggregation. Some classes are segmentable by DINOv3 out-of-the-box (such as cars in Potsdam), while others produce almost completely unstructured cost maps (such as Low Veg. in Potsdam). These unstructured cost maps benefit the most from cost aggregation.

Note that the cost map of a class is aggregated relative to other classes in the given semantic set, so changing the input set of classes will change the cost map of a given class.

\begin{table}[h]
\centering
\caption{Prompt wrappers for CAFe-DINO. ``\{\}'' is a placeholder for a desired semantic class.}
\begin{tabular}{l p{13cm}}
    \hline
    a photo of \{\}\\
    an image of \{\}\\
    a photograph of \{\}\\
    a picture of \{\}\\
    a photo of a \{\}\\
    an image of a \{\}\\
    a photo of the \{\}\\
    an image of the \{\}\\
    a close-up photo of \{\}\\
    a cropped image featuring \{\} \\
    \hline
\label{tab:prompts}
\end{tabular}
\end{table}

\clearpage

\section{COCO-Stuff Subset}

We use a subset of the COCO-Stuff dataset containing only the following classes:
\begin{description}[leftmargin=0pt]
\item[Classes:] \small bicycle, car, motorcycle, airplane, bus, train, truck,
boat, bridge, building, bush, dirt, fence, grass, gravel, ground, hill, house,
leaves, metal, mountain, mud, pavement, plant, platform, playing field, railing,
railroad, river, road, rock, roof, sand, sea, skyscraper, snow, stone,
structural, tree, water, wood.
\end{description}

Our heuristic in forming this dataset was to remove semantic classes that could never exist in satellite imagery at current resolution capabilities, such as household objects and food items. Using a reduced semantic set brings the benefit of smaller cost volumes, resulting in a substantial decrease in GPU memory and FLOPs for both training and inference. Additionally, we find that training convergence with the reduced set is very short (45,000 iterations).

\section{Additional Cost Maps}
\label{sup:costmap}

\begin{figure*}[]
    \centering
    \setlength{\tabcolsep}{1.5pt}

    \begin{tabular}{c*{7}{c}}
        & Road & Building & Low Veg. & Tree & Car  & Prediction \\

        \raisebox{26pt}{CAFe-DINO} &
        \includegraphics[width=0.14\textwidth]{img/costmaps_outr/costmap_0.jpg} &
        \includegraphics[width=0.14\textwidth]{img/costmaps_outr/costmap_1.jpg}&
        \includegraphics[width=0.14\textwidth]{img/costmaps_outr/costmap_2.jpg}&
        \includegraphics[width=0.14\textwidth]{img/costmaps_outr/costmap_3.jpg}&
        \includegraphics[width=0.14\textwidth]{img/costmaps_outr/costmap_4.jpg}&
        \includegraphics[width=0.14\textwidth]{img/costmaps_outr/segmentation.jpg} \\

        \raisebox{26pt}{DINOv3} &
        \includegraphics[width=0.14\textwidth]{img/costmaps_dino/costmap_0.jpg} &
        \includegraphics[width=0.14\textwidth]{img/costmaps_dino/costmap_1.jpg}&
        \includegraphics[width=0.14\textwidth]{img/costmaps_dino/costmap_2.jpg}&
        \includegraphics[width=0.14\textwidth]{img/costmaps_dino/costmap_3.jpg}&
        \includegraphics[width=0.14\textwidth]{img/costmaps_dino/costmap_4.jpg}&
        \includegraphics[width=0.14\textwidth]{img/costmaps_dino/segmentation.jpg} \\
        &&&&&\raisebox{26pt}{True Mask}&
            \includegraphics[width=0.14\textwidth]{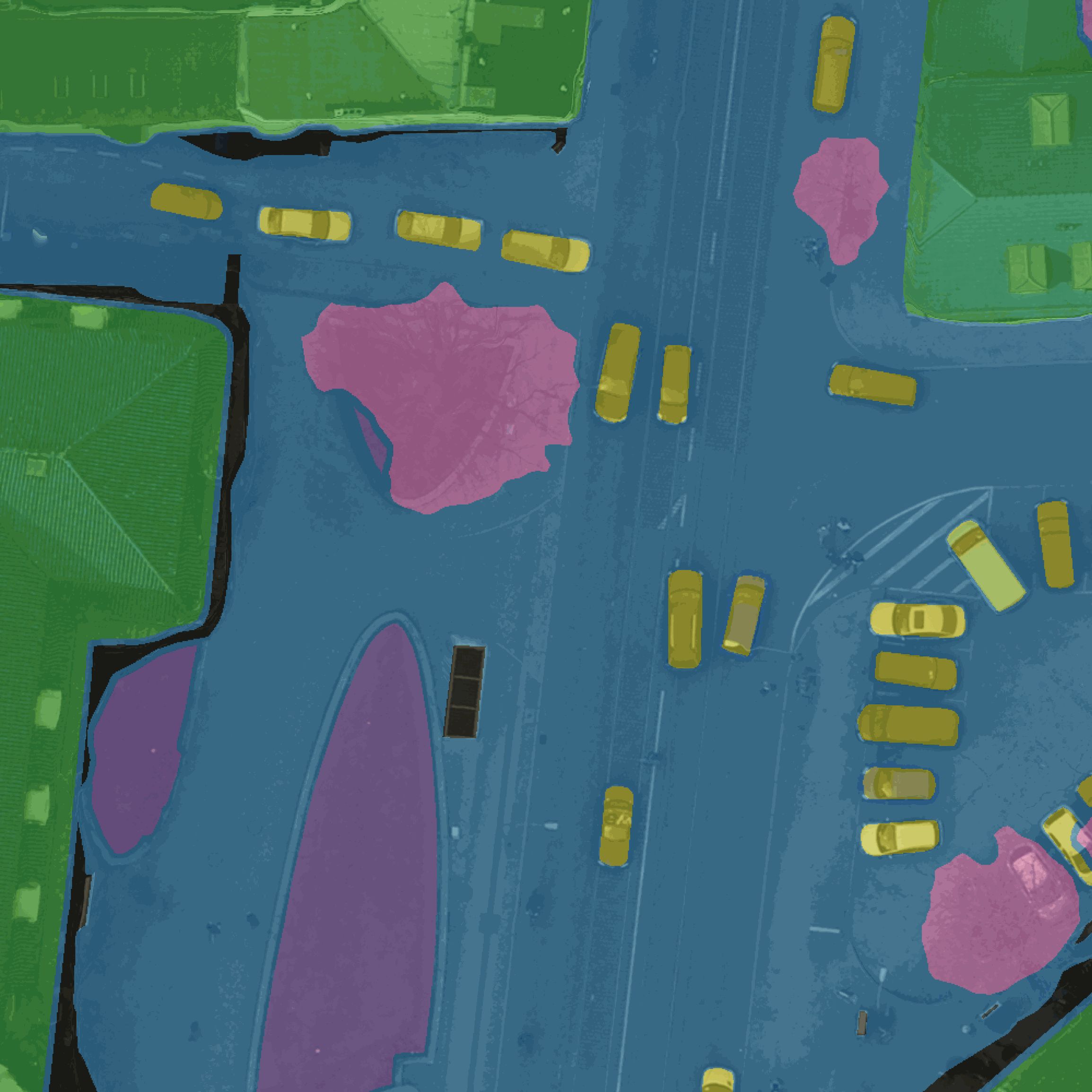} \\
    \end{tabular}

    \caption{Cost maps for a Potsdam image.}
    \label{fig:grid1}
\end{figure*}

\begin{figure*}[]
    \centering
    \setlength{\tabcolsep}{1.5pt}
    \begin{tabular}{c*{7}{c}}
        & Road & Building & Low Veg. & Tree & Car  & Prediction \\

        \raisebox{26pt}{CAFe-DINO} &
        \includegraphics[width=0.14\textwidth]{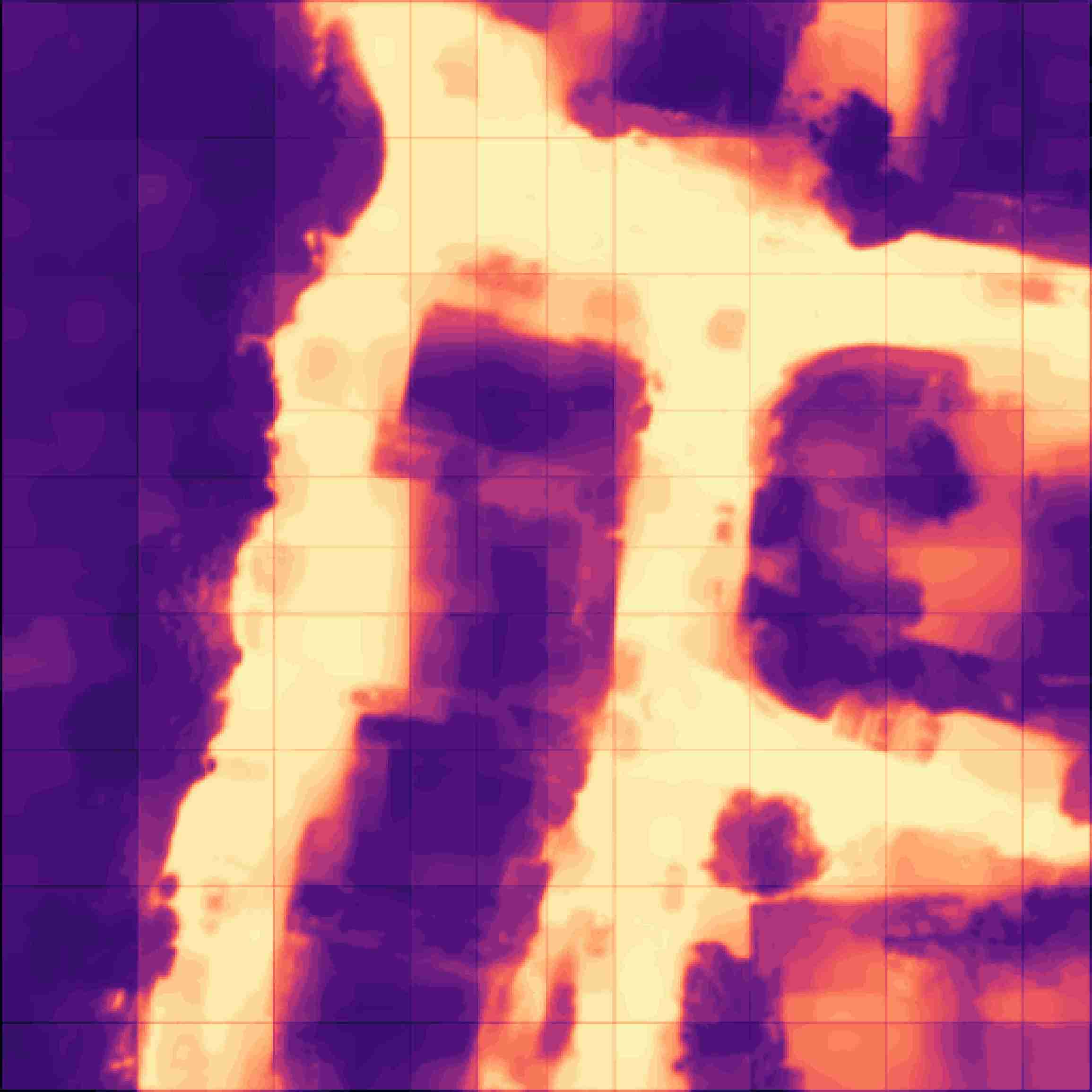} &
        \includegraphics[width=0.14\textwidth]{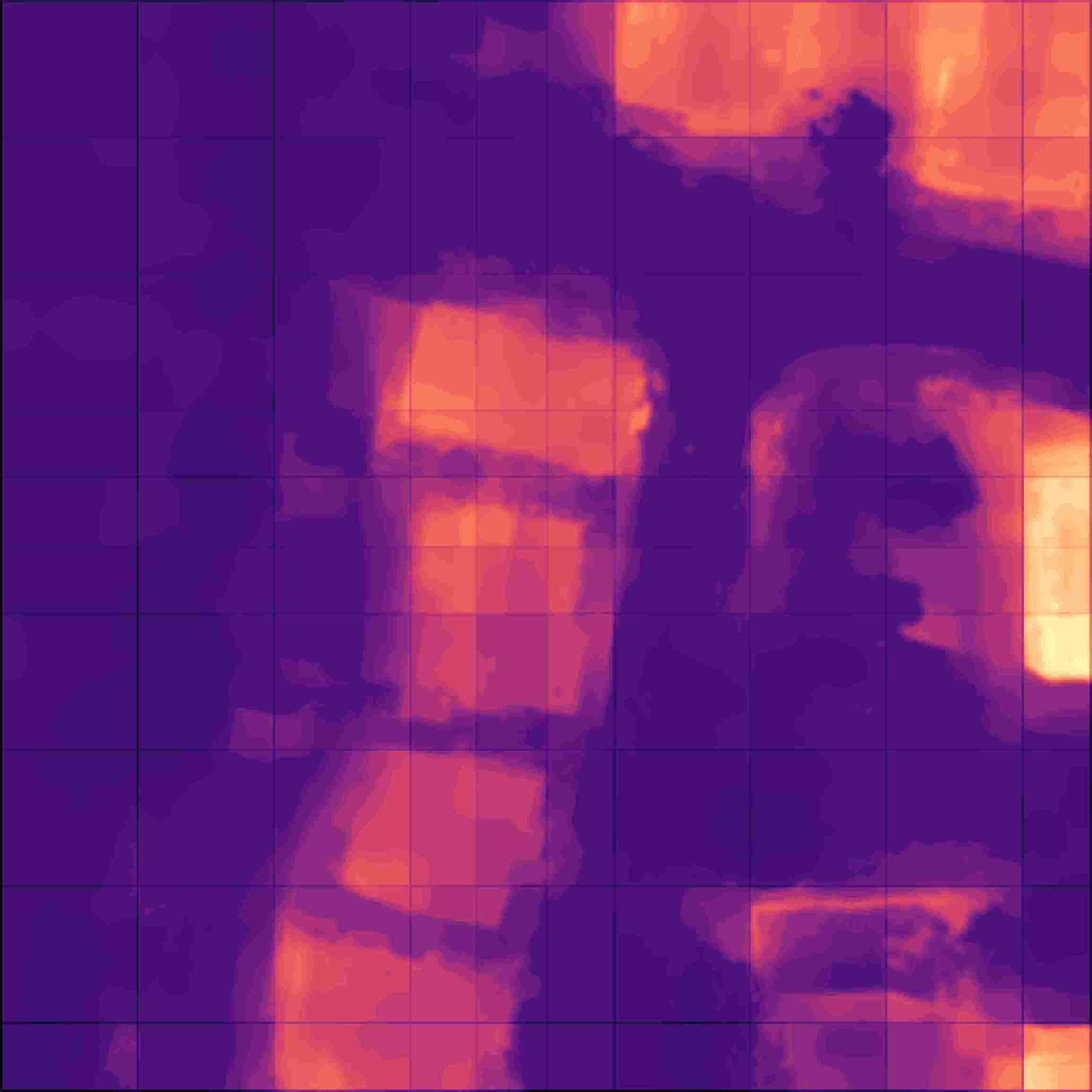}&
        \includegraphics[width=0.14\textwidth]{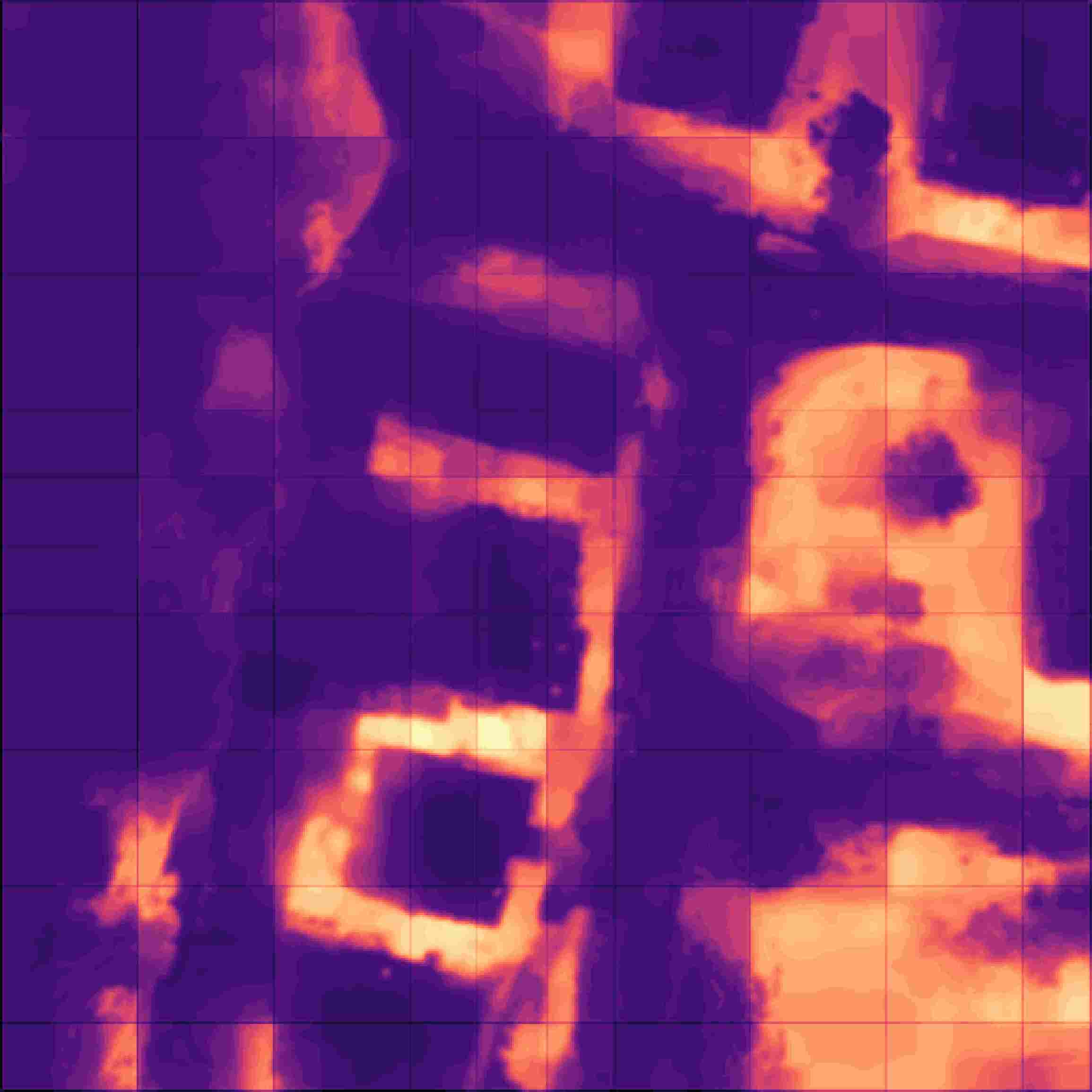}&
        \includegraphics[width=0.14\textwidth]{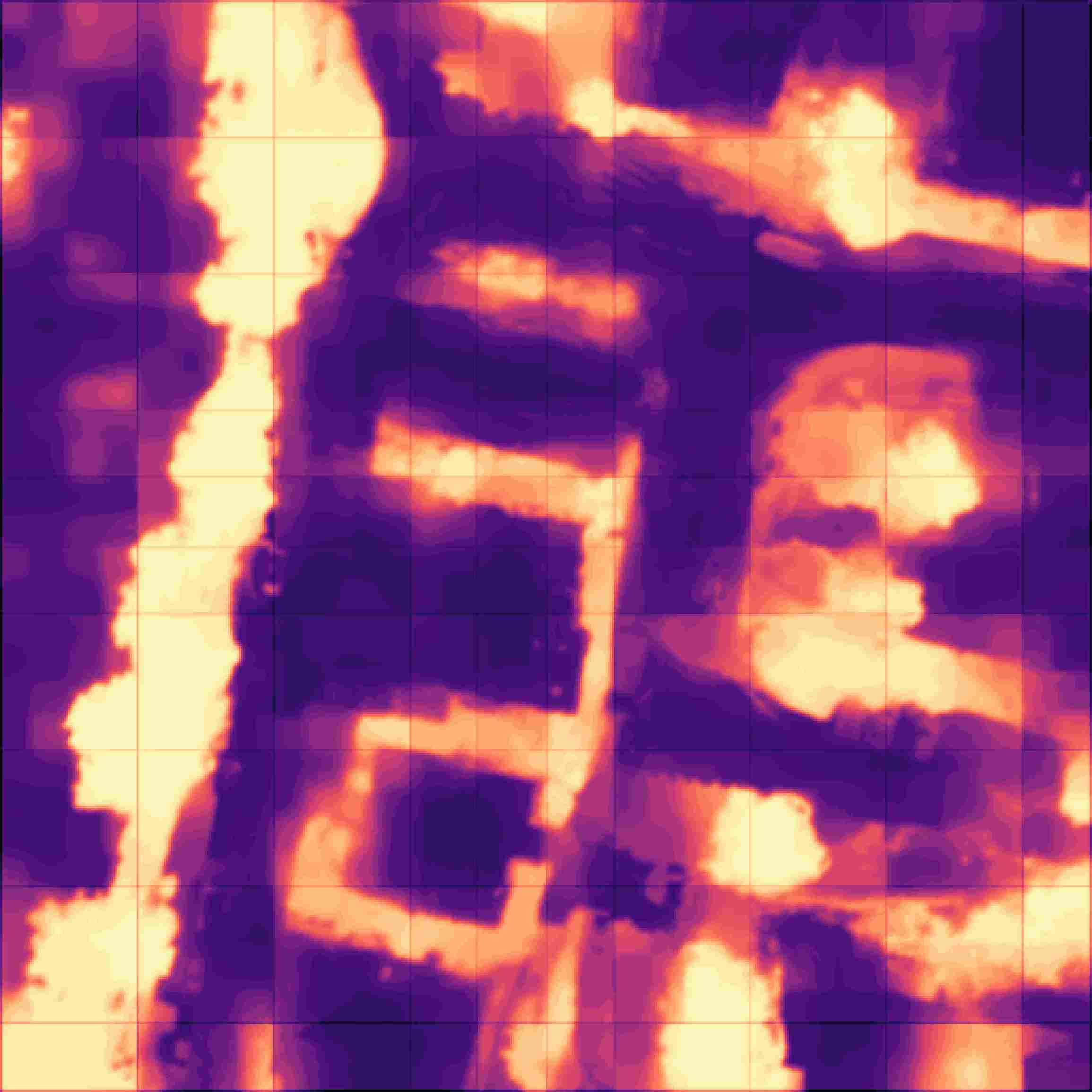}&
        \includegraphics[width=0.14\textwidth]{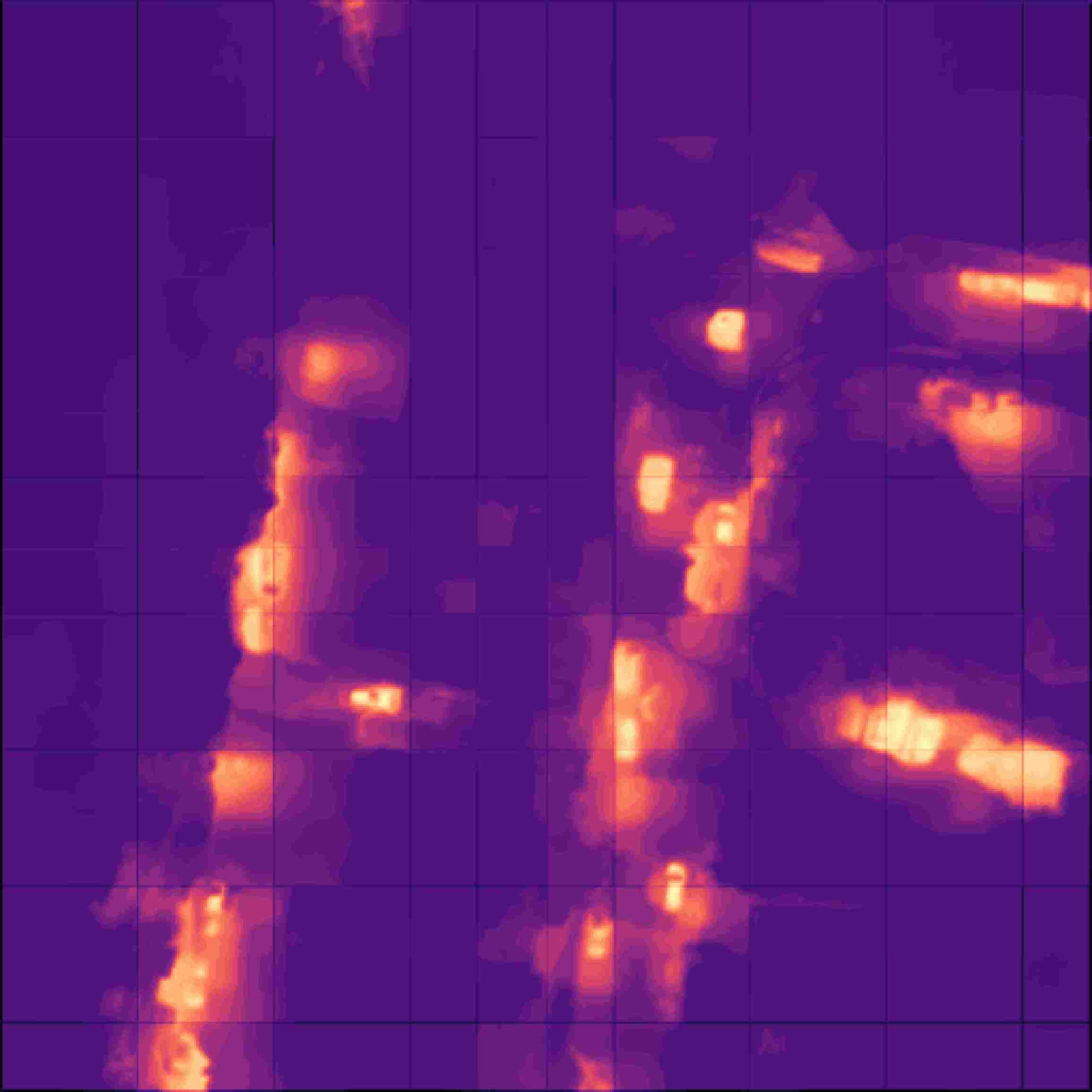}&
        \includegraphics[width=0.14\textwidth]{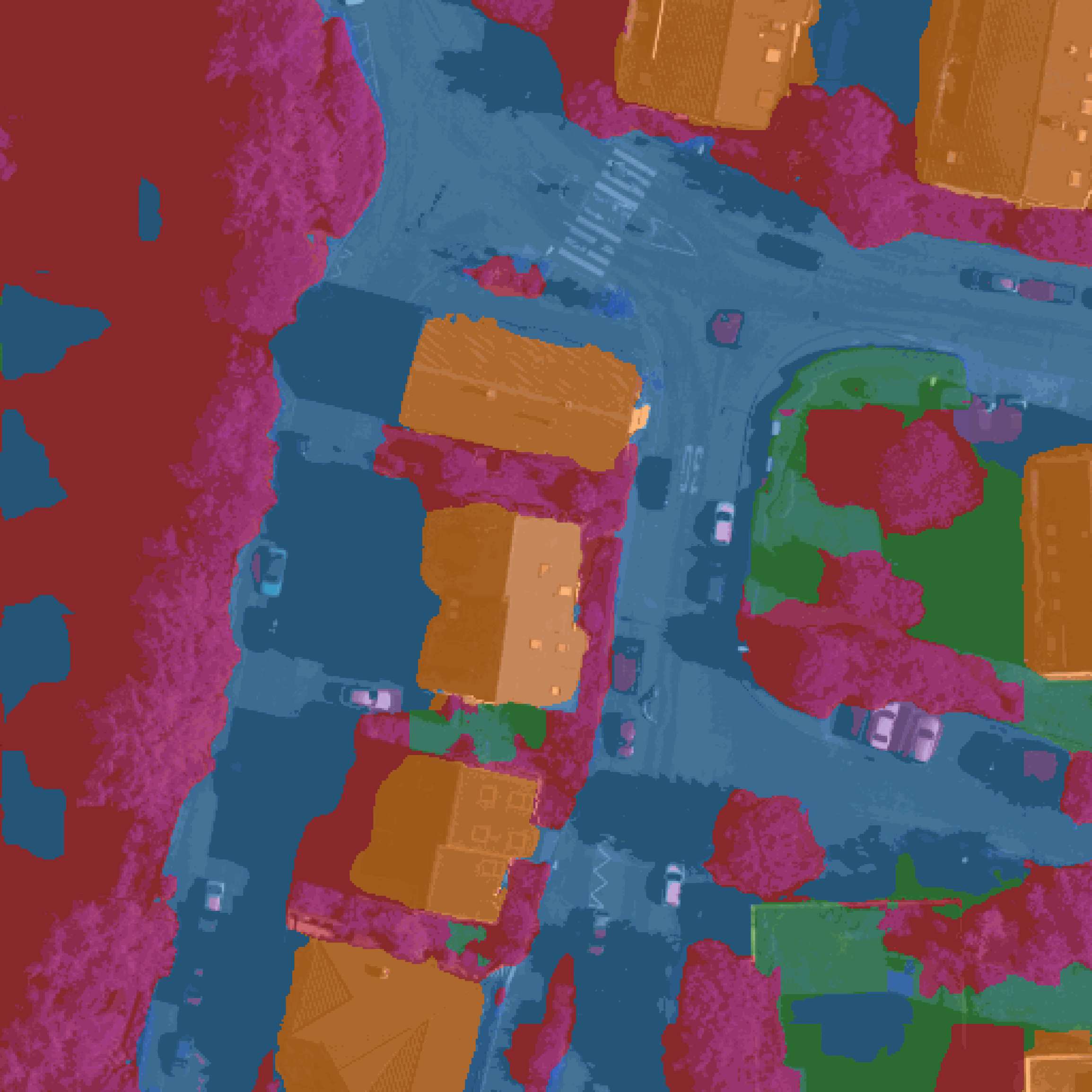} \\

        \raisebox{26pt}{DINOv3} &
        \includegraphics[width=0.14\textwidth]{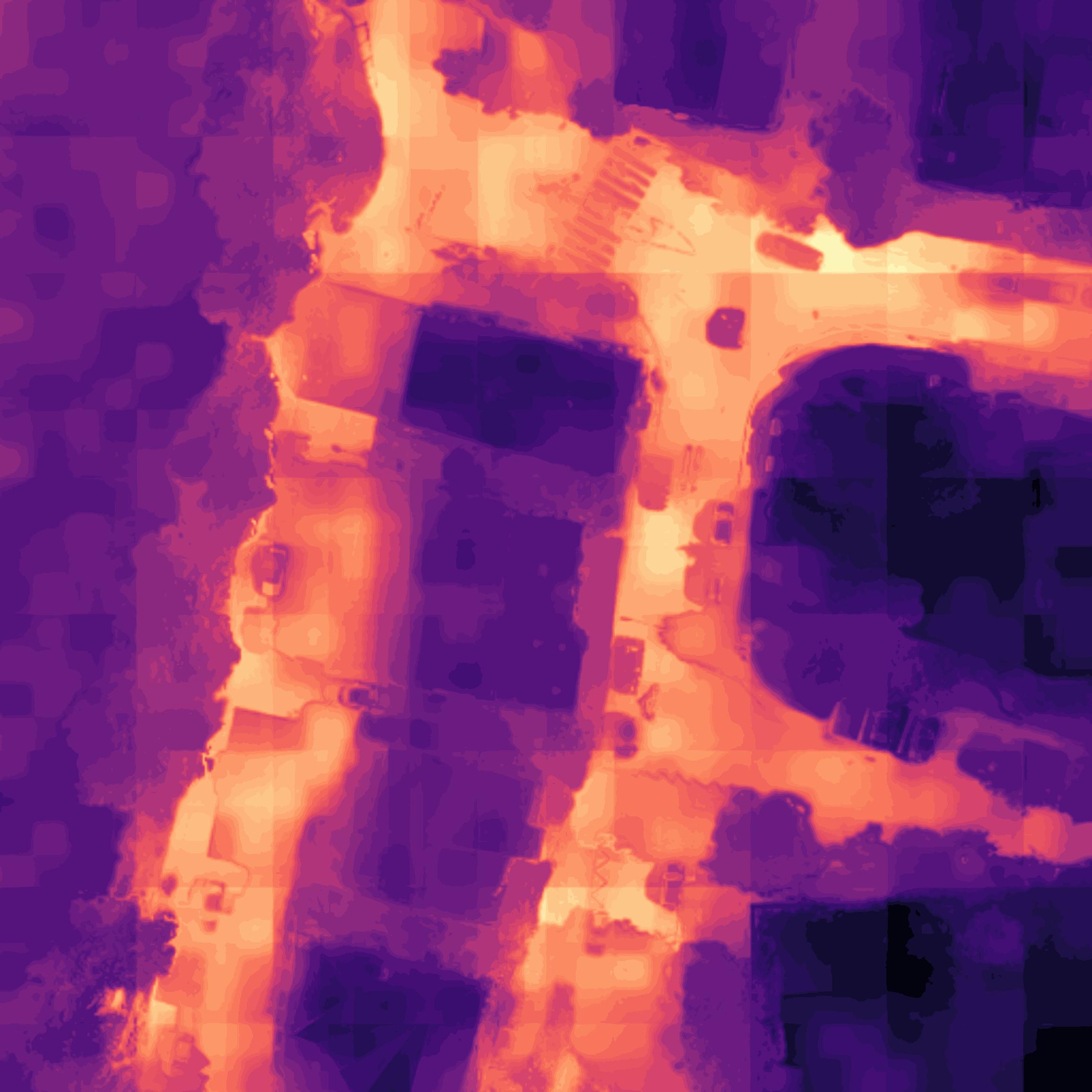} &
        \includegraphics[width=0.14\textwidth]{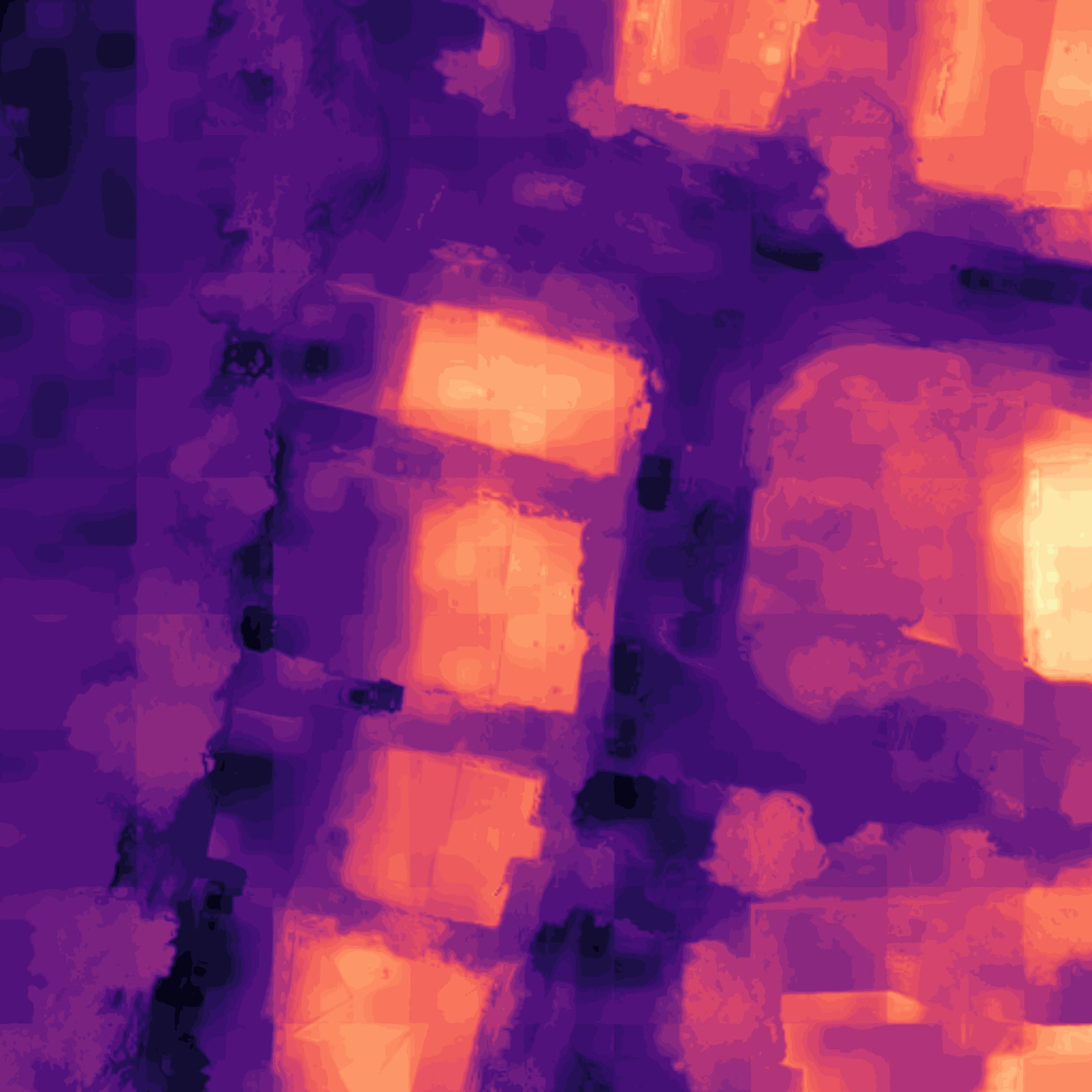} &
        \includegraphics[width=0.14\textwidth]{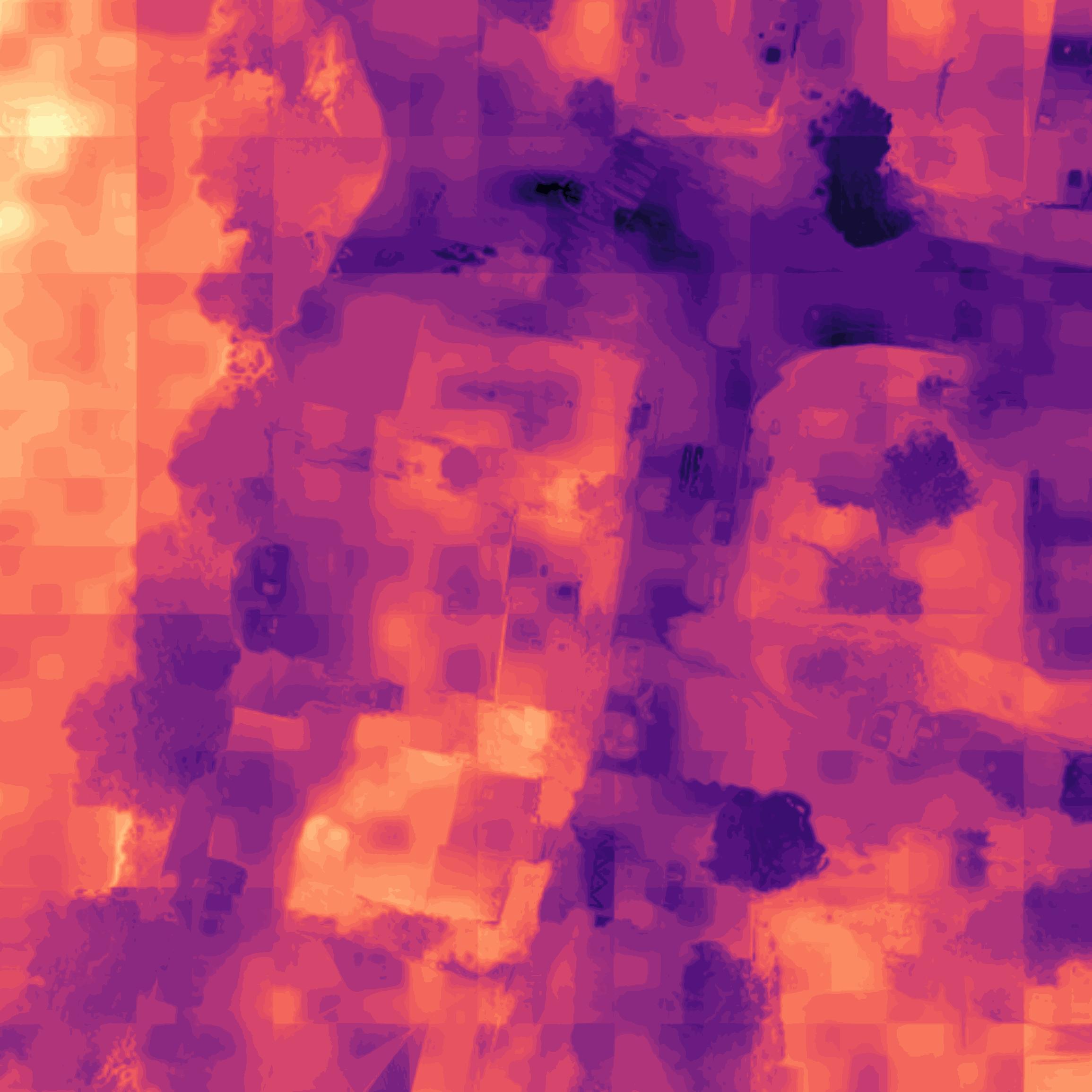} &
        \includegraphics[width=0.14\textwidth]{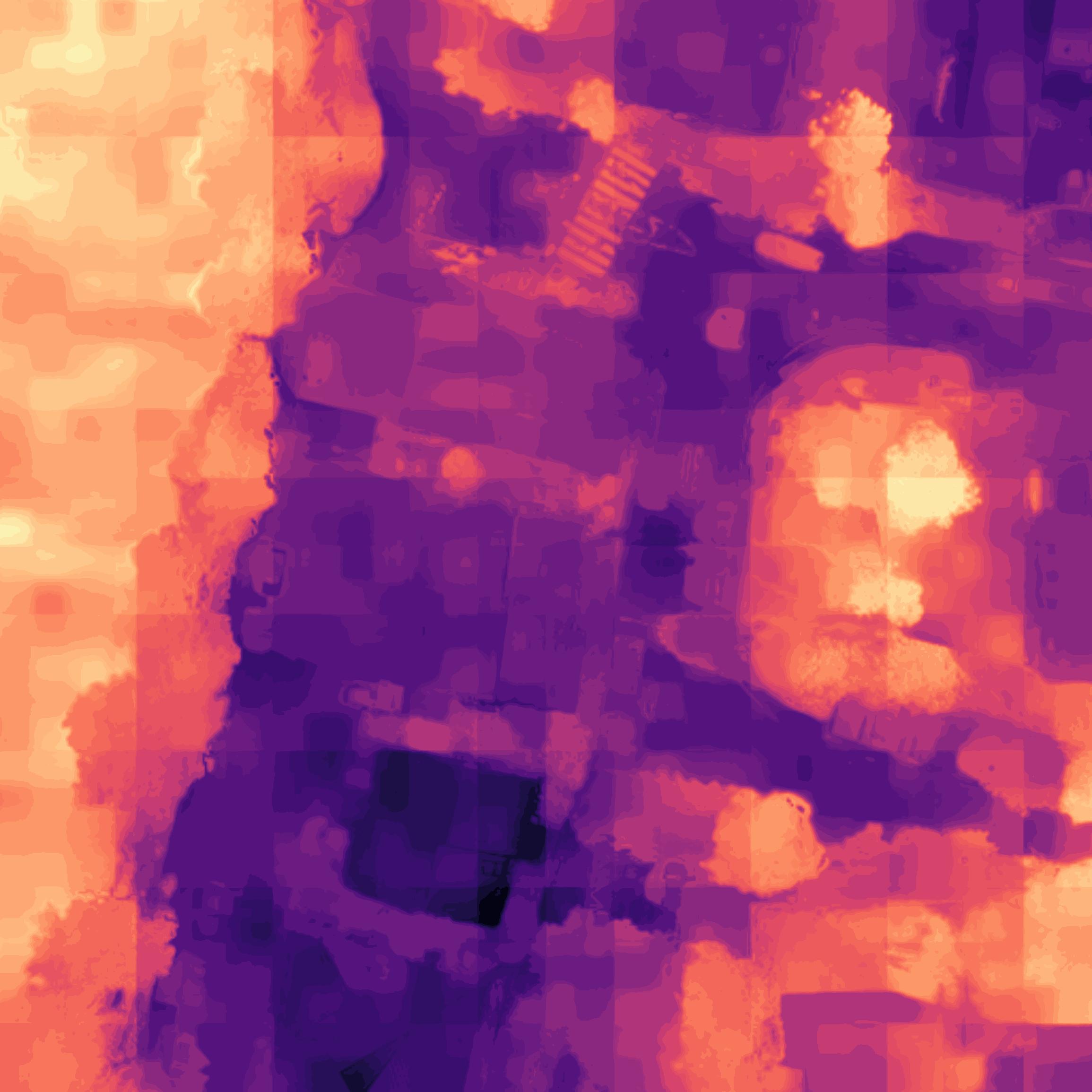} &
        \includegraphics[width=0.14\textwidth]{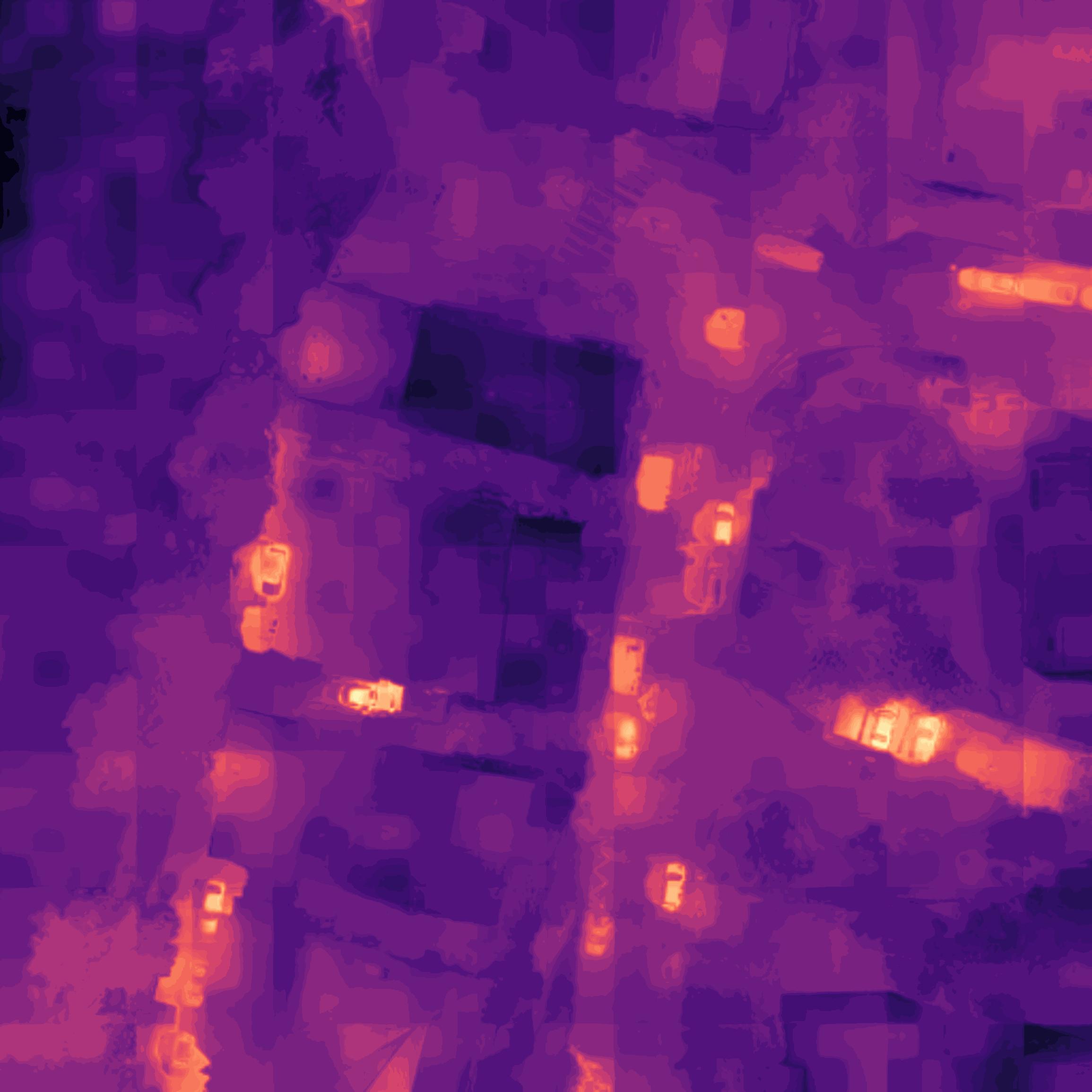} &
        \includegraphics[width=0.14\textwidth]{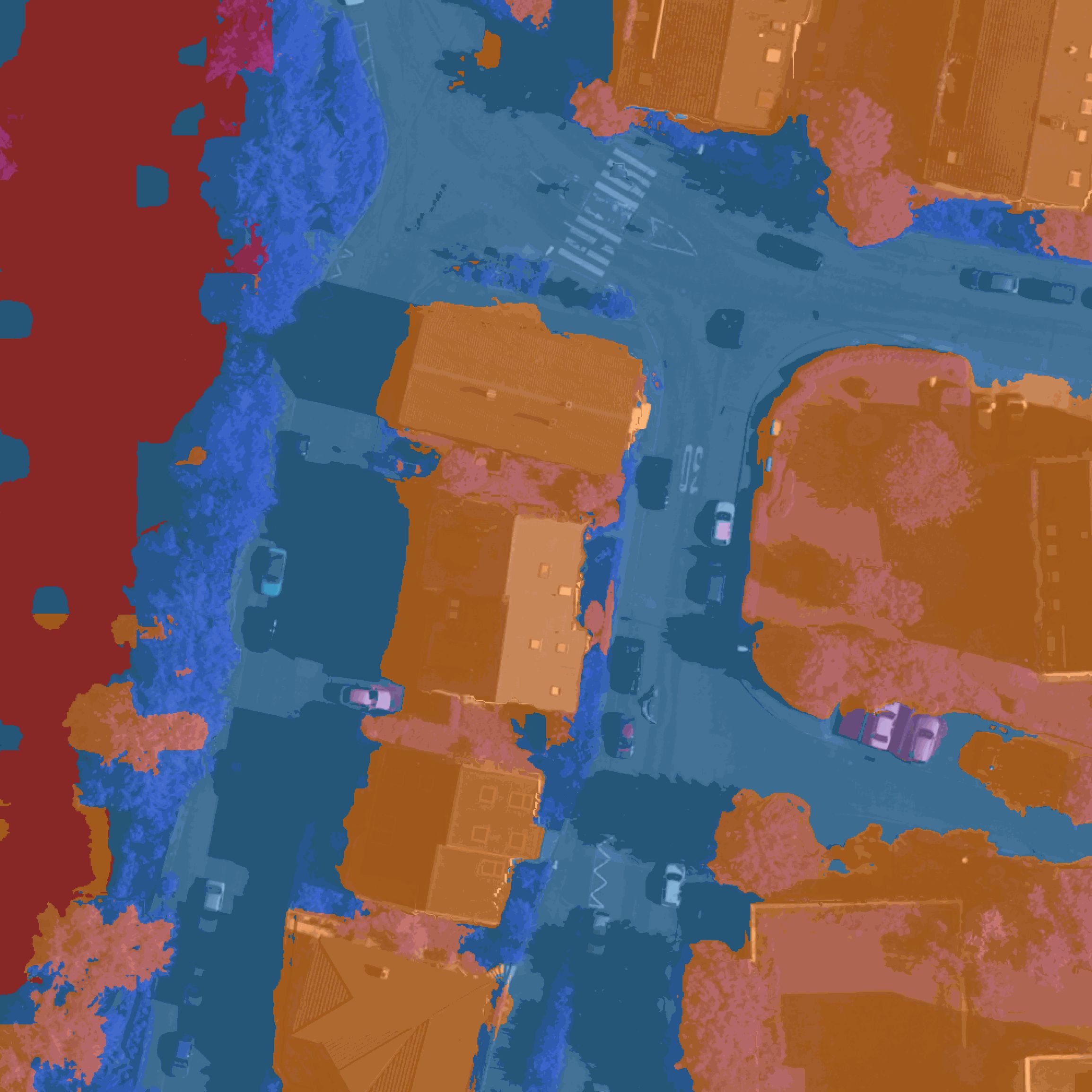} \\
        &&&&&\raisebox{26pt}{True Mask}&
            \includegraphics[width=0.14\textwidth]{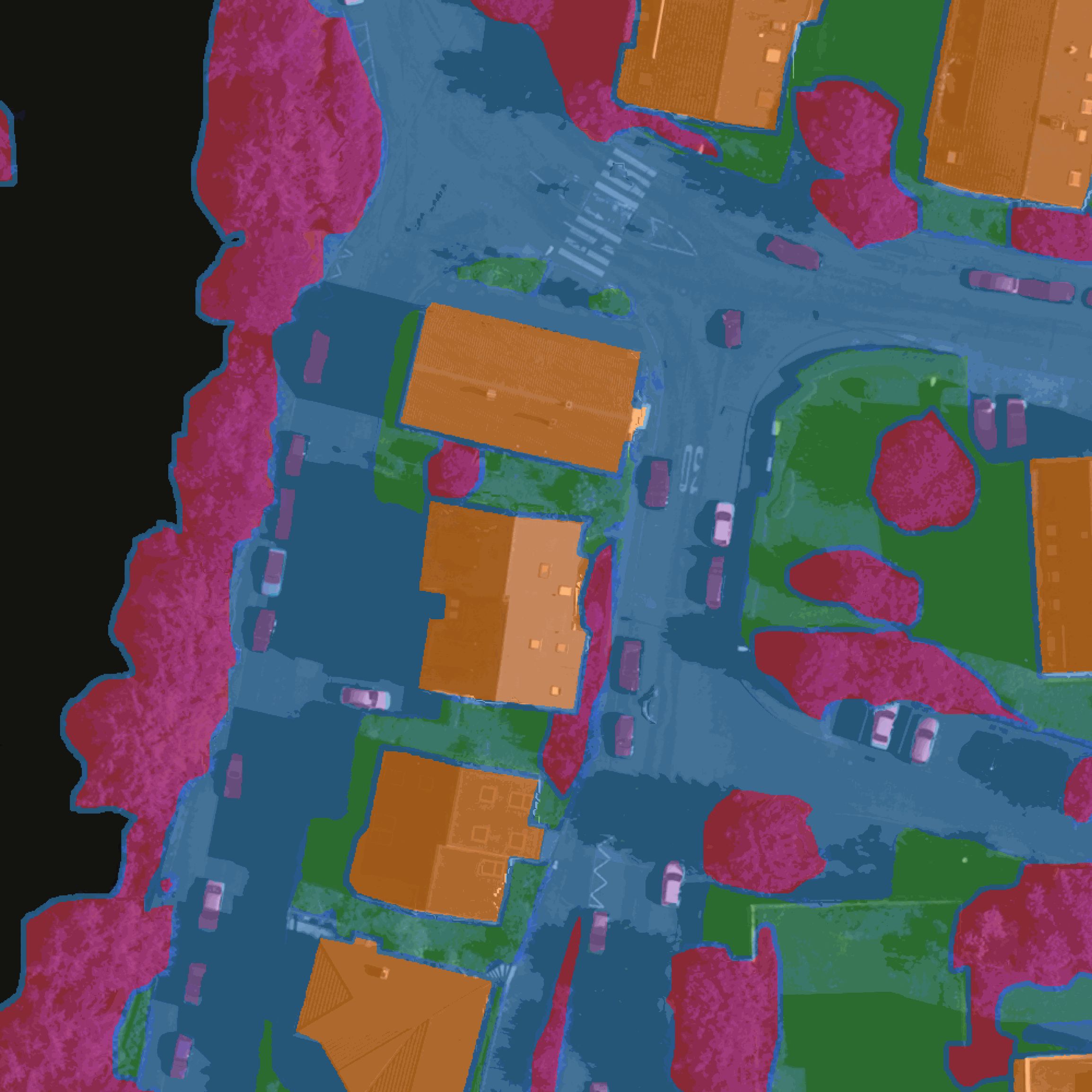} \\
    \end{tabular}

    \caption{Cost maps for a Vaihingen image.}
    \label{fig:grid2}
\end{figure*}

\begin{figure*}[]
    \centering
    \setlength{\tabcolsep}{1.5pt}

    \begin{tabular}{c*{7}{c}}
        & Building & Road & Water & Barren & Tree & Farm & Prediction\\

        \raisebox{26pt}{CAFe-DINO} &
        \includegraphics[width=0.12\textwidth]{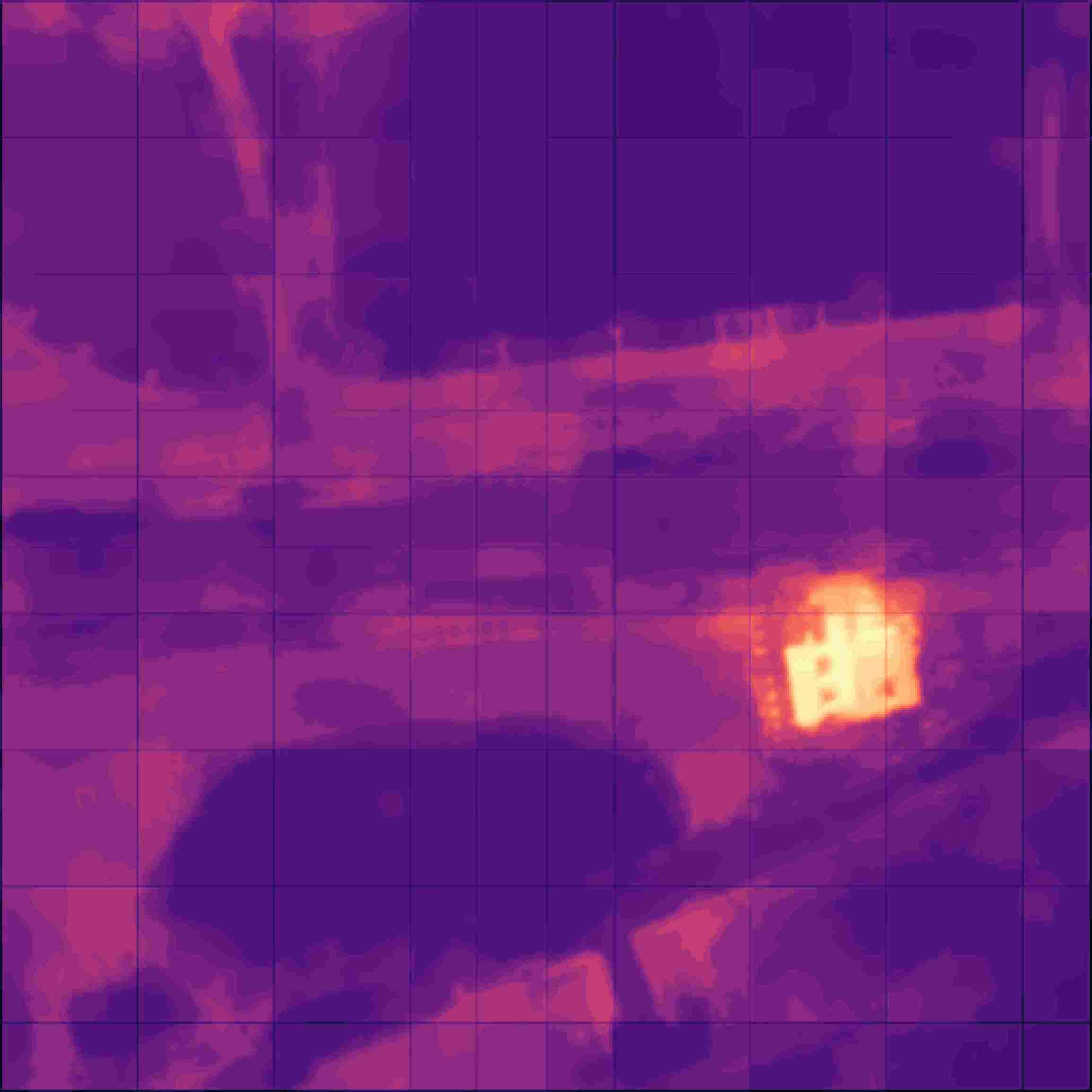} &
        \includegraphics[width=0.12\textwidth]{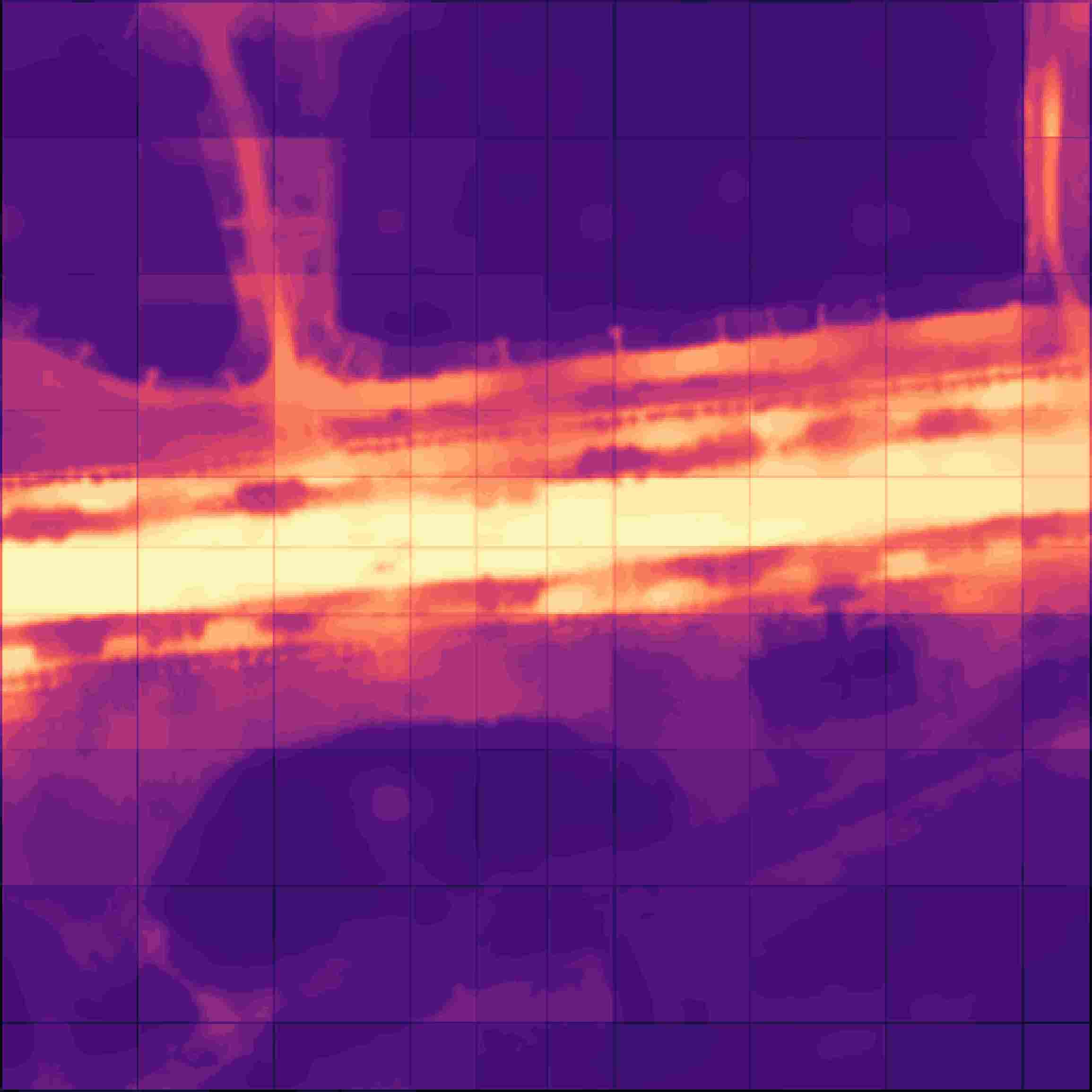}&
        \includegraphics[width=0.12\textwidth]{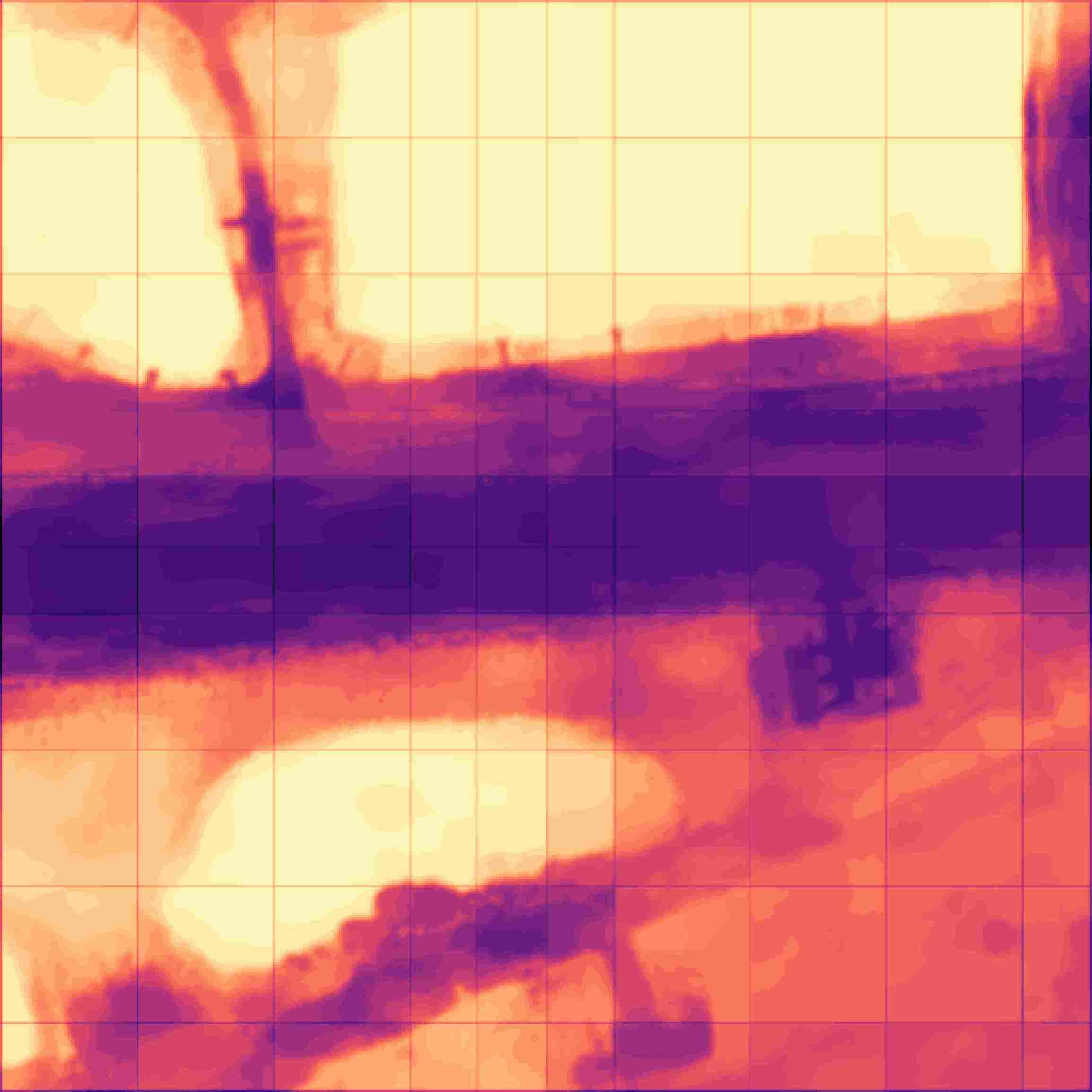}&
        \includegraphics[width=0.12\textwidth]{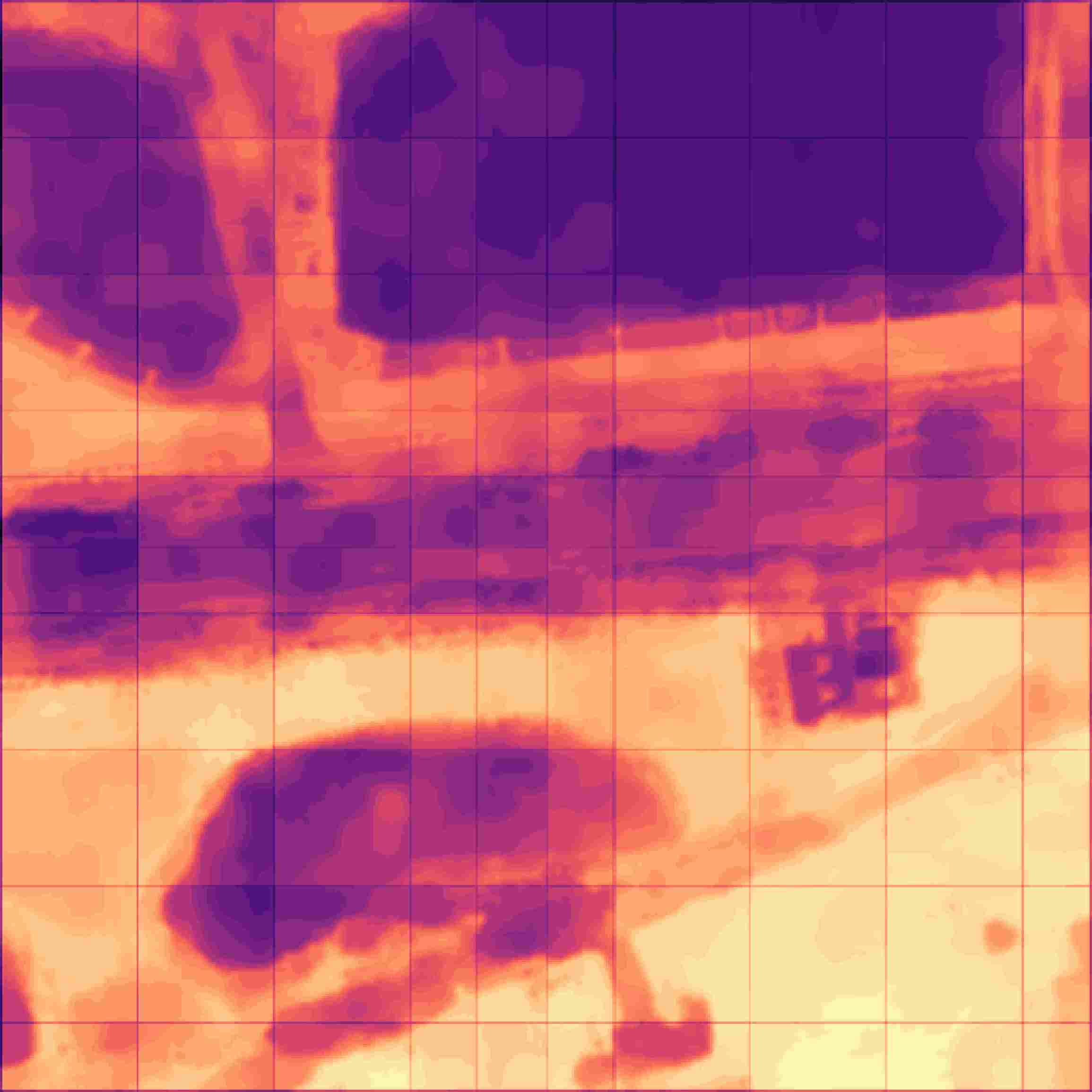}&
        \includegraphics[width=0.12\textwidth]{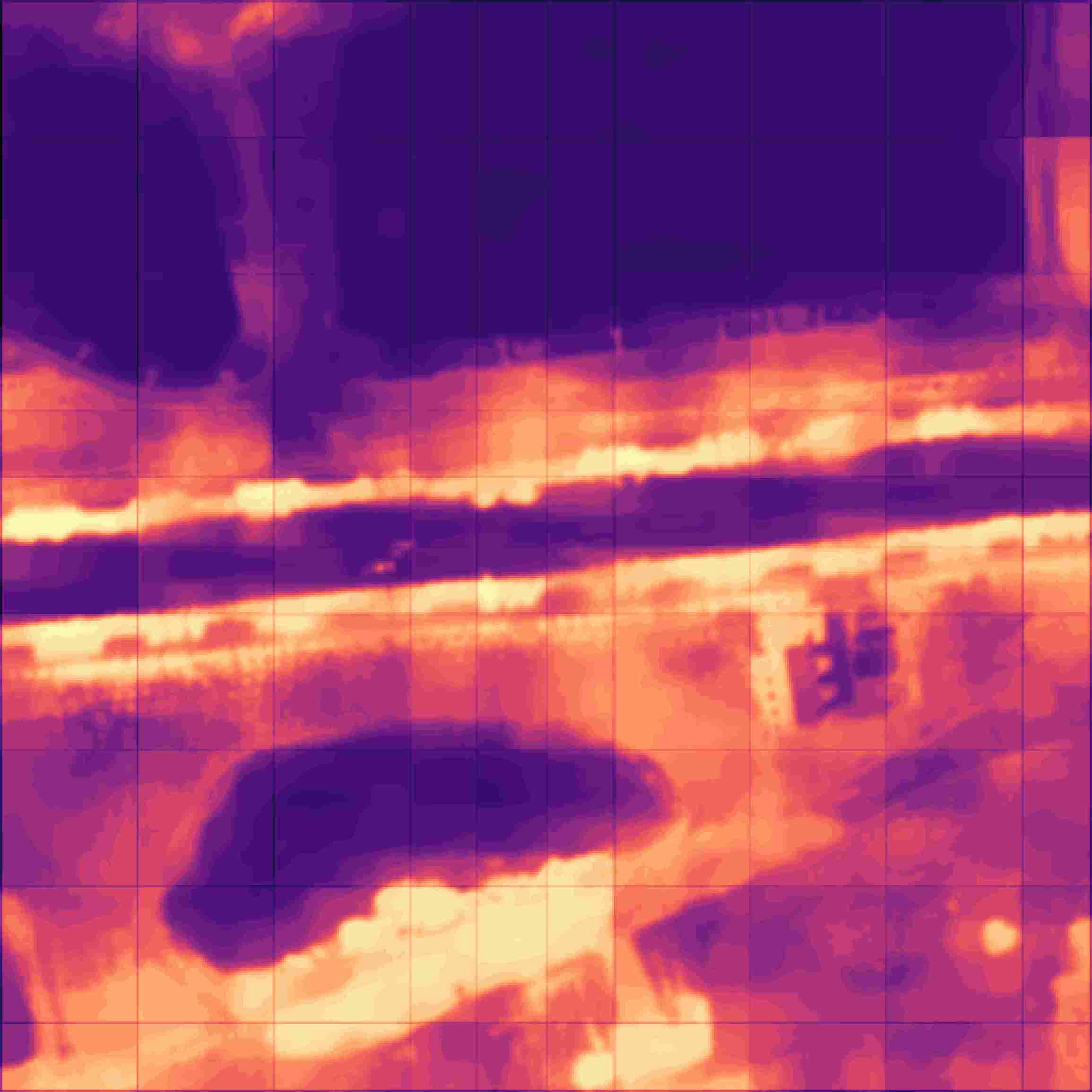}&
        \includegraphics[width=0.12\textwidth]{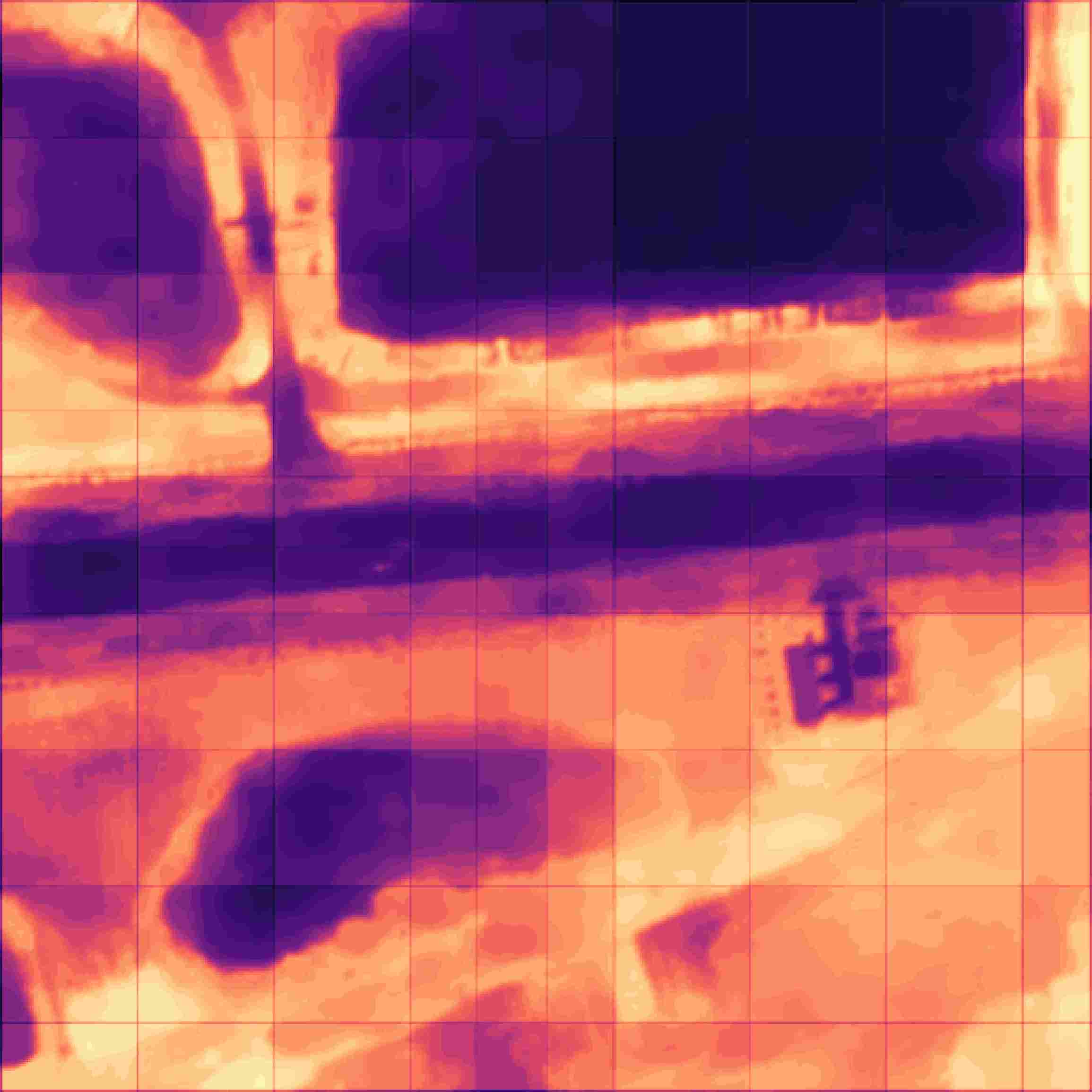}&
        \includegraphics[width=0.12\textwidth]{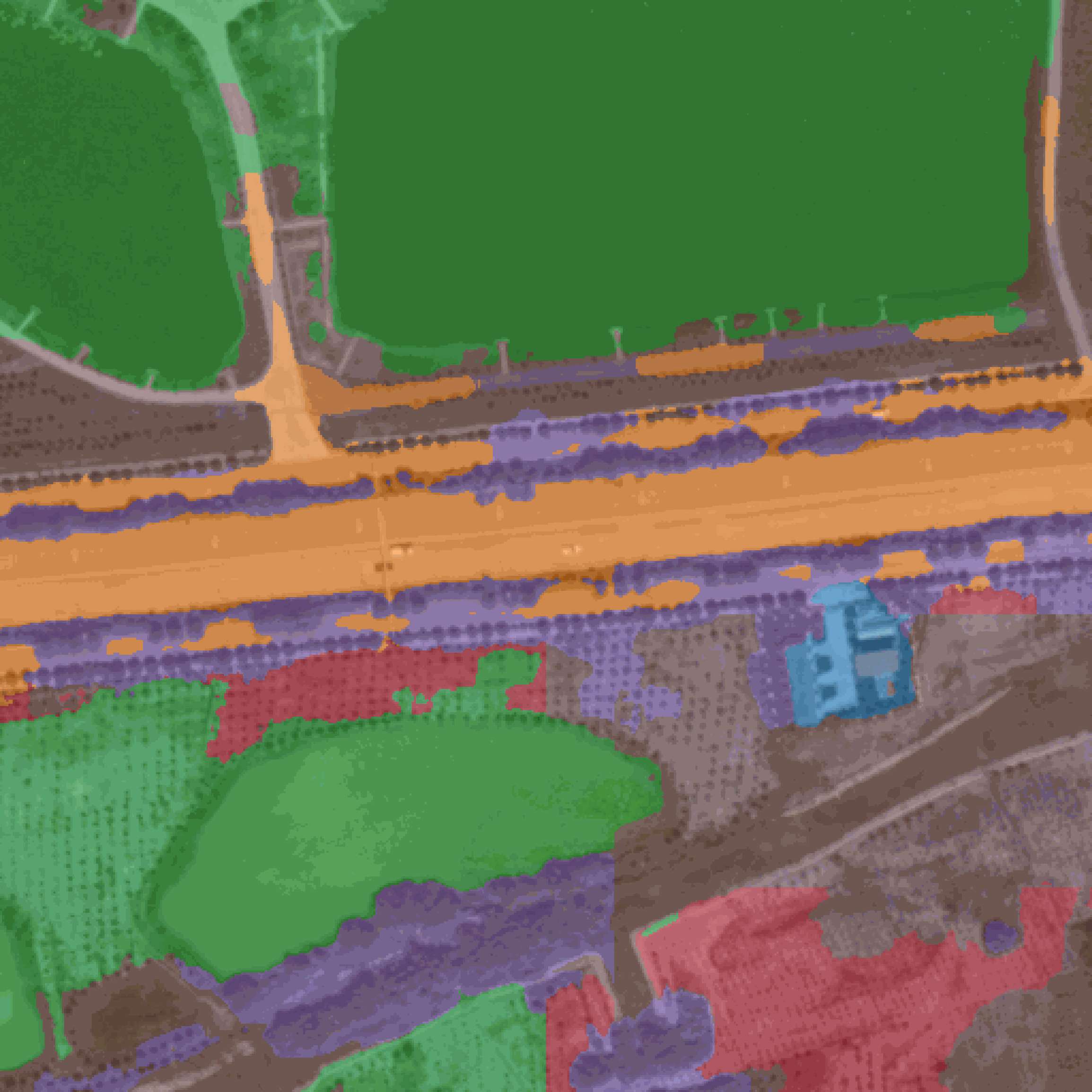} \\

        \raisebox{26pt}{DINOv3} &
        \includegraphics[width=0.12\textwidth]{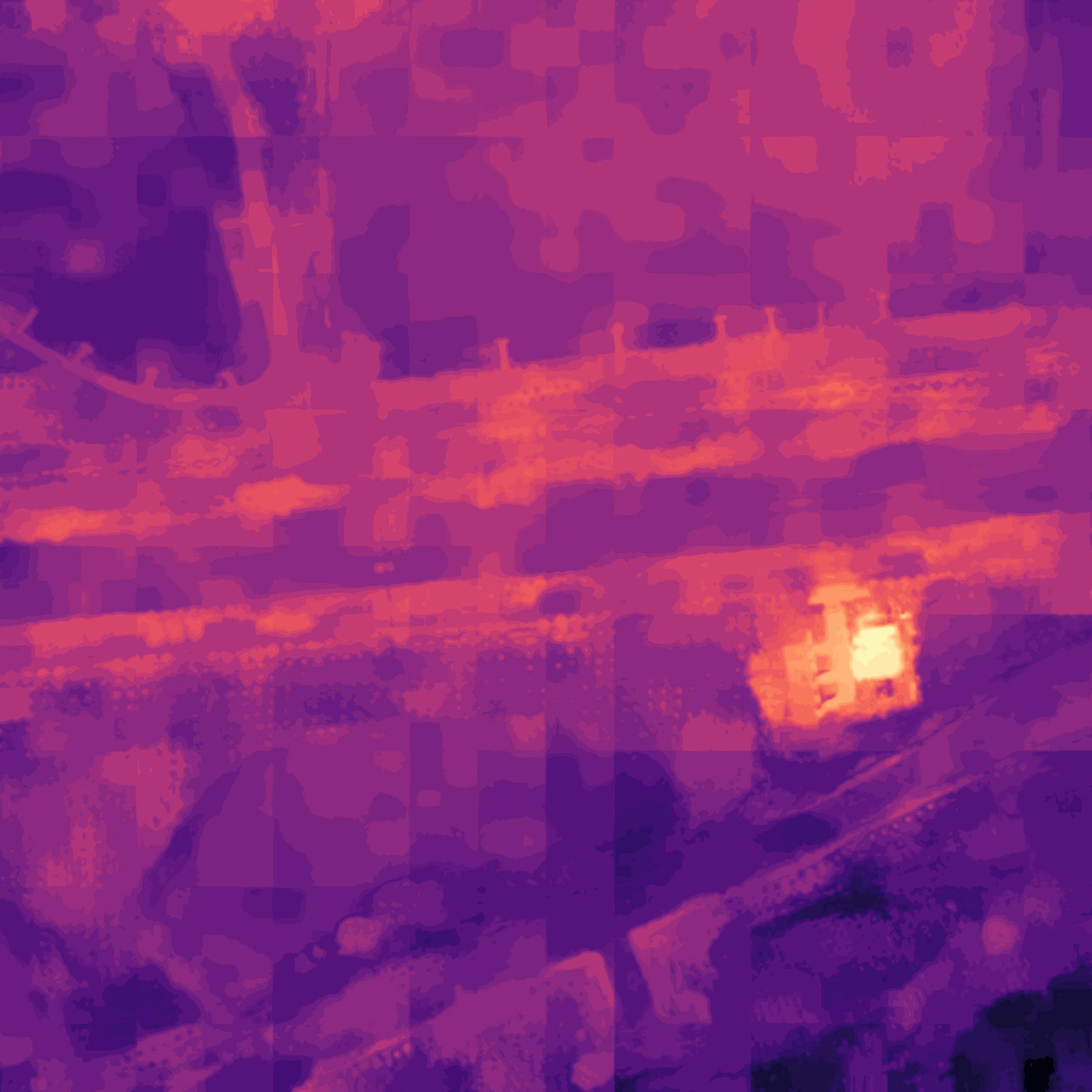} &
        \includegraphics[width=0.12\textwidth]{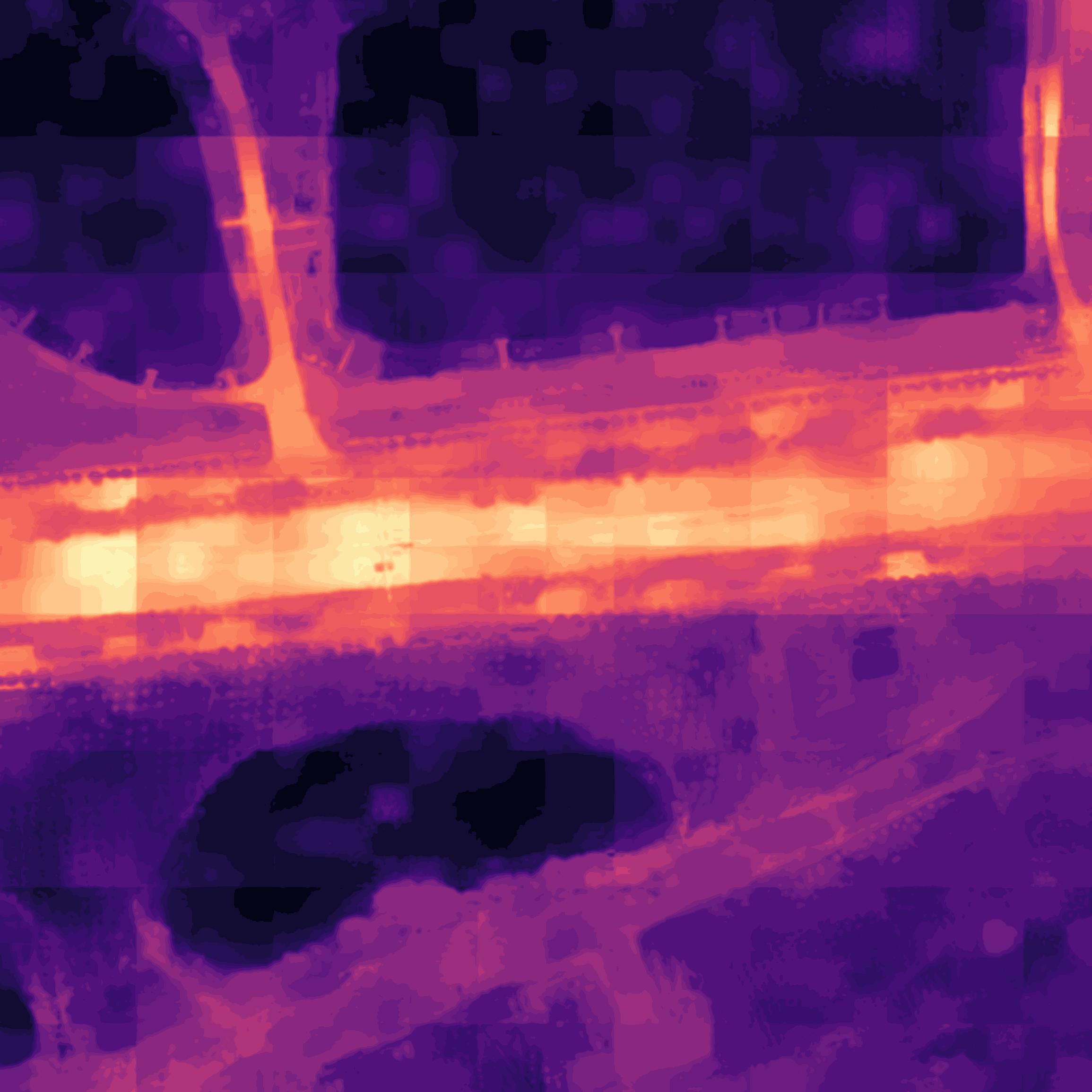} &
        \includegraphics[width=0.12\textwidth]{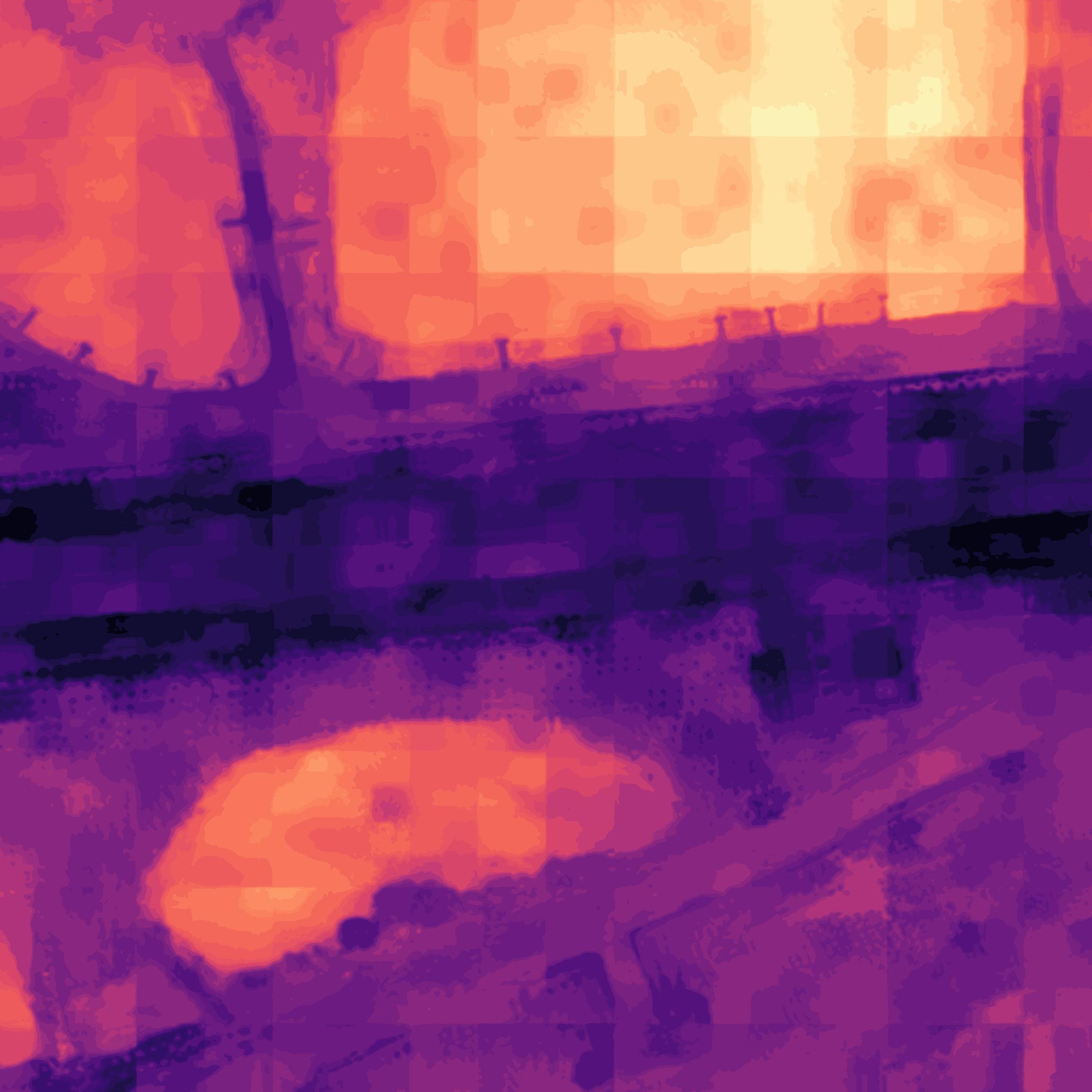} &
        \includegraphics[width=0.12\textwidth]{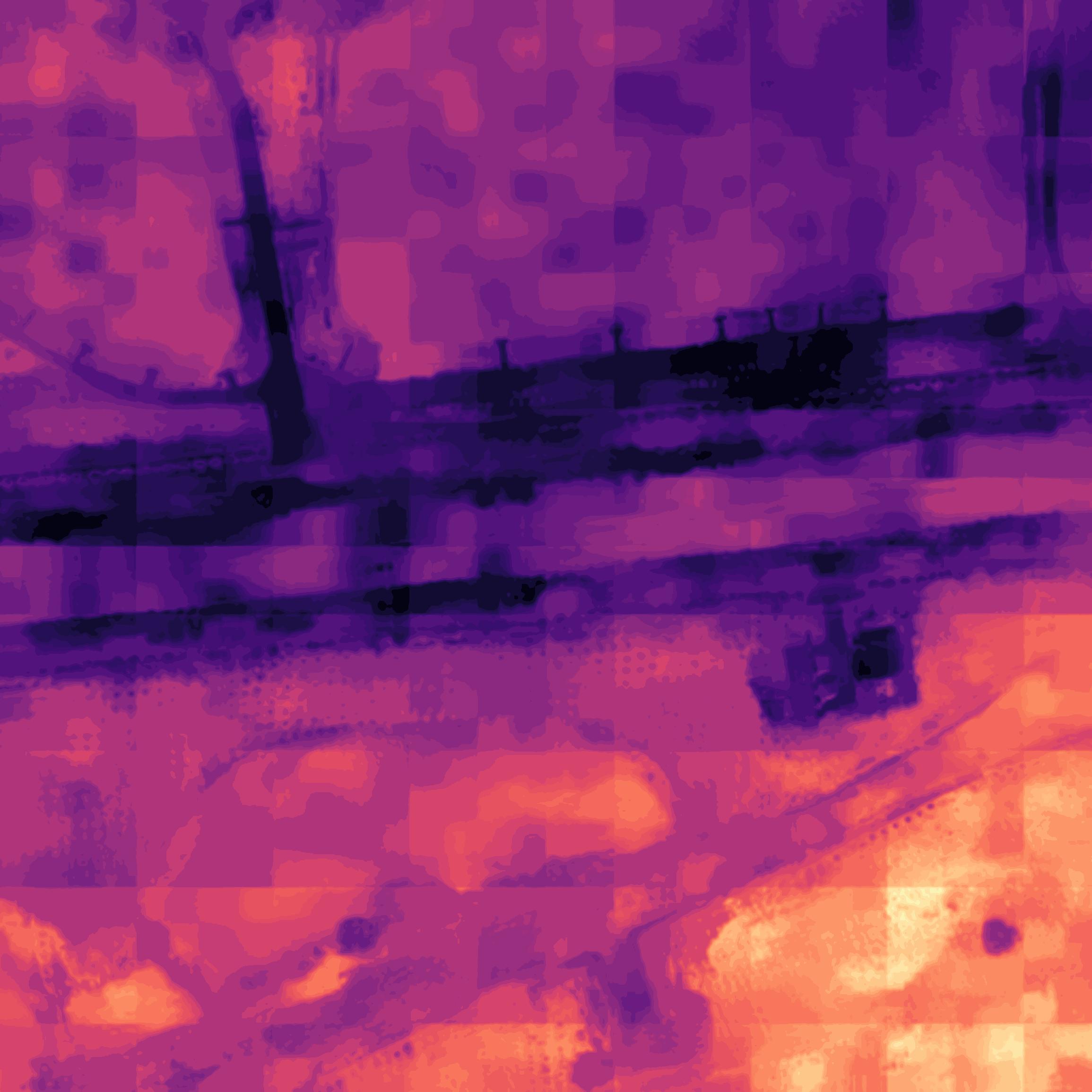} &
        \includegraphics[width=0.12\textwidth]{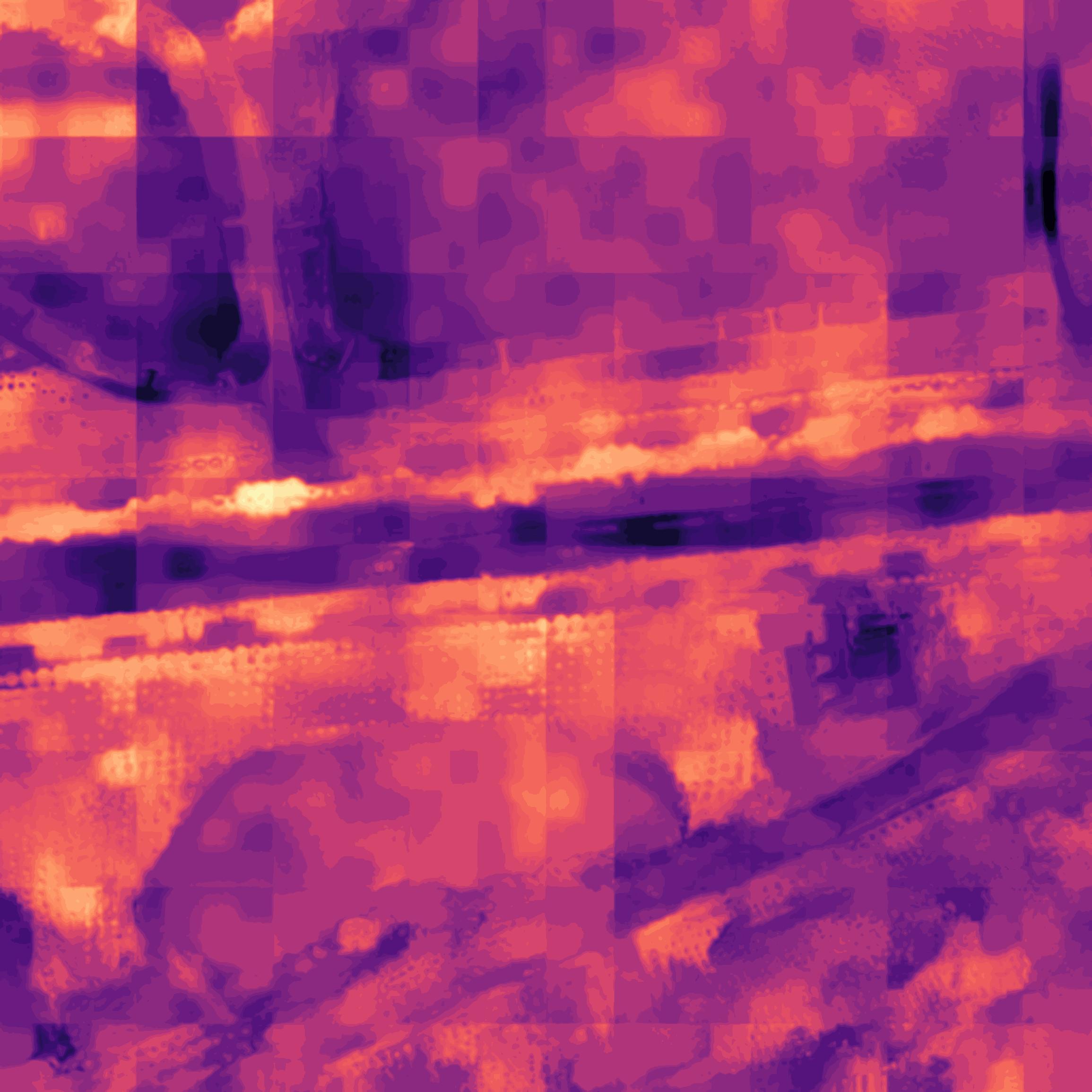} &
        \includegraphics[width=0.12\textwidth]{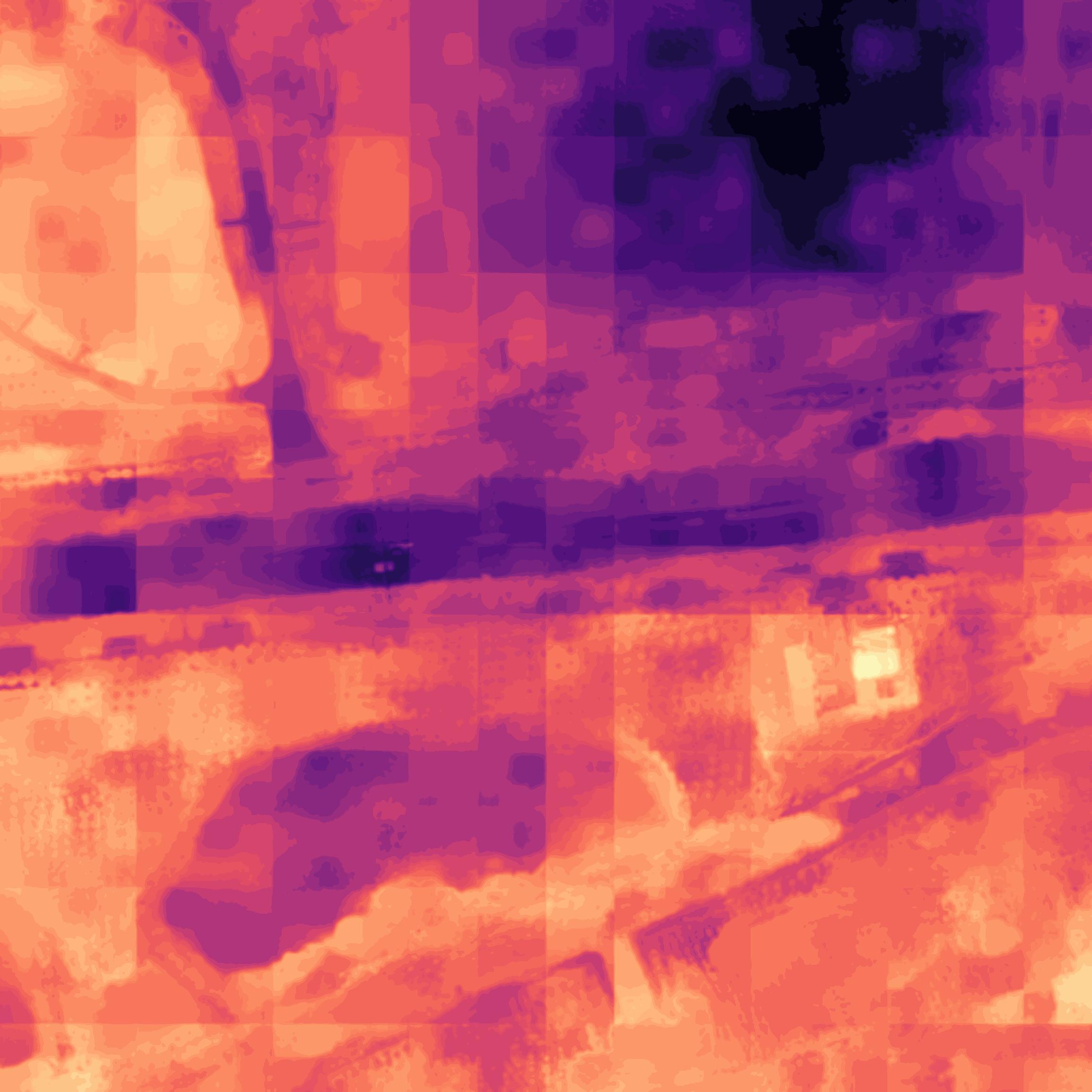} &
        \includegraphics[width=0.12\textwidth]{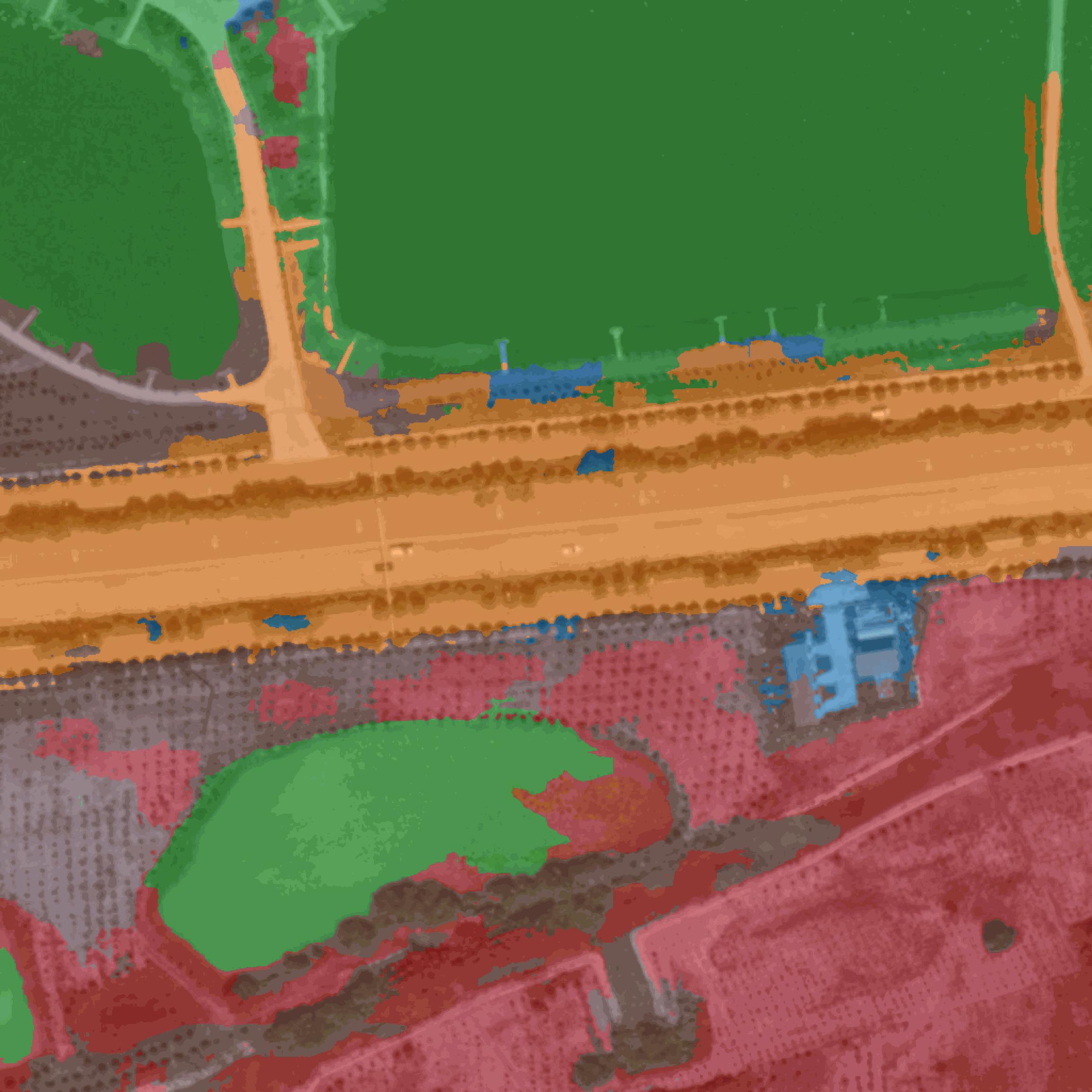}\\

        &&&&&&\raisebox{26pt}{True Mask}&
            \includegraphics[width=0.12\textwidth]{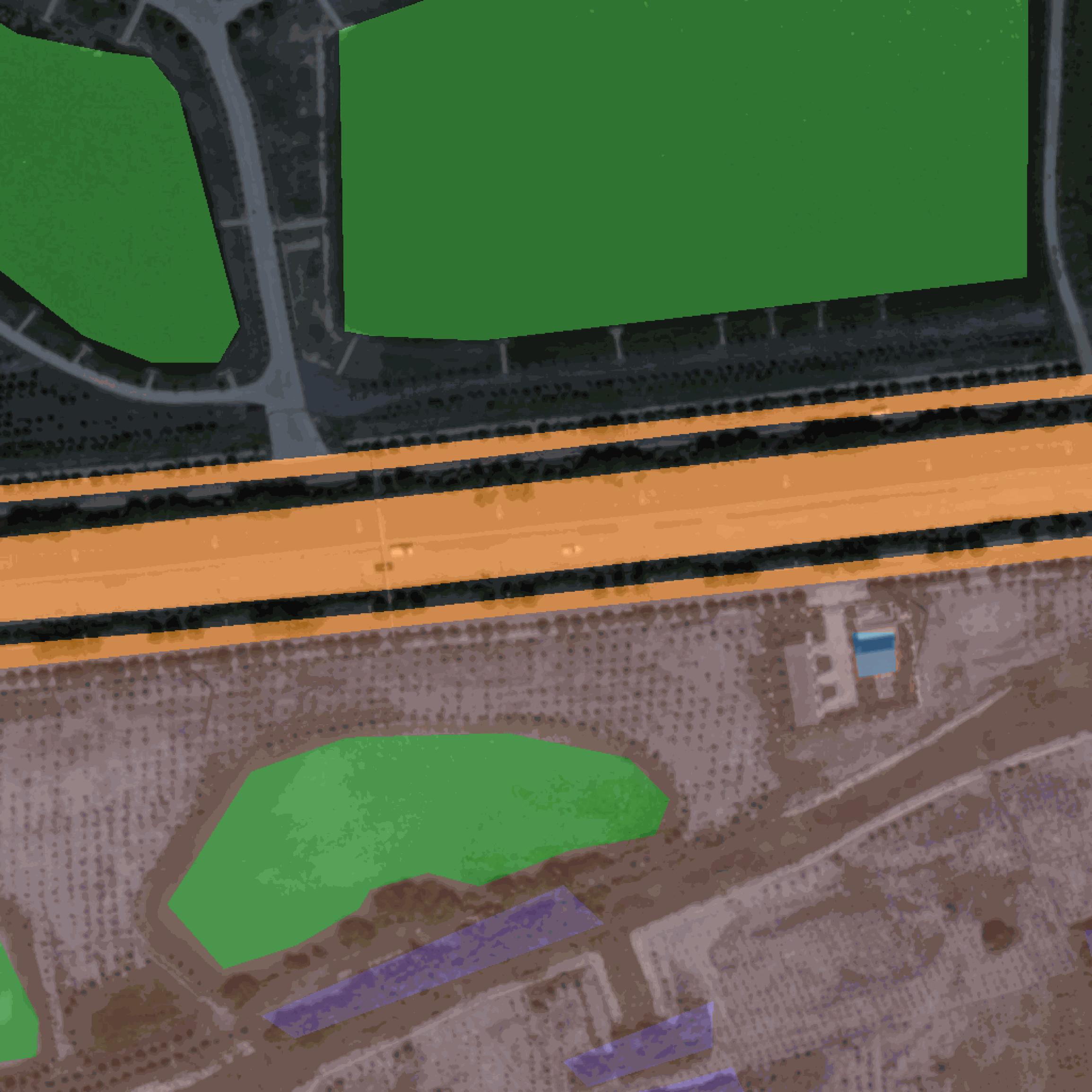} \\
    \end{tabular}

    \caption{Cost maps for a LoveDA image.}
    \label{fig:grid3}
\end{figure*}

\begin{figure*}[]
    \centering
    \setlength{\tabcolsep}{1.5pt}

    \begin{tabular}{c*{9}{c}}
        & Barren & Grass & Pavement & Road & Tree & Water & Cropland & Building & Prediction\\

        \raisebox{21pt}{CAFe-DINO} &
        \includegraphics[width=0.09\textwidth]{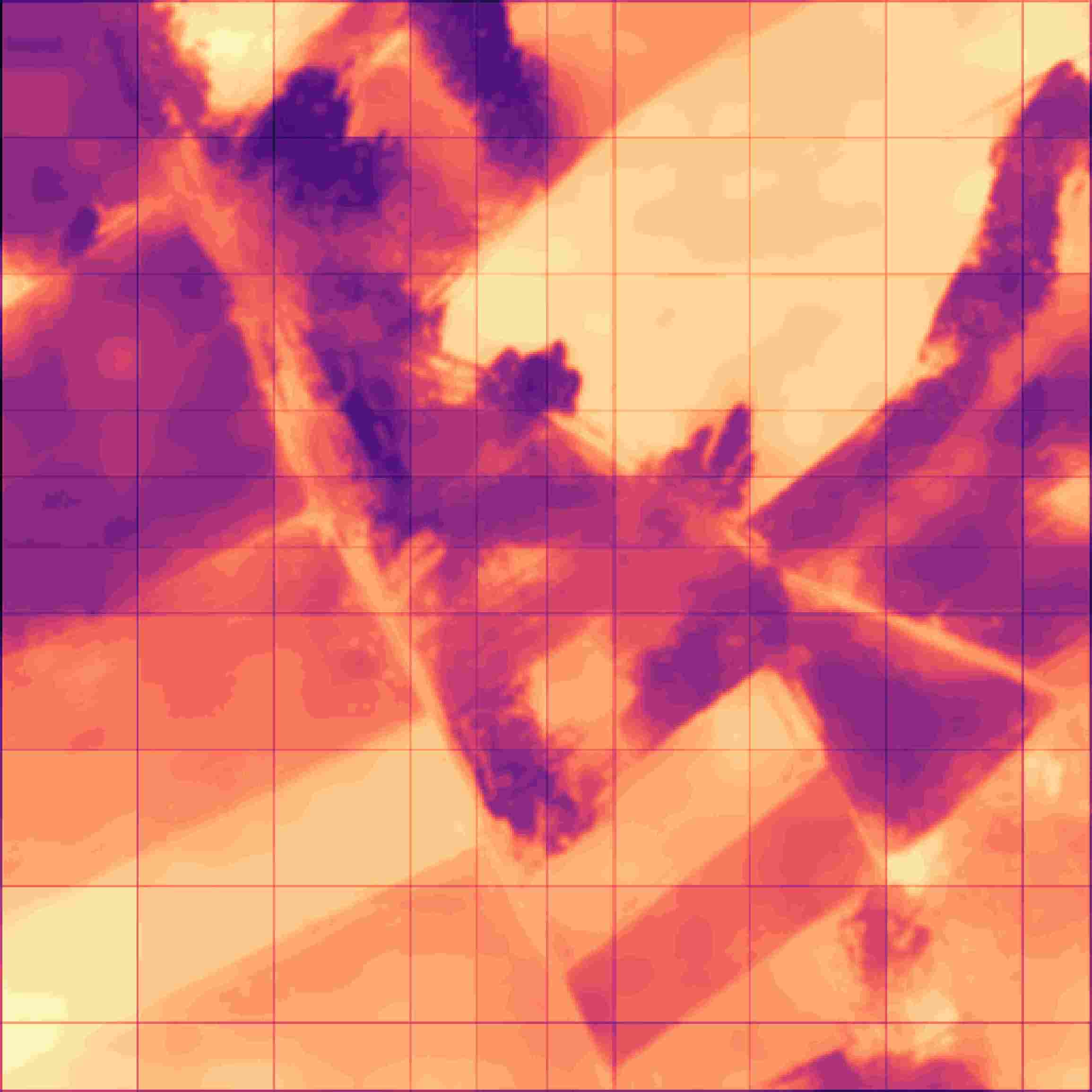} &
        \includegraphics[width=0.09\textwidth]{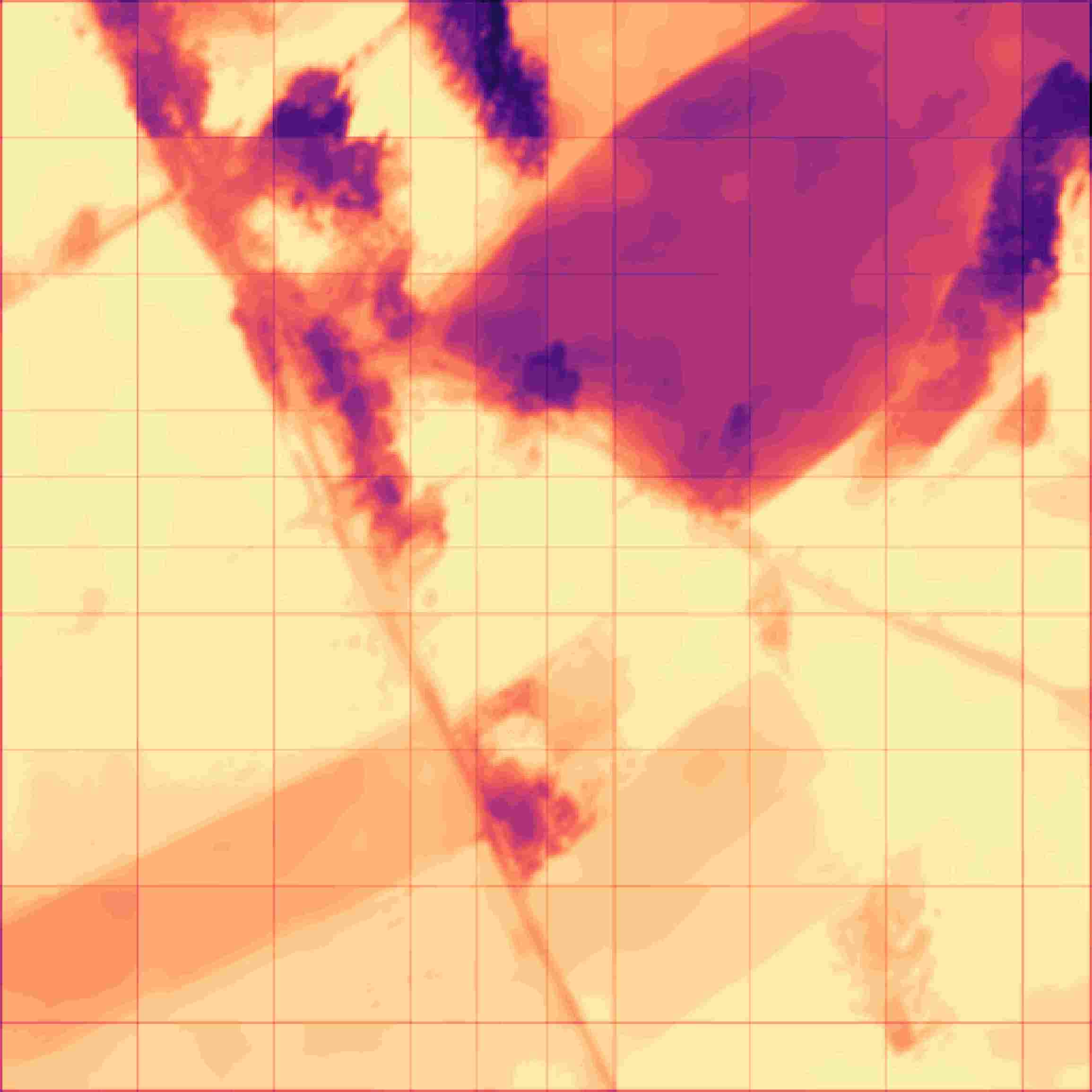}&
        \includegraphics[width=0.09\textwidth]{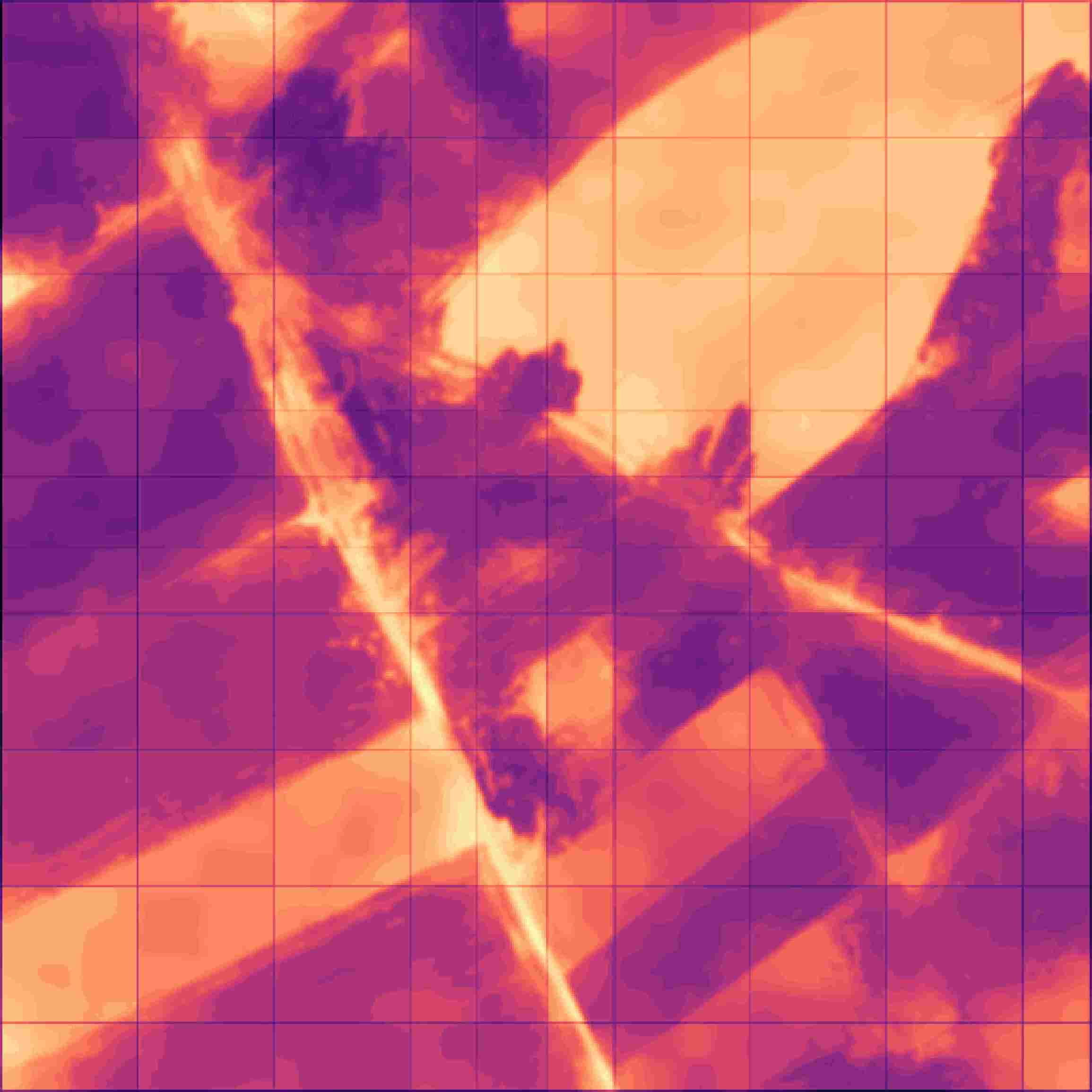}&
        \includegraphics[width=0.09\textwidth]{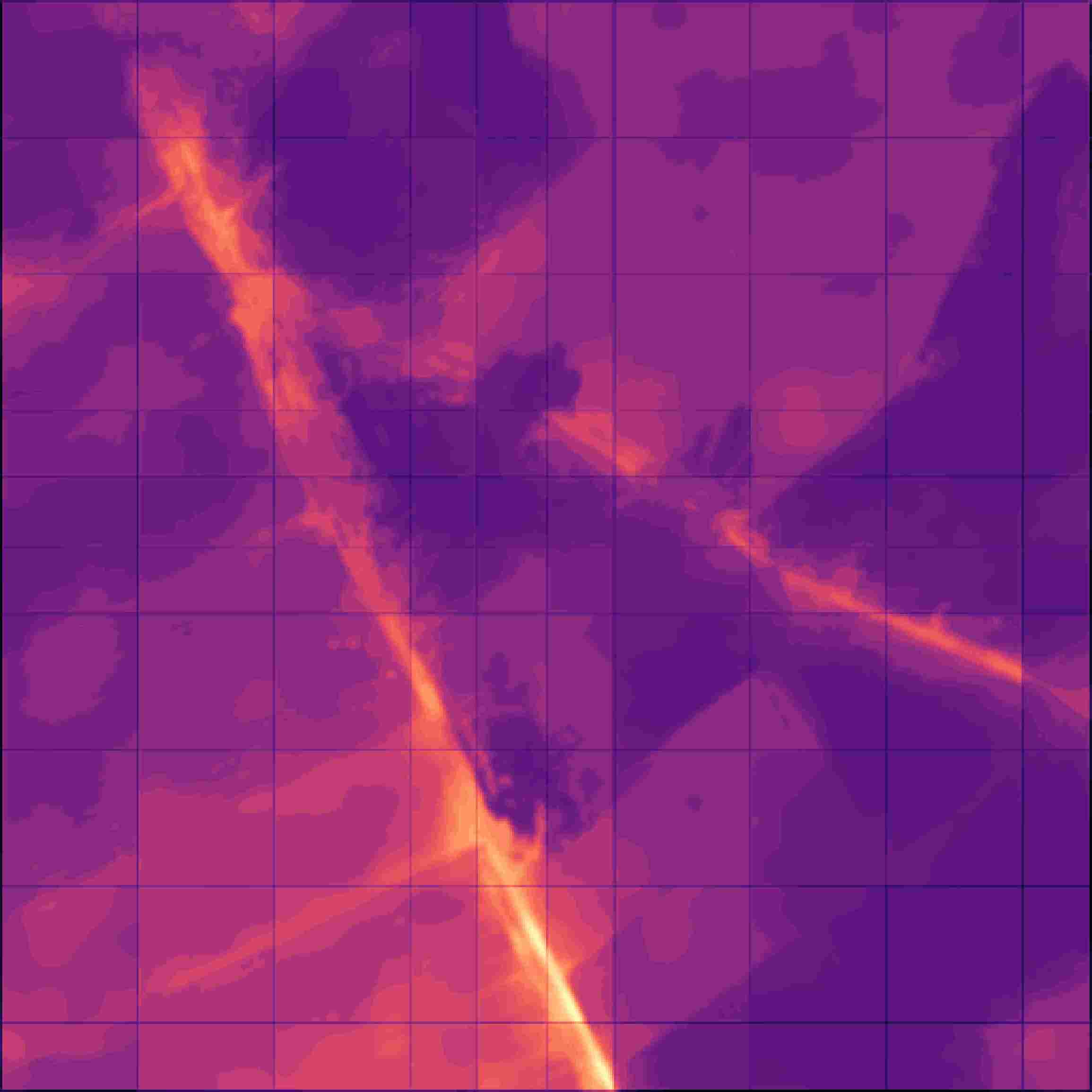}&
        \includegraphics[width=0.09\textwidth]{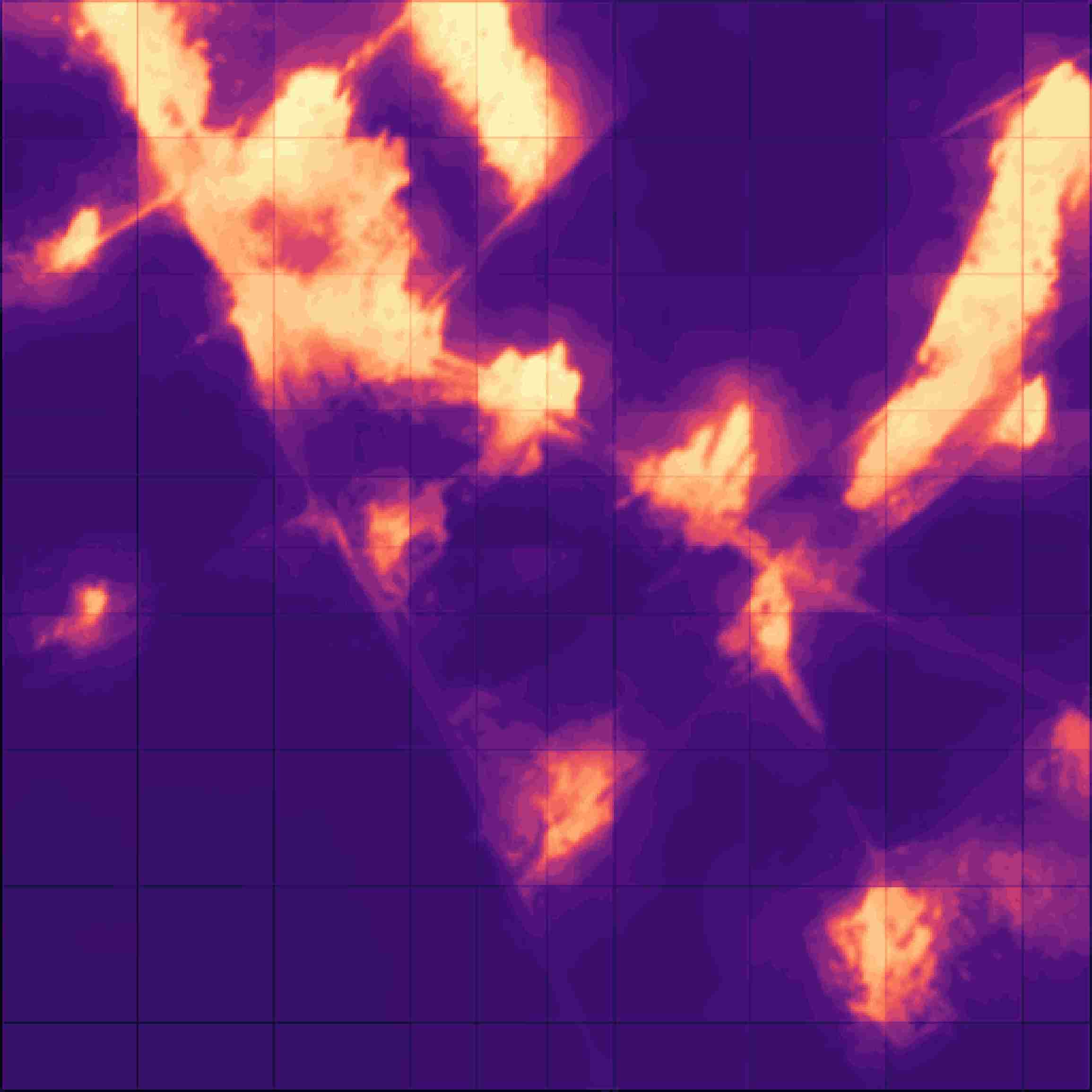}&
        \includegraphics[width=0.09\textwidth]{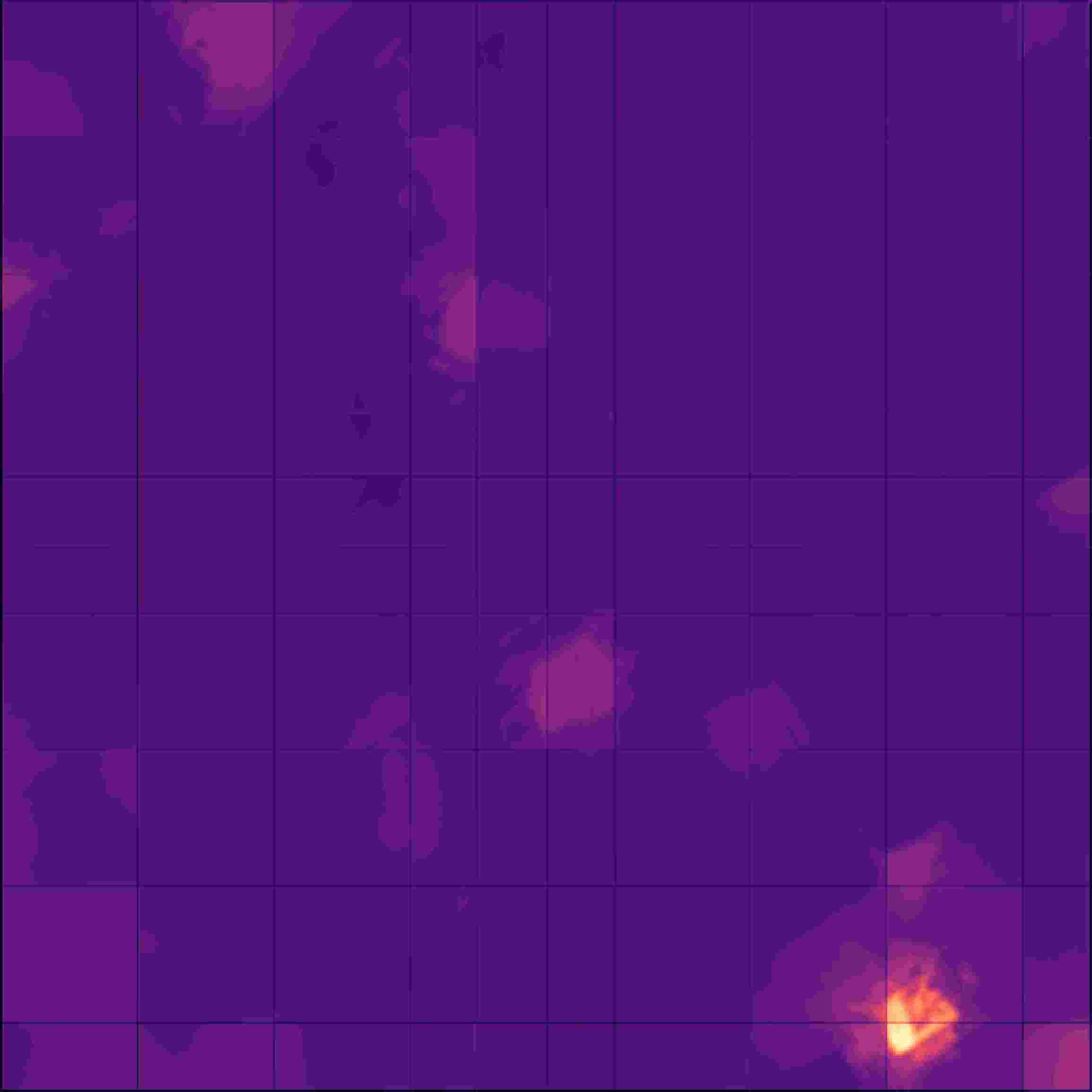}&
        \includegraphics[width=0.09\textwidth]{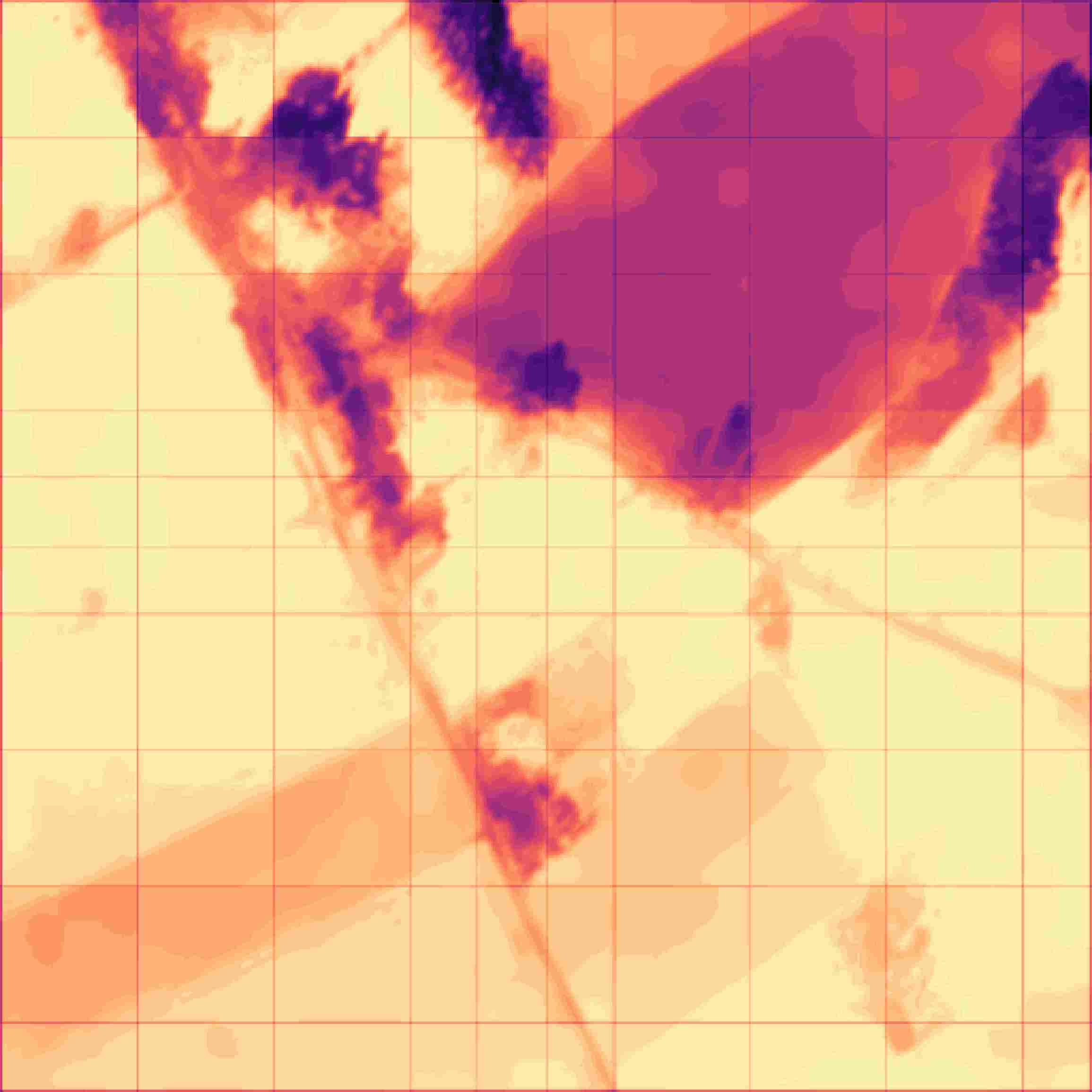}&
        \includegraphics[width=0.09\textwidth]{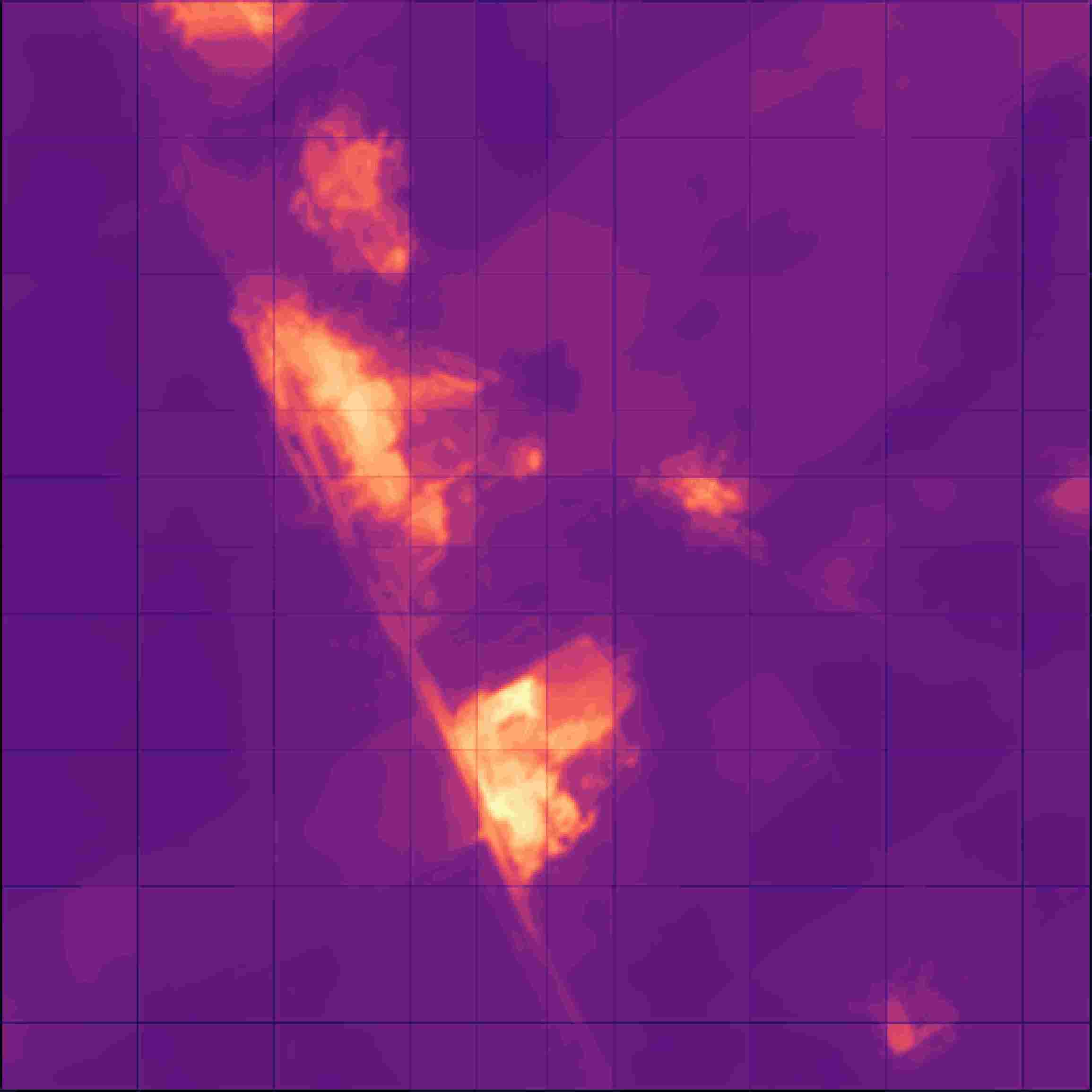}&
        \includegraphics[width=0.09\textwidth]{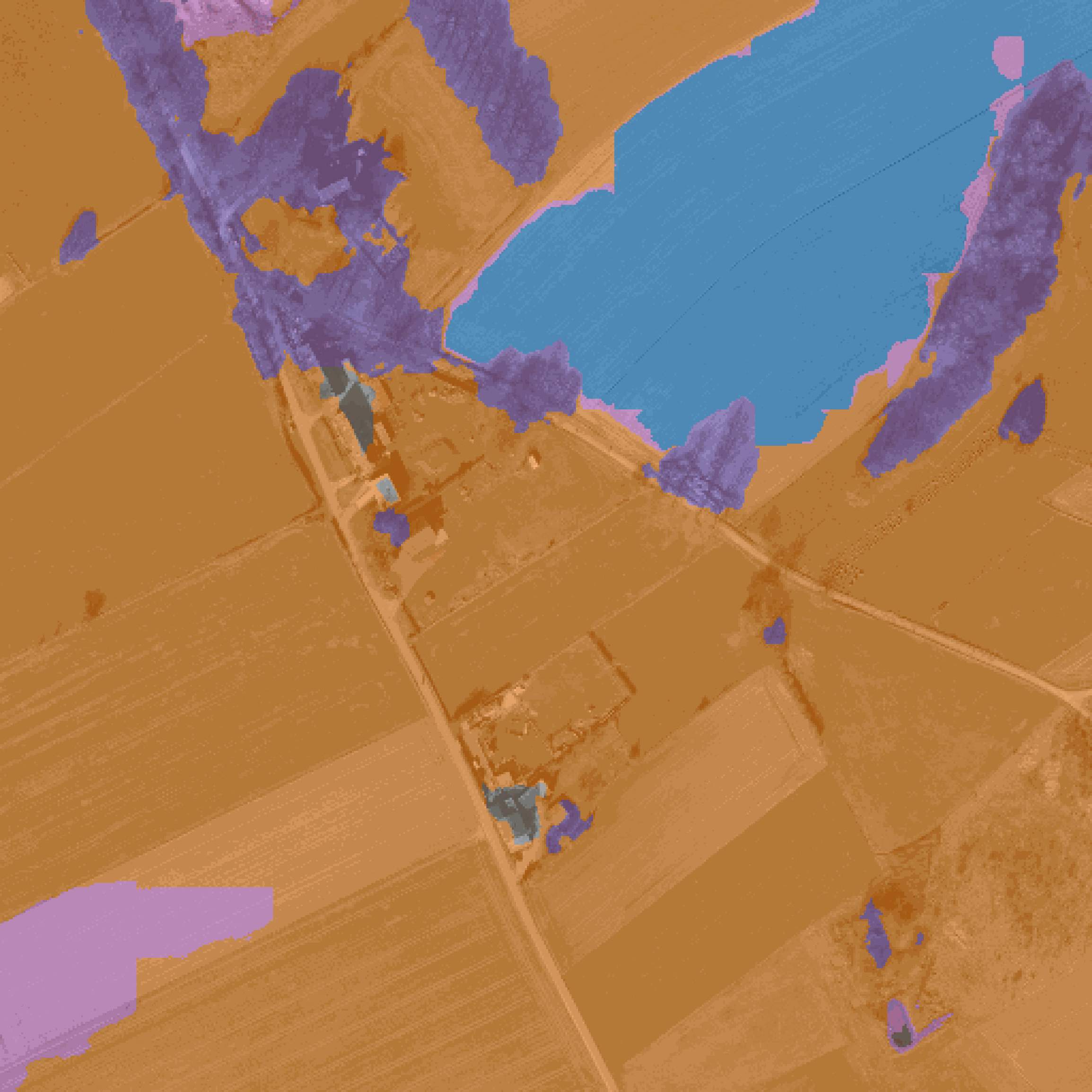} \\

        \raisebox{21pt}{DINOv3} &
        \includegraphics[width=0.09\textwidth]{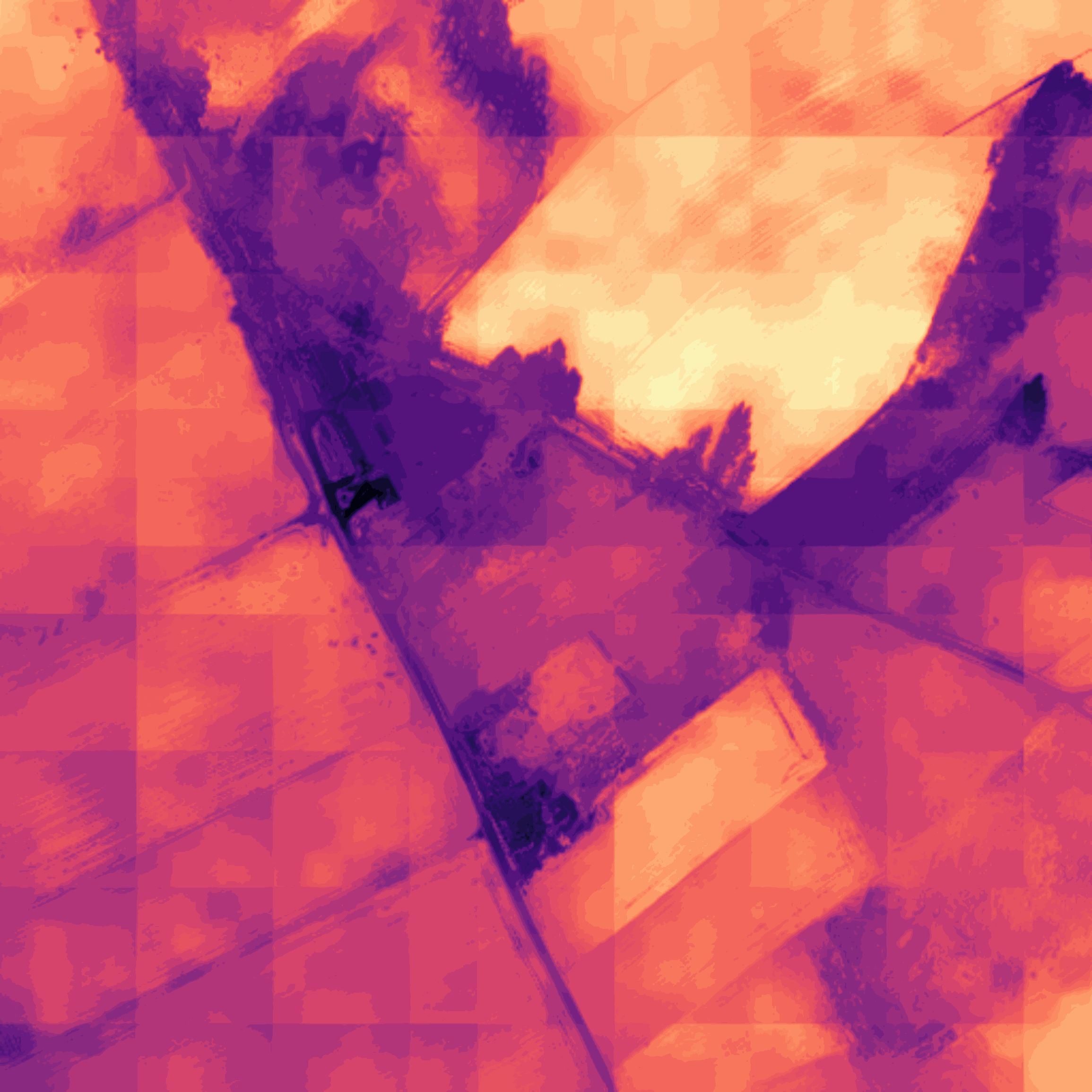} &
        \includegraphics[width=0.09\textwidth]{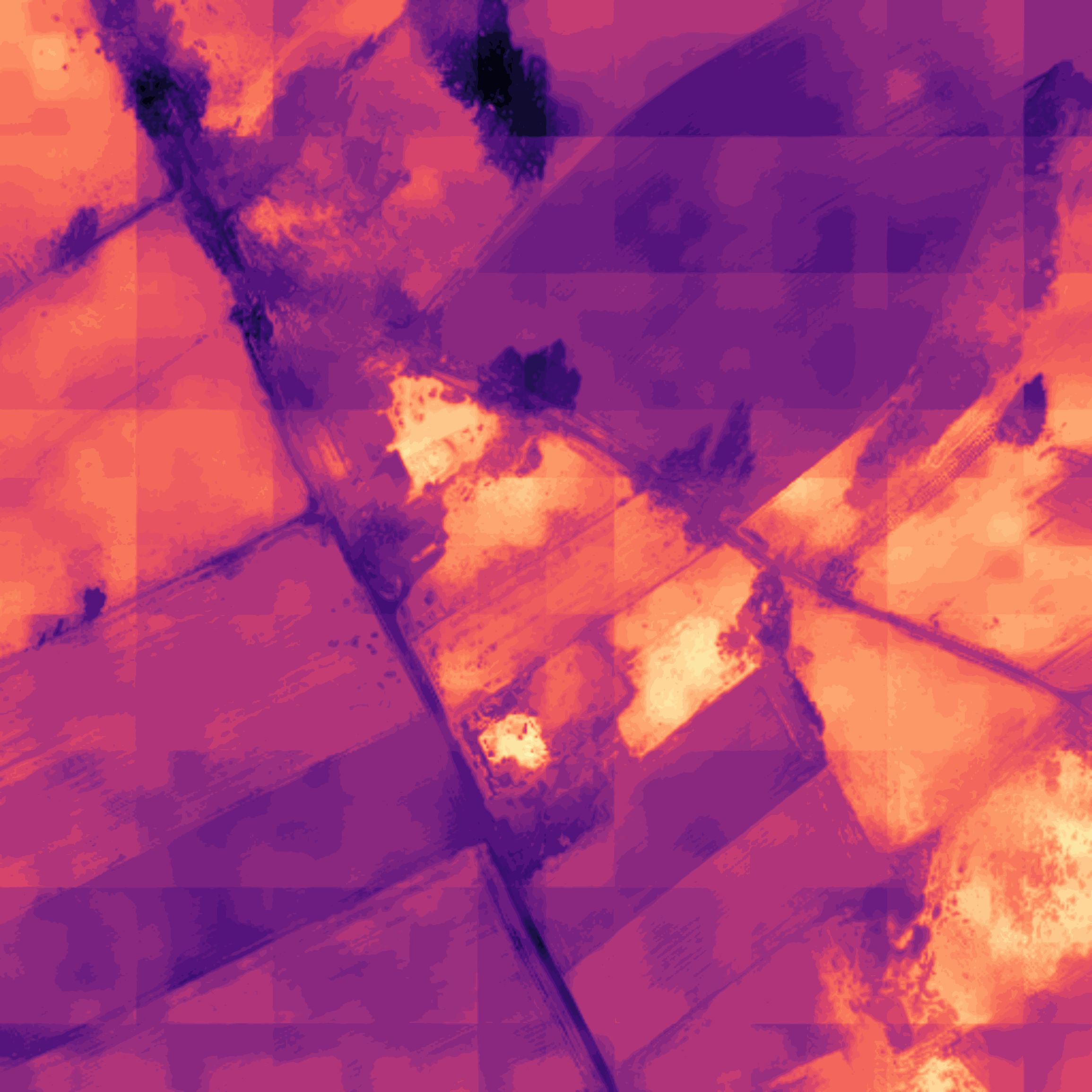} &
        \includegraphics[width=0.09\textwidth]{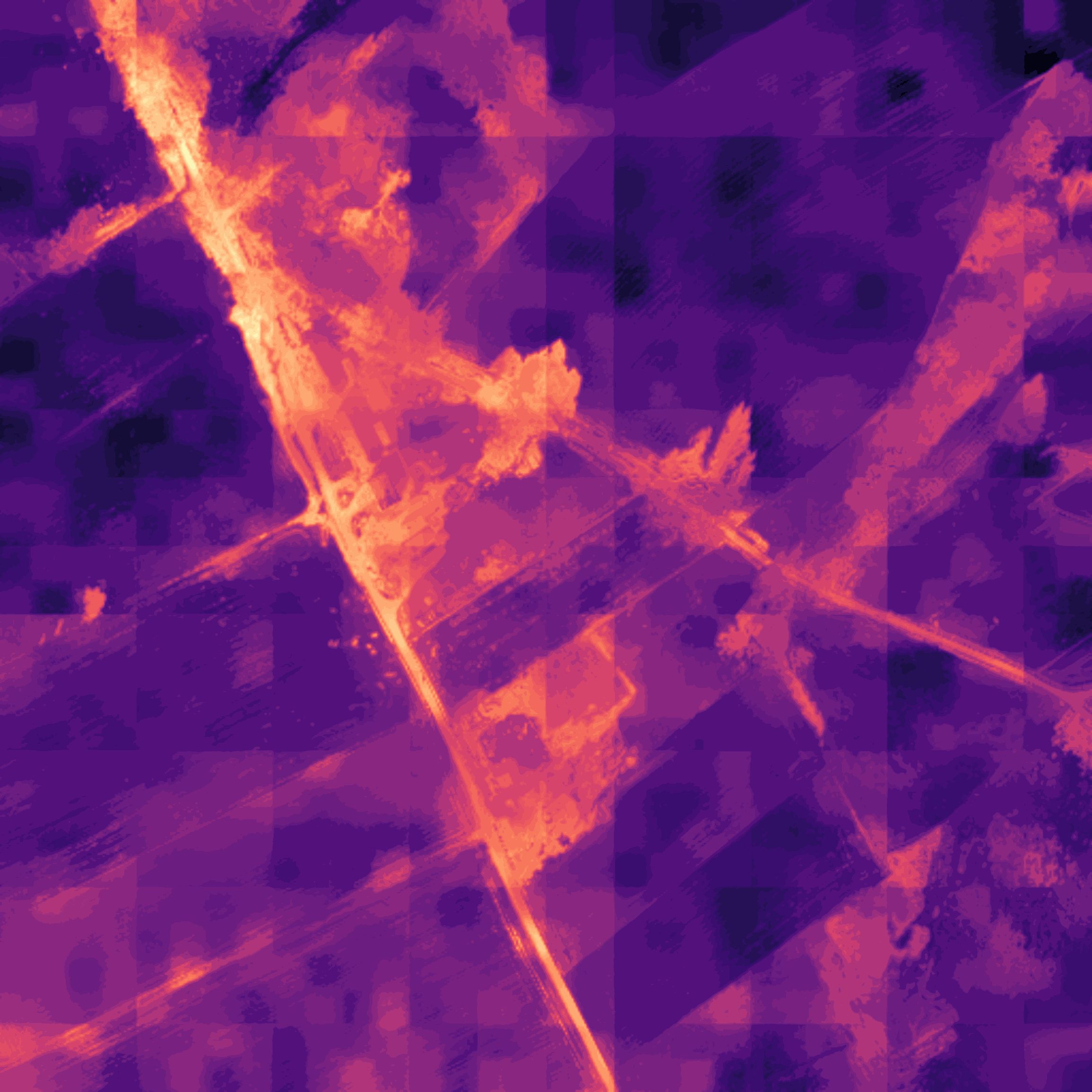} &
        \includegraphics[width=0.09\textwidth]{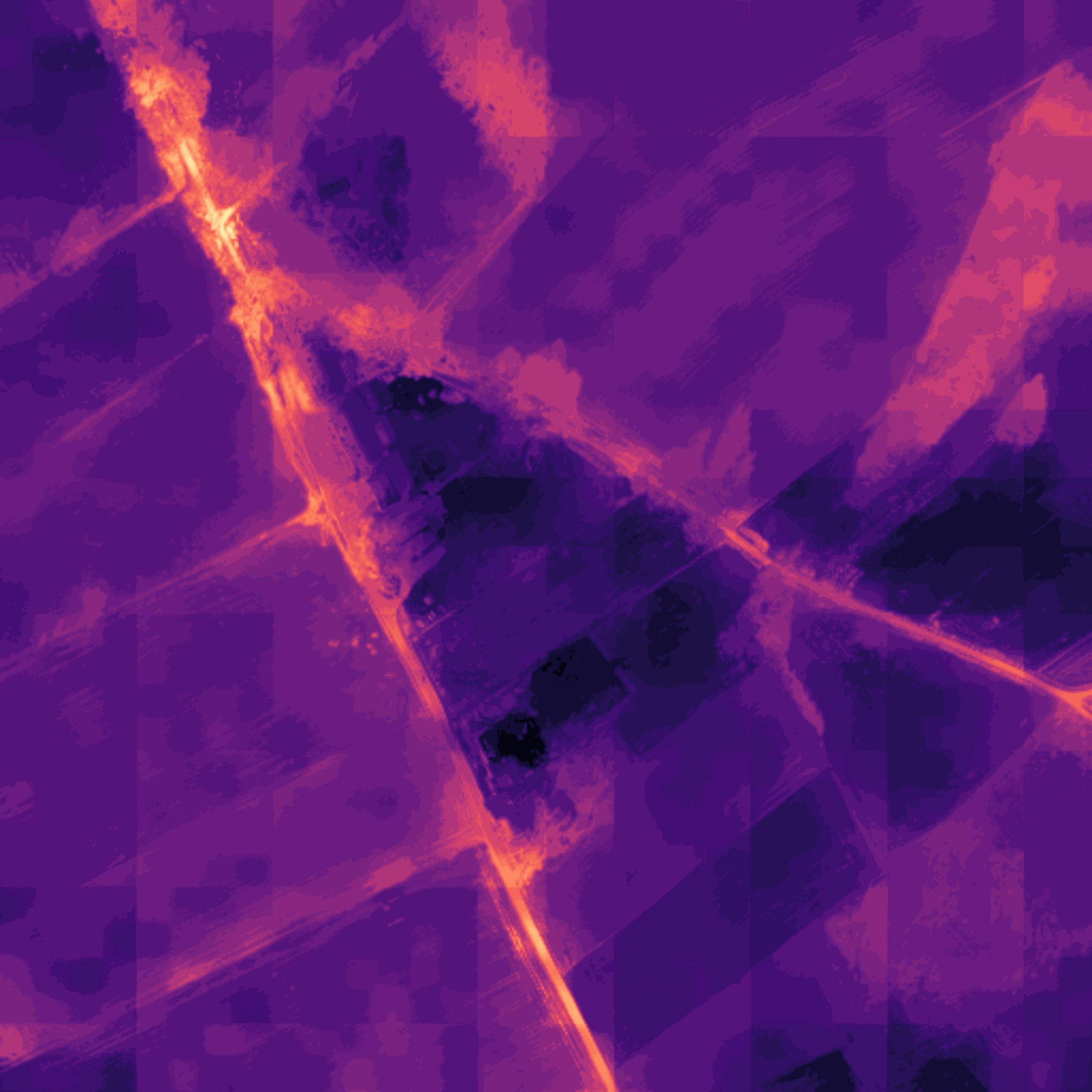} &
        \includegraphics[width=0.09\textwidth]{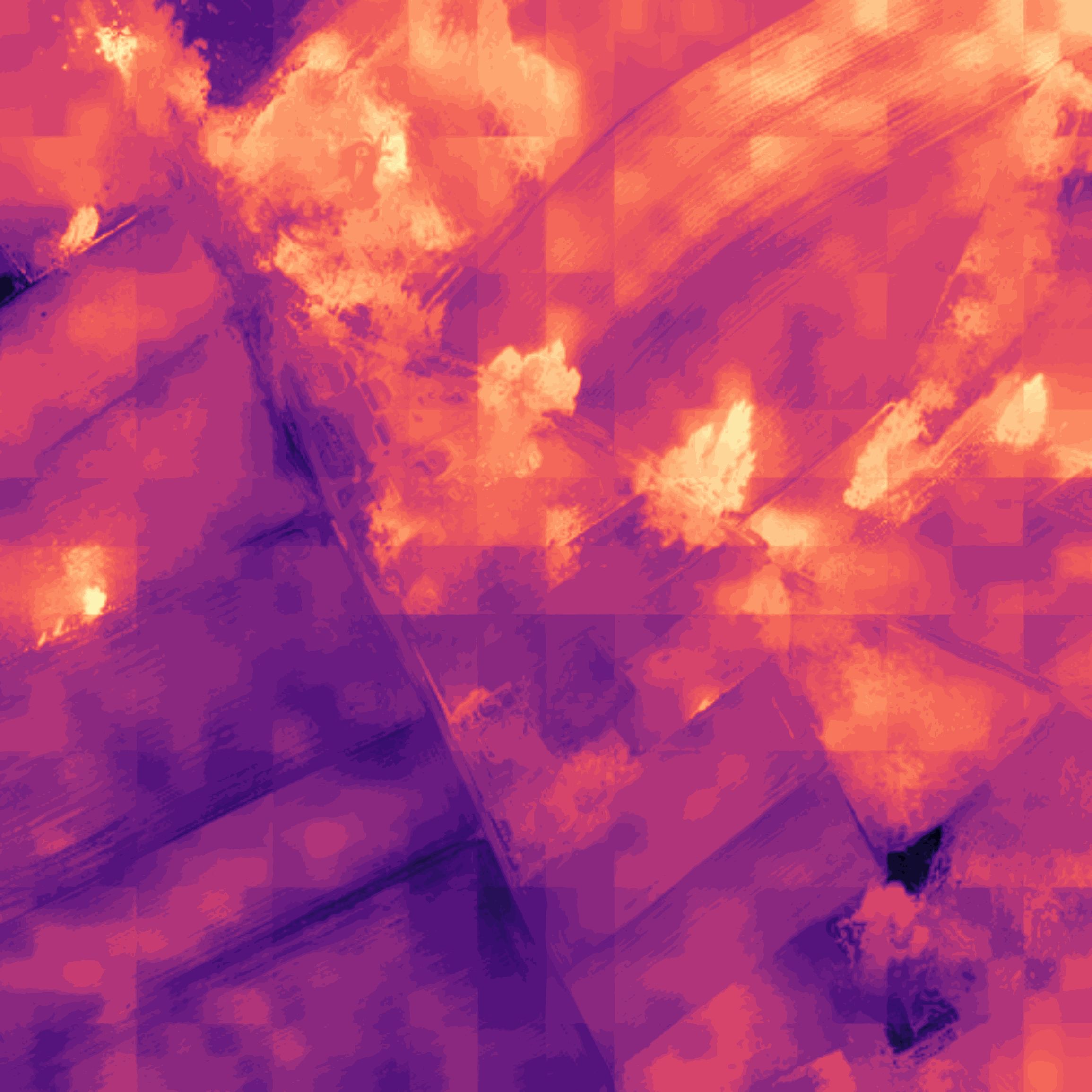} &
        \includegraphics[width=0.09\textwidth]{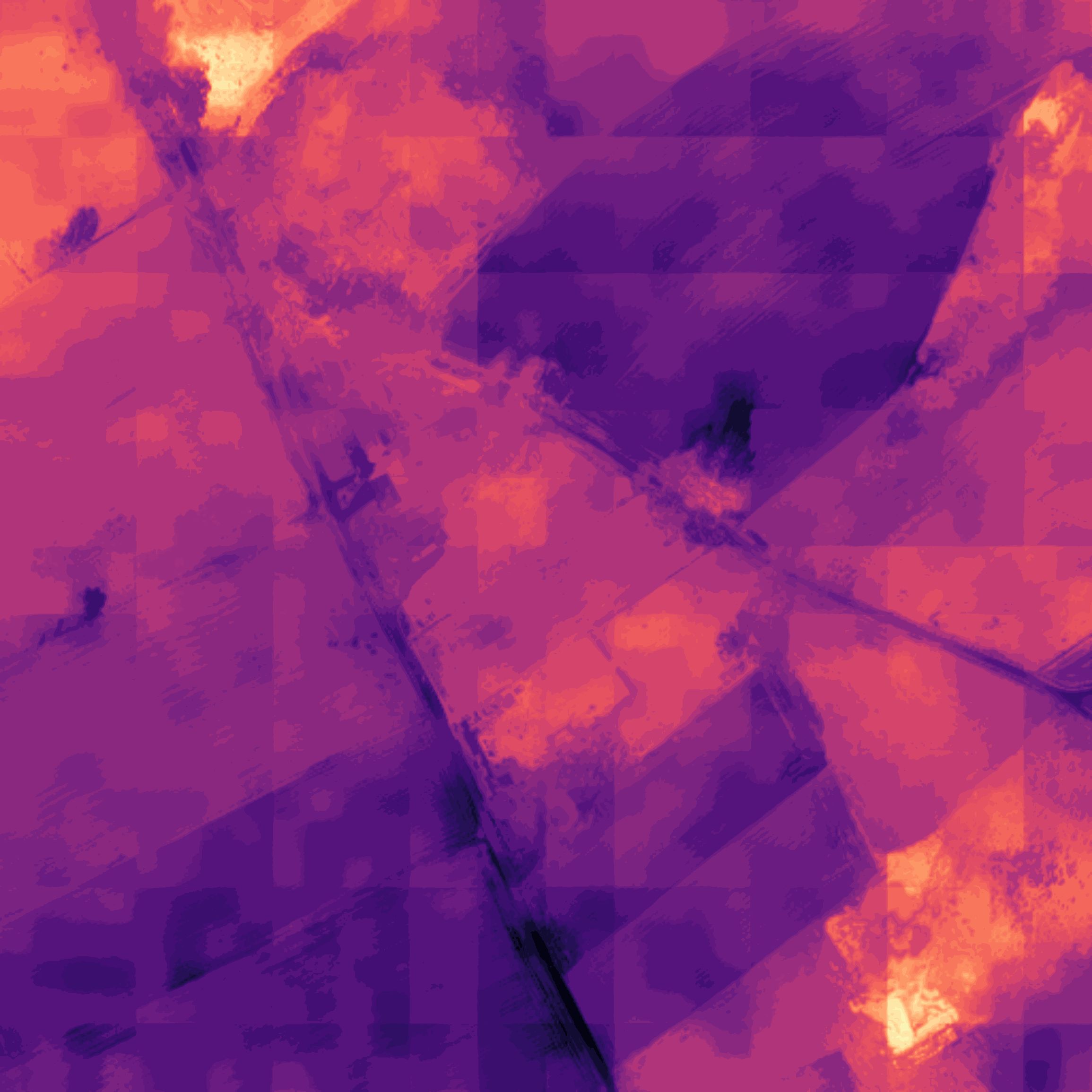} &
        \includegraphics[width=0.09\textwidth]{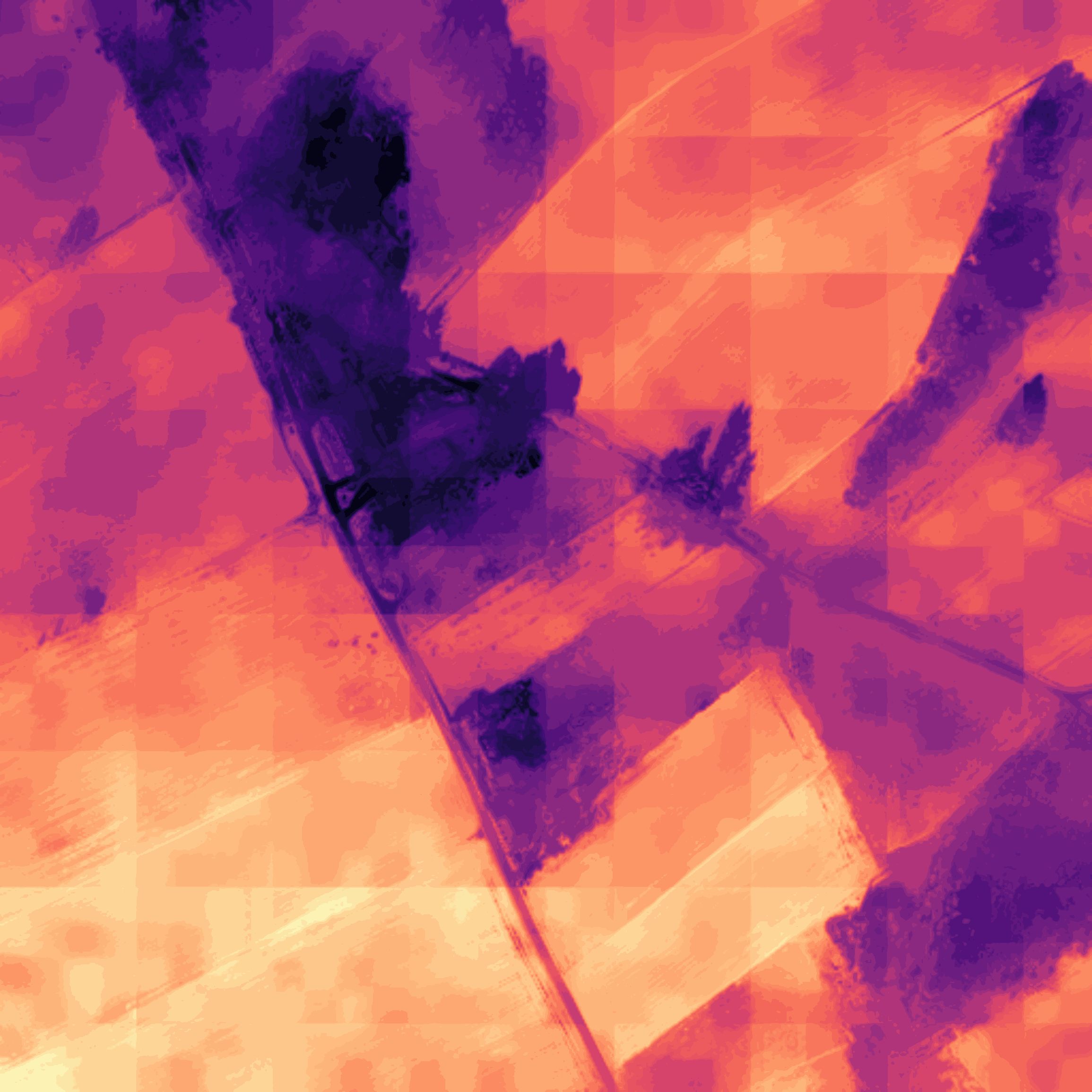} &
        \includegraphics[width=0.09\textwidth]{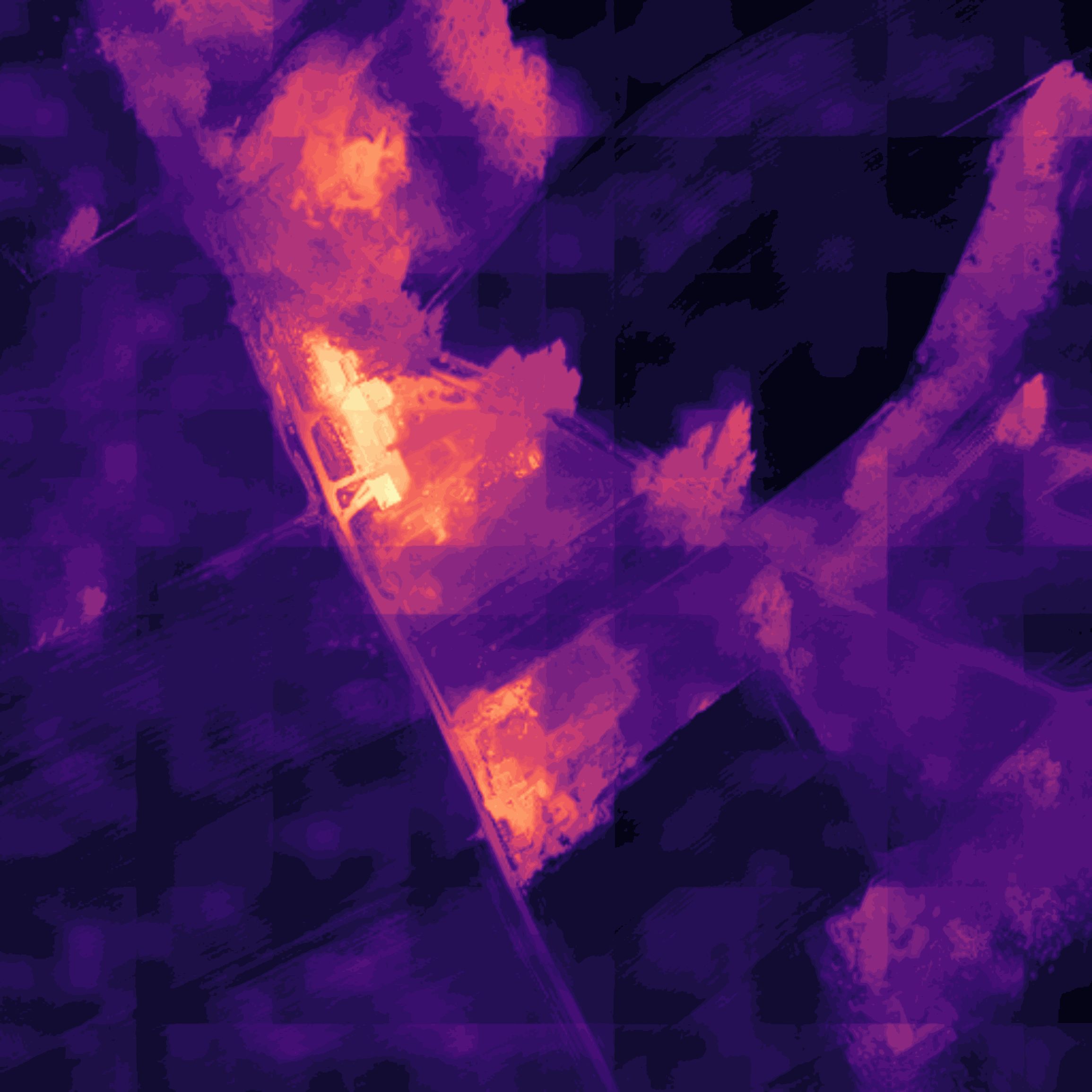} &
        \includegraphics[width=0.09\textwidth]{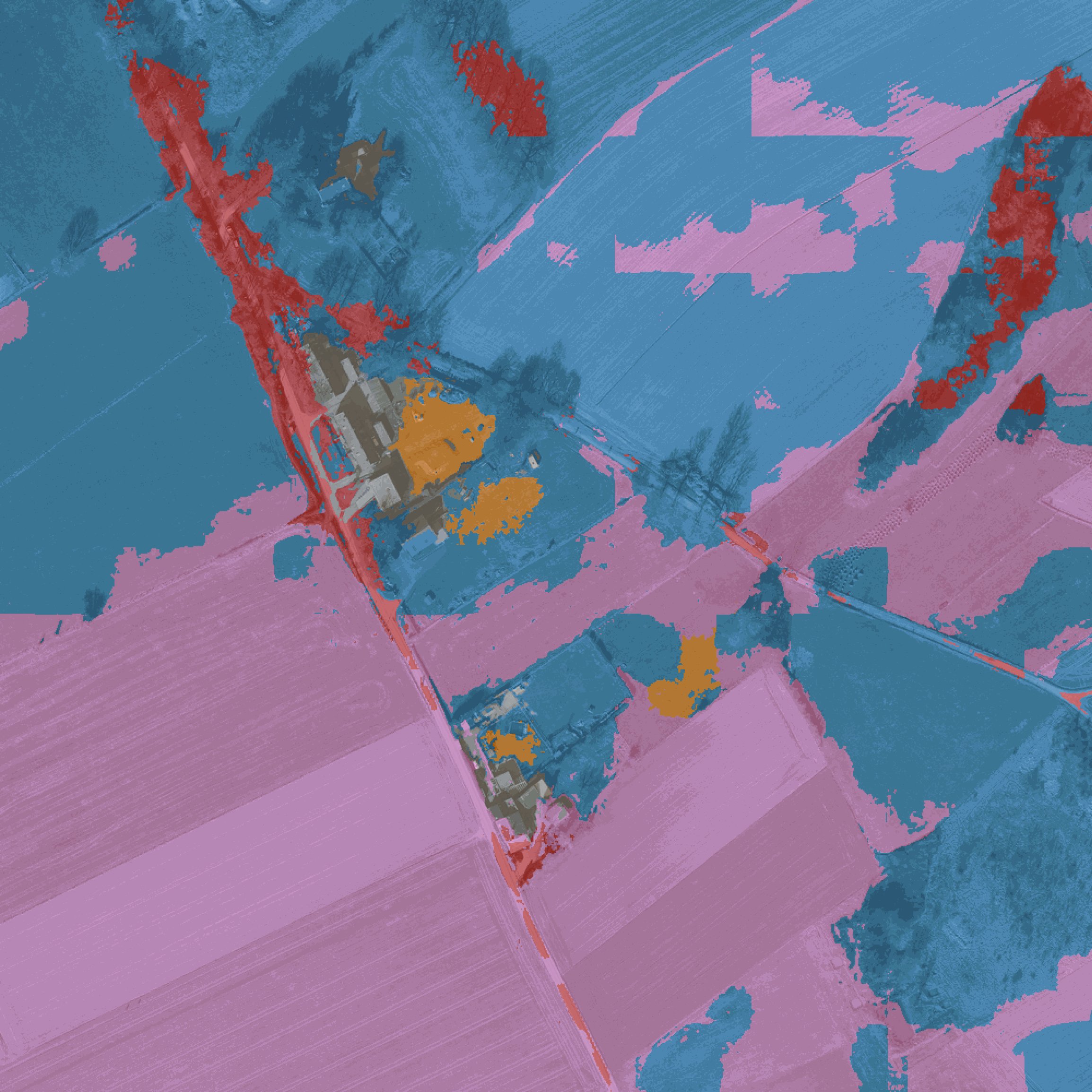} \\
        
        &&&&&&&&\raisebox{21pt}{True Mask}&
            \includegraphics[width=0.09\textwidth]{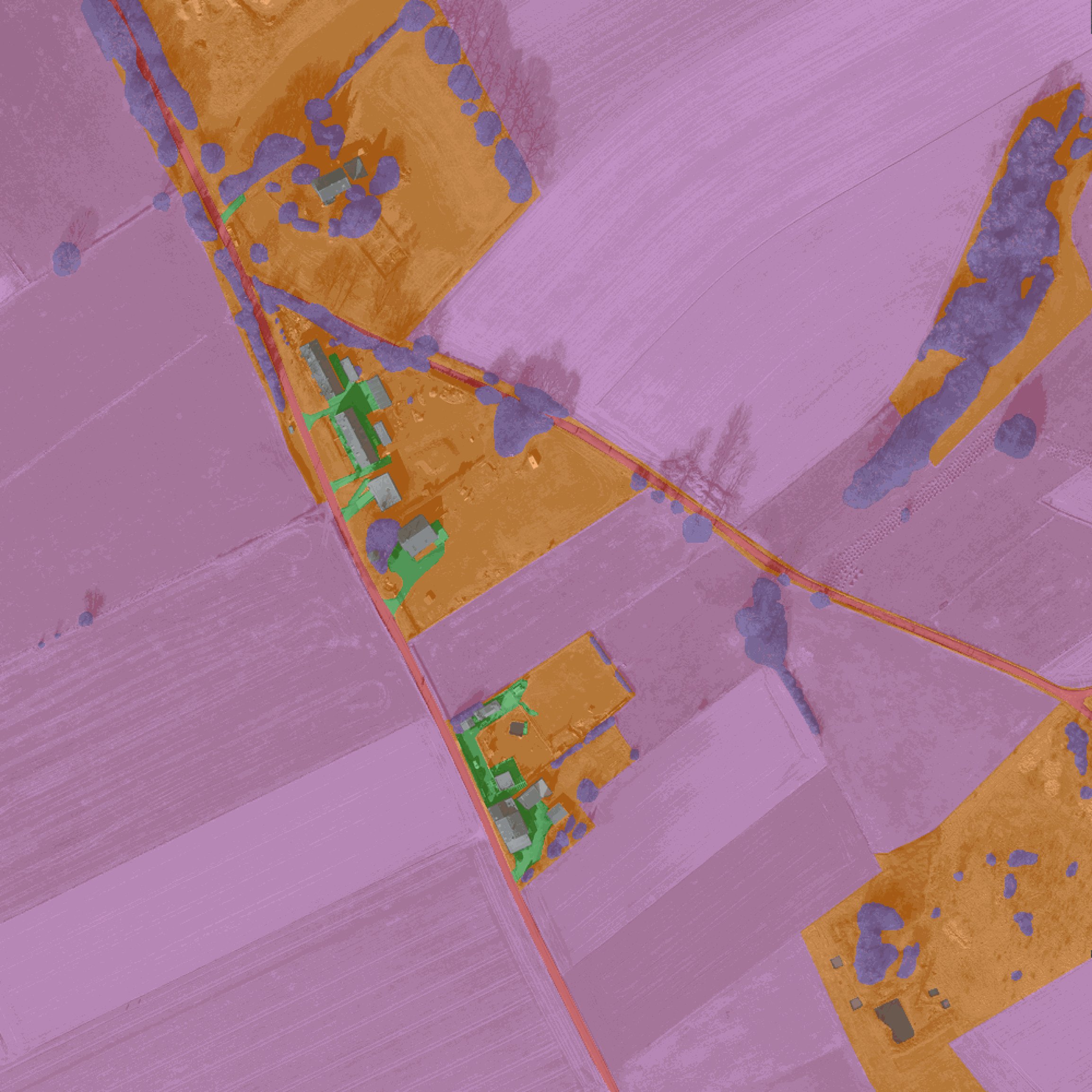} \\
    \end{tabular}

    \caption{Cost maps for an OEM image.}
    \label{fig:grid4}
\end{figure*}

On the following pages, we provide additional cost maps for samples from the
Potsdam (\cref{fig:grid1}),
Vaihingen (\cref{fig:grid2}),
LoveDA (\cref{fig:grid3}), and
OEM (\cref{fig:grid4}) datasets.

\end{document}